\documentclass[10pt]{article} 
\usepackage{amsmath,amsthm,amssymb,hyperref,cleveref,graphicx,url,rotating}
\usepackage{supertabular,url} 
\usepackage{algorithm,blkarray}
\usepackage{fullpage}

\newtheorem{Property}{Property}
\newtheorem{Definition}{Definition}
\newtheorem{Corollary}{Corollary}
\newtheorem{Theorem}{Theorem}
\newtheorem{Example}{Example}

\graphicspath{{Fig/}}

\title{An elementary introduction to information geometry}

\author{Frank Nielsen\\ Sony Computer Science Laboratoties Inc\\ Tokyo, Japan}

 \date{}

\def\calA{\mathcal{A}}

\def\calD{\mathcal{D}}
\def\calE{\mathcal{E}}
\def\calF{\mathcal{F}}

\def\calL{\mathcal{L}}
\def\calM{\mathcal{M}}

\def\calP{\mathcal{P}}

\def\calS{\mathcal{S}}

\def\calU{\mathcal{U}}

\def\calX{\mathcal{X}}
\def\calY{\mathcal{Y}}

\def\bbR{\mathbb{R}}

\def\bbX{\mathbb{X}}

\def\frF{\mathfrak{F}}

\def\frX{\mathfrak{X}}

\def\eqnota{{:=:}}

\def\LC{\mathrm{LC}}
\def\ds{\mathrm{d}s}
\def\dx{\mathrm{d}x}

\def\calF{\mathcal{F}}
\def\ddp{\mathrm{d}p}
\def\Inner#1#2{{\left\langle #1,#2 \right\rangle}}
\def\inner#1#2{{\langle #1,#2 \rangle}}

\def\eqdef{{:=}}
\def\bbR{\mathbb{R}}

\def\bbX{\mathbb{X}}

\def\st{\mbox{\ such that\ }}

\def\calM{\mathcal{M}}

\def\calP{\mathcal{P}}

\def\dnu{\mathrm{d}\nu}

\def\eqEinstein{\stackrel{{\tiny\Sigma}}{=}}
\def\eqsum{\eqEinstein}
\def\eqSum{\eqEinstein}

\def\dtheta{\mathrm{d}\theta}
\def\KL{\mathrm{KL}}
\def\dmu{\mathrm{d}\mu}

\def\leftsup#1{{{}^{#1}}}
\def\leftsub#1{{{}_{#1}}}
\def\leftsubsup#1#2{{{}_{#1}^{#2}}}

\def\idxSup#1{{{}^{#1}}}

\def\eqNota{{:=:}}

\def\nablaLC{{\idxSup{\LC}\nabla}}

\def\Cov{\mathrm{Cov}}

\def\CF{{\idxSup{F}C}}
\def\gF{{\idxSup{F}g}}
\def\CD{{\idxSup{D}C}}
\def\gD{{\idxSup{D}g}}
\def\gDstar{{\idxSup{D^*}g}}
\def\nablaD{{\idxSup{D}\nabla}}
\def\nablaDstar{{\idxSup{D^*}\nabla}}
\def\GammaD{{\idxSup{D}\Gamma}}
\def\GammaDstar{{\idxSup{D^*}\Gamma}}

\def\calL{\mathcal{L}}

\def\leftsup#1{\ ^{#1}}

\def\bbR{\mathbb{R}}
\def\inner#1#2{\langle #1,#2\rangle}
\def\calF{\mathcal{F}}
\def\NG{\mathrm{NG}}
\def\nablaNG{\leftsup{\NG}\nabla}
\def\eqdef{:=}
\def\IS{\mathrm{IS}}
\def\KL{\mathrm{KL}}
\def\dmu{\mathrm{d}\mu}
\def\bbR{\mathbb{R}}
\def\dtheta{\mathrm{d}\theta}
\def\tr{\mathrm{tr}}

\def\Bi{\mathrm{Bi}}
\def\dt{\mathrm{d}t}

\begin{document}

\maketitle
 
\begin{abstract}
In this survey, we describe the fundamental differential-geometric structures of information manifolds, state the fundamental theorem of information geometry, and illustrate some use cases of these information manifolds in information sciences. 
The exposition is self-contained by concisely introducing the necessary concepts of differential geometry, but proofs are omitted for brevity. 
\end{abstract}


\section{Introduction}

\subsection{Overview of information geometry} 
We present a concise and modern view of the basic structures lying at the heart of {\em Information Geometry} (IG), and report some applications of those information-geometric manifolds (herein termed ``information manifolds'') in statistics (Bayesian hypothesis testing) and machine learning (statistical mixture clustering).

By analogy to {\em Information Theory} (IT) (pioneered by Claude Shannon in his celebrated 1948's paper~\cite{Shannon-1948}) which considers primarily the communication of messages over noisy transmission channels,  we may define {\em Information Sciences} (IS) as the fields that study 
``communication'' between
 (noisy/imperfect) data and families of models (postulated as {\em a priori} knowledge). 
In short, information sciences seek methods to {\em distill} information from data to models.
Thus information sciences encompass information theory but also include the fields of Probability \& Statistics, Machine Learning (ML), Artificial Intelligence (AI), Mathematical Programming, just to name a few.

We review some key milestones of information geometry and report some  definitions of the field by its pioneers in~\S\ref{sec:history}.
Professor Shun-ichi Amari, the founder of modern information geometry, defined information geometry in the preface of his latest textbook~\cite{IG-2016} as follows: ``Information geometry is a method of exploring the world of information by means of modern geometry.''
In short, information geometry geometrically investigates information sciences. 
It is a mathematical endeavour to define and bound the term geometry itself as geometry is open-ended.
Often, we start by studying the invariance of a problem (eg., invariance of distance between probability distributions) and get as a result a novel geometric structure (eg., a ``statistical manifold''). However, a geometric structure is ``pure'' and thus may be applied to other application areas beyond the scope of the original problem (eg, use of the dualistic structure of statistical manifolds in mathematical programming~\cite{IG-MP-2013}): the method of geometry~\cite{IG-2000} thus yields a pattern of abduction~\cite{Peirce-1998,Schurz-2008}.

\begin{figure}
\begin{center}
\includegraphics[width=0.4\textwidth]{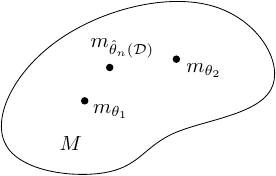}
\end{center}

\caption{The parameter inference $\hat{\theta}$ of a model from data $\calD$ can also be interpreted as a decision making problem: Decide which parameter of a parametric family of models $M=\{m_\theta\}_{\theta\in\Theta}$ suits the ``best'' the data. Information geometry provides a differential-geometric  structure on manifold $M$ which useful for designing and studying statistical decision rules. \label{fig:inference}}

\end{figure}

A narrower definition of information geometry can be stated as the field that studies the {\em geometry of decision making}.
This definition also includes  {\em model fitting} (inference) which can be interpreted as a decision problem as illustrated in Figure~\ref{fig:inference}:
Namely, deciding which model parameter to choose from a family of parametric models.
This framework was advocated by Abraham Wald~\cite{Wald-1949,Wald-1950,GeometryDetectionTheory-1993} who considered all statistical problems as statistical decision problems.
Dissimilarities (also loosely called distances among others) play a crucial role not only for measuring the {\em goodness-of-fit} of data to model (say, likelihood in statistics, classifier loss functions in ML, objective functions in mathematical programming or operations research, etc.) but also for measuring the discrepancy (or deviance) between models.

One may ponder why adopting a geometric approach? 
Geometry allows one to study {\em invariance} of ``figures'' in a coordinate-free framework.
The {\em geometric language} (e.g., line, ball or projection) also provides affordances that help us reason intuitively about problems.
Note that although figures can be visualized (i.e., plotted in coordinate charts), they should be thought of as purely abstract objects, namely, geometric figures.  

Geometry also allows one to study {\em equivariance}:
For example, the centroid $c(T)$ of a triangle is equivariant under any affine transformation $A$: $c(A.T)=A.c(T)$.
In Statistics, the Maximum Likelihood Estimator (MLE) is equivariant under 
a monotonic transformation $g$ of the model parameter $\theta$: $\widehat(g(\theta))=g(\hat{\theta})$, where the MLE of $\theta$ is denoted by $\hat{\theta}$.

\subsection{Outline of the survey}

This survey is organized as follows:

In the first part (\S\ref{sec:dg}), we start by concisely introducing the necessary background on differential geometry  in order to define a manifold structure $(M,g,\nabla)$, ie., a manifold $M$ equipped with a metric tensor field $g$ and an affine connection $\nabla$.
We explain how this framework generalizes the Riemannian manifolds $(M,g)$ by stating the fundamental theorem of Riemannian geometry that defines a unique torsion-free metric-compatible Levi-Civita connection which can be derived from the metric tensor.

In the second part (\S\ref{sec:conjmfd}), we explain the dualistic structures of information manifolds: We present the conjugate connection manifolds $(M,g,\nabla,\nabla^*)$, the statistical manifolds $(M,g,C)$ where $C$ denotes a cubic tensor, and show how to derive a family of information manifolds $(M,g,\nabla^{-\alpha},\nabla^\alpha)$ for $\alpha\in\bbR$ provided any given pair $(\nabla=\nabla^{-1},\nabla^*=\nabla^1)$ of conjugate connections. 
We explain how to get conjugate connections $\nabla$ and $\nabla^*$ coupled to the metric $g$ from any smooth (potentially asymmetric) distances (called divergences), present the dually flat manifolds obtained when considering Bregman divergences, and define, when dealing with parametric family of probability models, the exponential connection $\leftsup{e}\nabla$ and the mixture connection $\leftsup{m}\nabla$ that are dual connections coupled to the Fisher information metric. 
We discuss the concept of statistical invariance for the metric tensor and the notion of information monotonicity for statistical divergences~\cite{Corcuera-1998,IG-2016}. 
It follows that the Fisher information metric  is the unique invariant metric (up to a scaling factor), and 
that the $f$-divergences are the unique separable invariant divergences. 

In the third part (\S\ref{sec:AppIG}), we illustrate how to use these information-geometric structures in   simple applications:
First, we described the natural gradient descent method in~\S\ref{sec:appng} and its relationships with the Riemannian gradient descent and the Bregman mirror descent.
Second, we consider two applications in dually flat spaces in~\S\ref{sec:AIG}:
In the first application, we consider the problem of Bayesian hypothesis testing and show how Chernoff information (which defines the best error exponent) 
can be geometrically  characterized on the dually flat structure of an exponential family manifold.
In the second application, we show how to cluster statistical mixtures sharing the same component distributions on the dually flat mixture family manifold.

Finally, we conclude in~\S\ref{sec:concl}  by summarizing the important concepts and structures of information geometry, and by providing  further references and textbooks~\cite{IG-2014,IG-2016} for further readings to more advanced structures and applications of information geometry.
We also mention recent studies of generic classes of principled distances and divergences.

In the Appendix~\S\ref{sec:MCfdiv}, we show how to estimate the statistical $f$-divergences between two probability distributions in order to ensure that the estimates are non-negative in \S\ref{sec:mvn}, and
report the canonical decomposition of the multivariate Gaussian family, an example of exponential family which admits a dually flat structure.

At the beginning of each part, we start by outlining its contents.
A summary of the notations used throughout this survey is provided page~\pageref{sec:notations}.

\section{Prerequisite: Basics of differential geometry}\label{sec:dg}

In~\S\ref{sec:odg}, we review the very basics of Differential Geometry (DG) for defining a manifold $(M,g,\nabla)$ equipped with both a metric tensor field $g$ and an affine connection $\nabla$.
We explain these two {\em independent} metric/connection structures in~\S\ref{sec:mt} and in~\S\ref{sec:affc}, respectively.
From an affine connection $\nabla$, we show how to derive the notion of  covariant derivative in~\S\ref{sec:codev},  parallel transport in~\S\ref{sec:partransp} 
and   geodesics in~\S\ref{sec:geo}. We further explain the {\em intrinsic curvature and torsion} of manifolds induced by the connection in~\S\ref{sec:curvaturetorsion}, and state the fundamental theorem of Riemannian geometry in~\S\ref{sec:fthmrie}: The existence of a unique torsion-free Levi-Civita connection $\nablaLC$ compatible with the metric (metric connection) that can be derived from the metric tensor $g$. 
Thus the Riemannian geometry $(M,g)$ is obtained as a special case of the more general manifold structure $(M,g,\nablaLC)$: $(M,g)\equiv (M,g,\nablaLC)$.
Information geometry shall further consider a dual structure $(M,g,\nabla^*)$ associated to $(M,g,\nabla)$, and the pair of dual structures shall form an information manifold $(M,g,\nabla,\nabla^*)$.

\subsection{Overview of differential geometry: Manifold $(M,g,\nabla)$}\label{sec:odg}

Informally speaking, a {\em smooth $D$-dimensional manifold $M$} is a topological space that locally behaves like the $D$-dimensional  Euclidean space $\bbR^D$.
Geometric objects  (e.g., points, balls, and vector fields) and entities (e.g., functions and differential operators) live on $M$,  and are {\em coordinate-free} but can conveniently be expressed in {\em any} local coordinate system of an atlas $\calA=\{(\calU_i,x_i)\}_i$ of charts$(\calU_i,x_i)$'s (fully covering the manifold) for calculations.
Historically, Ren\'e Descartes (1596-1650) allegedly invented the global Cartesian coordinate system while wondering how to locate  a fly on the ceiling from his bed. In practice, we shall use the most expedient coordinate system to facilitate calculations.
In information geometry,  we usually handle a single chart fully covering the manifold.

A $C^k$ manifold is obtained when the {\em change of chart transformations} are $C^k$.
The manifold is said smooth when it is $C^\infty$. 
At each point $p\in M$, a tangent plane $T_p$ locally best linearizes the manifold.
On any smooth manifold $M$, we can define two \underline{\em independent} structures:
\begin{enumerate}
\item a metric tensor $g$,  and 
\item an affine connection $\nabla$.
\end{enumerate}

The metric tensor $g$  induces on each tangent plane $T_p$ an {\em inner product space} that allows one to measure vector magnitudes (vector ``lengths'') and angles/orthogonality between vectors. 
The affine connection $\nabla$ is a differential operator that allows one to define:

\begin{enumerate}
\item the {\em covariant derivative operator} which provides a way to calculate differentials of a vector field $Y$ with respect to another vector field $X$: Namely, the covariant derivative $\nabla_X Y$, 

\item the {\em parallel transport} $\prod^\nabla_c$ which defines a way to transport vectors between tangent planes along any smooth curve $c$,  

\item the notion of $\nabla$-geodesics $\gamma_\nabla$  which are defined as autoparallel curves, thus extending the ordinary notion of Euclidean straightness,  

\item the intrinsic curvature and torsion of the manifold.
\end{enumerate}

\subsection{Metric tensor fields $g$}\label{sec:mt}

The {\em tangent bundle} of $M$ is defined as the ``union'' of all tangent spaces:
\begin{equation}
TM \eqdef \cup_p T_p=\{(p,v),\quad p\in M, v\in T_p\}.
\end{equation}
Thus the tangent bundle $TM$ of a $D$-dimensional manifold $M$ is of dimension $2D$. 
(The tangent bundle is a particular example of a fiber bundle with base manifold $M$.)

Informally speaking, a {\em tangent vector} $v$ plays the role of a directional derivative, with $v f$ informally meaning the derivative of a smooth function $f$ 
 (belonging to the space of smooth functions $\frF(M)$) along the direction $v$.
Since the manifolds are abstract and not embedded in some Euclidean space, we do not view a vector as an ``arrow'' anchored on the manifold. Rather, vectors can be understood in several ways in differential geometry like directional derivatives or  equivalent class of smooth curves at a point. That is, tangent spaces shall be considered as the manifold abstract too.

\begin{figure}
\begin{center}
\includegraphics[width=0.4\textwidth]{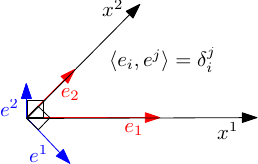}
\end{center}

\caption{Primal basis (red) and reciprocal basis (blue) of an inner product $\inner{\cdot}{\cdot}$ space.
The primal/reciprocal basis are mutually orthogonal: 
$e^1$ is orthogonal to $e_2$, and $e_1$ is orthogonal to $e^2$.\label{fig:pdbasis}}
\end{figure}

A smooth {\em vector field} $X$ is defined as a ``cross-section'' of the tangent bundle: $X\in\frX(M)=\Gamma(TM)$, where
 $\frX(M)$ or $\Gamma(TM)$ denote the space of smooth vector fields.
A basis $B=\{b_1,\ldots, b_D\}$ of a finite $D$-dimensional vector space is 
a {\em maximal linearly independent set of vectors}:
A set of vectors $B=\{b_1,\ldots, b_D\}$ is linearly independent if and only if $\sum_{i=1}^D \lambda_i b_i=0$ iff $\lambda_i=0$ for all $i\in [D]$.
That is, in a linearly independent vector set, no vector of the set can be represented as a linear combination of the remaining vectors.
A vector set is linearly independent  maximal when we cannot add another linearly independent vector.
Tangent spaces carry algebraic structures of vector spaces.
Furthermore, to any vector space $V$, we can associate a dual covector space $V^*$ which is the vector space of real-valued linear mappings. We do not enter into details here to preserve this gentle introduction to information geometry with as little intricacy as possible.
Using local coordinates on a chart $(\calU,x)$, the vector field $X$ can be expressed as 
$X=\sum_{i=1}^D  X^i e_i \eqSum X^i e_i$ using Einstein summation convention on dummy indices (using notation $\eqSum$), where
$(X)_B\eqdef (X^i)$ denotes the {\em contravariant vector components} (manipulated as ``column vectors'' in algebra) in the {\em natural basis} $B=\{e_1=\partial_1,\ldots, e_D=\partial_D\}$ with $\partial_i\eqNota \frac{\partial}{\partial x_i}$.  
A tangent plane (vector space) equipped with an {\em inner product} $\inner{\cdot}{\cdot}$ yields an {\em inner product space}.
We define a {\em reciprocal basis} $B^*=\{{e^*}^i=\partial^i\}_i$  of $B=\{e_i=\partial_i\}_i$ so that vectors can also be expressed 
using the {\em covariant vector components} in the natural reciprocal basis. 
The primal and reciprocal basis are {\em mutually orthogonal} by construction as illustrated in Figure~\ref{fig:pdbasis}.
 
For any vector $v$, its contravariant components $v^i$'s (superscript notation) and its covariant components $v_i$'s (subscript notation) can be retrieved from $v$ using  the inner product with the use of the reciprocal and primal basis, respectively:
\begin{eqnarray}
v^i &=& \inner{v}{{e^*}^i},\\
v_i &=& \inner{v}{e_i}.
\end{eqnarray}

The inner product defines a {\em metric tensor} $g$ and a {\em dual metric tensor} $g^*$:
\begin{eqnarray}
g_{ij}&\eqdef &\inner{e_i}{e_j},\\
{g^*}^{ij}&\eqdef&\inner{{e^*}^i}{{e^*}^j}.
\end{eqnarray}

Technically speaking, the metric tensor $g_p: T_pM \times T_pM\rightarrow \mathbb{R}$ is a $2$-covariant tensor field:
\begin{equation}
g \eqsum g_{ij} \dx_i \otimes \dx_j,
\end{equation} 
where $\otimes$ is the dyadic tensor product performed on pairwise covector basis $\{\dx_i\}_i$ (the covectors corresponding to the reciprocal vector basis).
We do not describe tensors in details for sake of brevity. A tensor is a geometric entity of a tensor space that can also be interpreted as a multilinear map. A contravariant vector lives in a vector space while a covariant vector lives in the dual covector space. We recommend the textbook~\cite{tensor-Muhlich-2017} for a concise and well-explained description of tensors. 

Let $G=[g_{ij}]$ and $G^*=[{g^*}^{ij}]$ denote the $D\times D$ matrices
It follows by construction of the reciprocal basis that $G^*=G^{-1}$.
The reciprocal basis vectors ${{e^*}^i}$'s and primal basis vectors $e_i$'s can be expressed using the dual metric ${g^*}$ and metric $g$  on the primal basis vectors $e_j$'s and reciprocal basis vectors ${e^*}^j$'s, respectively:
\begin{eqnarray}
{{e^*}^i} &\eqSum& {g^*}^{ij}e_j,\\
e_i &\eqSum& g_{ij}{e^*}^j.
\end{eqnarray}

The {\em metric tensor field} $g$ (``metric tensor'' or ``metric'' for short)  defines a smooth symmetric positive-definite {\em bilinear form} on  the tangent bundle so that
for $u,v\in T_p,\quad g(u,v)\geq 0\in\bbR$. 
We can also write equivalently $g_p(u,v)\eqnota\inner{u}{v}_{p}\eqnota\inner{u}{v}_{g(p)}\eqnota\inner{u}{v}$.
Two vectors $u$ and $v$ are said orthogonal, denoted by $u\perp v$, iff $\inner{u}{v}=0$.
The  length of a vector is induced from the {\em norm} $\|u\|_p\eqnota \|u\|_{g(p)}=\sqrt{\inner{u}{u}_{g(p)}}$.
Using local coordinates of a chart $(\calU, x)$, we  get the vector contravariant/covariant components, and compute the metric tensor using matrix algebra (with column vectors by convention) as follows:
\begin{equation}
g(u,v) = (u)_B^\top \times G_{x(p)} \times(v)_B = (u)_{B^*}^\top \times G_{x(p)}^{-1} \times (v)_{B^*},
\end{equation}
since it follows from the primal/reciprocal basis that $G \times G^{*}=I$, the identity matrix.  
Thus on any tangent plane $T_p$, we get a {\em Mahalanobis distance}:
\begin{equation}
M_G(u,v)\eqdef \|u-v\|_G=\sqrt{ \sum_{i=1}^D \sum_{j=1}^D G_{ij} (u^i-v^i)(u^j-v^j) }.
\end{equation}

The inner product of two vectors $u$ and $v$ is a scalar (a $0$-tensor) that can be equivalently calculated  as:
\begin{equation}
\inner{u}{v} \eqdef g(u,v) \eqsum u^i v_i \eqsum u_i v^i.
\end{equation}

A metric tensor $g$ of manifold $M$ is said {\em conformal} when $\inner{\cdot}{\cdot}_p = \kappa(p) \inner{\cdot}{\cdot}_{\mathrm{Euclidean}}$.
That is, when the inner product is a scalar function $\kappa(\cdot)$ of the Euclidean dot product.
More precisely, we define the notion of a metric $g'$ conformal to another metric $g$ when these metrics define the same angles between vectors $u$ and $v$ of  a tangent plane $T_p$: 
\begin{equation}
\frac{g'_p(u,v)}{\sqrt{g'_p(u,u)}\sqrt{g'_p(v,v)}}=\frac{g_p(u,v)}{\sqrt{g_p(u,u)}\sqrt{g_p(v,v)}}.
\end{equation}
 Usually $g'$ is chosen as the Euclidean metric.
In conformal geometry, we can measure   angles between vectors  in  tangent planes as if we were in an Euclidean space, without any deformation.
This is handy for checking orthogonality in charts. 
For example, Poincar\'e disk model of hyperbolic geometry is conformal but Klein disk model is not conformal (except at the origin), see~\cite{HVD-2010}.

\subsection{Affine connections $\nabla$}\label{sec:affc}

An affine connection $\nabla$ is a differential operator defined on a manifold that allows us to define (1) a covariant derivative of vector fields, (2) a parallel transport of vectors on tangent planes along a smooth curve, and (3) geodesics. 
Furthermore, an affine connection fully characterizes the curvature and torsion of a manifold.

\subsubsection{Covariant derivatives $\nabla_X Y$ of vector fields}\label{sec:codev}
A connection defines a {\em covariant derivative} operator that tells us how
to differentiate a vector field $Y$ according to another vector field $X$.
The  covariant derivative operator is denoted using the traditional gradient symbol $\nabla$. 
Thus a covariate derivative $\nabla$ is a function:
\begin{equation}
\nabla:\frX(M)\times \frX(M) \rightarrow \frX(M),
\end{equation}
 that has its own special subscript notation $\nabla_X Y \eqnota \nabla(X,Y)$ for indicating that it is differentiating a vector field $Y$ according to another vector field $X$.

By prescribing $D^3$ smooth \underline{functions} $\Gamma_{ij}^k=\Gamma_{ij}^k(p)$, called the {\em Christoffel symbols of the second kind}, we define the unique
{\em affine connection} $\nabla$ that satisfies in local coordinates of chart $(\calU,x)$ the following equations:

\begin{equation}
\nabla_{\partial_i} \partial_j  =  \Gamma_{ij}^k \partial_k.
\end{equation} 

The Christoffel symbols can also be written as $\Gamma_{ij}^k := (\nabla_{\partial_i} \partial_j)^k$, where $(\cdot)^k$ denote the $k$-th coordinate.
The $k$-th component $\left(\nabla_X Y\right)^k$ of the covariant derivative of vector field $Y$ with respect to vector field $X$ is given by:
\begin{equation}
\left(\nabla_X Y\right)^k \eqSum X^i (\nabla_i Y)^k \eqSum X^i \left(\frac{\partial Y^k}{\partial x^i} + \Gamma^k_{ij} Y^j\right).
\end{equation}

The Christoffel symbols are {\it not} tensors (fields) because the transformation rules induced by a {change of basis} do not obey the tensor contravariant/covariant rules.

\begin{figure}
\begin{center}
\includegraphics[width=0.4\textwidth]{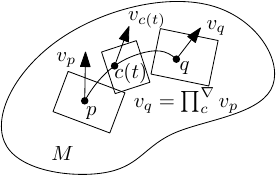}
\end{center}
\caption{Illustration of the parallel transport of vectors on tangent planes along a smooth curve. For a smooth curve $c$, with $c(0)=p$ and $c(1)=q$, a vector $v_p\in T_p$ is parallel transported smoothly to a vector $v_q\in T_q$ such that for any $t\in [0,1]$, we have $v_{c(t)}\in T_{c(t)}$.}\label{fig:pt}
\end{figure}

\subsubsection{Parallel transport $\prod^\nabla_c$ along a smooth curve $c$}\label{sec:partransp}
Since the manifold is not embedded\footnote{Whitney embedding theorem states that any $D$-dimensional Riemannian manifold can be embedded into $\bbR^{2D}$.} in a Euclidean space, we cannot add a vector $v\in T_p$ to a vector $v'\in T_{p'}$ as the tangent vector spaces are unrelated to each others without a connection.\footnote{When embedded, we can implicitly use the ambient Euclidean connection 
$\leftsup{\mathrm{Euc}}\nabla$, see~\cite{Manopt-2009}.}
Thus a {\em connection} $\nabla$ defines how to associate vectors between infinitesimally close tangent planes $T_p$ and $T_{p+\ddp}$.
Then the connection allows  us to smoothly {\em transport} a vector $v\in T_p$ by sliding it (with infinitesimal moves) along a smooth curve $c(t)$ (with $c(0)=p$ and $c(1)=q$), so that the vector
$v_p\in T_p$  ``corresponds'' to a vector $v_q\in T_{q}$: This is called the {\em parallel transport}.
This mathematical prescription is necessary in order to study dynamics on manifolds (e.g., study the motion of a particle\footnote{Elie Cartan introduced the notion of affine connections~\cite{Cartan-1986,Cartan-2011} in the 1920's motivated by the {\em principle of inertia} in mechanics: A point particle, without any force acting on it, shall move along a straight line with constant velocity.} on the manifold).
We can express the parallel transport along the smooth curve $c$ as:
\begin{equation}
\forall v\in T_p,\forall t\in [0,1],\quad v_{c(t)} = \prod_{c(0)\rightarrow c(t)}^{\nabla} v \in T_{c(t)}
\end{equation}
The parallel transport is schematically illustrated in Figure~\ref{fig:pt}.

\subsubsection{$\nabla$-geodesics $\gamma_\nabla$: Autoparallel curves}\label{sec:geo}

A connection $\nabla$ allows one to define {\em $\nabla$-geodesics} as autoparallel curves, that are curves $\gamma$ such that we have:
\begin{equation}
\nabla_{\dot\gamma} \dot\gamma=0.
\end{equation}

That is, the {\em velocity vector} $\dot\gamma$ is moving along the curve parallel to itself (and all tangent vectors on the geodesics are mutually parallel): 
In other words, $\nabla$-geodesics generalize the notion of ``straight Euclidean'' lines.
In local coordinates $(\calU,x)$, $\gamma(t)=(\gamma^k(t))_k$, the autoparallelism amounts to solve the following second-order Ordinary Differential Equations (ODEs):

\begin{equation}
\ddot\gamma(t)+\Gamma_{ij}^k \dot\gamma(t)\dot\gamma(t)=0, \quad \gamma^l(t)=x^l \circ\gamma(t),
\end{equation}
where $\Gamma_{ij}^k$ are the {\em Christoffel symbols of the second kind}, with:
\begin{equation}
\Gamma_{ij}^k  \eqEinstein \Gamma_{ij,l} g^{lk}, \quad \Gamma_{ij,k}  \eqEinstein g_{lk}\Gamma_{ij}^l,
\end{equation}
where  $\Gamma_{ij,l}$ the {\em Christoffel symbols of the first kind}.
Geodesics are 1D autoparallel submanifolds and $\nabla$-hyperplanes are defined similarly as autoparallel submanifolds of dimension $D-1$.
We may specify in subscript the connection that yields the geodesic $\gamma$: $\gamma_\nabla$.

The geodesic equation $\nabla_{\dot\gamma(t)} \dot\gamma(t)=0$ may be either solved as an {\em Initial Value Problem} (IVP) or as a {\em Boundary Value Problem} (BVP):
\begin{itemize}
\item Initial Value Problem (IVP): fix the conditions $\gamma(0)=p$ and $\dot\gamma(0)=v$ for some vector $v\in T_p$.

\item Boundary Value Problem (BVP): fix the geodesic extremities $\gamma(0)=p$ and $\gamma(1)=q$.

\end{itemize}

\subsubsection{Curvature and torsion of a manifold}\label{sec:curvaturetorsion}
An affine connection $\nabla$	 defines a 4D\footnote{It follows from symmetry constraints that the number of independent components of the Riemann tensor is $\frac{D^2(D^2-1)}{12}$ in $D$ dimensions. } {\em curvature  tensor} $R$  (expressed using components $R_{jkl}^i$ of a $(1,3)$-tensor).
The coordinate-free equation of the curvature tensor is given by:
\begin{equation}\label{eq:ct}
R(X,Y)Z \eqdef \nabla_X\nabla_Y X - \nabla_Y \nabla_X Z - \nabla_{[X,Y]} Z,
\end{equation}
where $[X,Y](f)=X(Y(f))-Y(X(f))$ ($\forall f\in\frF(M)$) is the {\em Lie bracket} of vector fields.
When the connection is the metric Levi-Civita, the curvature is called {\em Riemann-Christoffel curvature  tensor}.
In a local coordinate system, we have:
\begin{equation}
R(\partial_j,\partial_k)\partial_i \eqSum R_{jki}^l \partial_l.
\end{equation}
Informally speaking, the curvature tensor as defined in Eq.~\ref{eq:ct} quantifies the amount of non-commutativity of the covariant derivative.

A manifold $M$ equipped with a connection $\nabla$ is said {\em flat} (meaning $\nabla$-flat) when $R=0$.
This holds in particular when finding a {\em particular}\footnote{For example, the Christoffel symbols vanish in a rectangular coordinate system of a plane but not in the polar coordinate system of it.} coordinate system $x$ of a chart $(\calU,x)$ such that $\Gamma_{ij}^k=0$, i.e., when all connection coefficients vanish.
 
A manifold is {\em torsion-free} when the connection is {\em symmetric}.
A symmetric connection satisfies the following coordinate-free equation:
\begin{equation}
\nabla_XY-\nabla_YX = [X,Y].
\end{equation}
Using local chart coordinates, this amounts to check that $\Gamma_{ij}^k=\Gamma_{ji}^k$.
The torsion tensor is a $(1,2)$-tensor defined by:
\begin{equation}
T(X,Y) \eqdef \nabla_X Y-\nabla_Y X - [X,Y].
\end{equation}

For a torsion-free connection, we have the first Bianchi identity:
\begin{equation}
R(X,Y)Z+R(Z,X)Y+R(Y,Z)X=0,
\end{equation}
and the second Bianchi identity:
\begin{equation}
(\nabla_V R)(X,Y)Z+(\nabla_X R)(Y,V)Z+(\nabla_Y R)(V,X)Z=0.
\end{equation}

\begin{figure}
\begin{center}
\includegraphics[height=0.15\textheight]{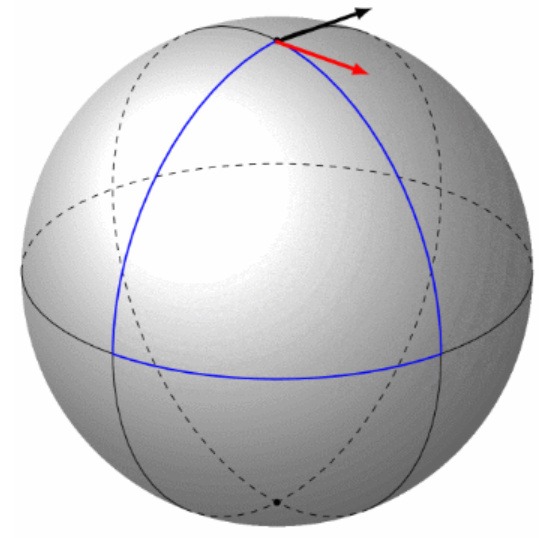}\hskip 1cm
\includegraphics[height=0.15\textheight]{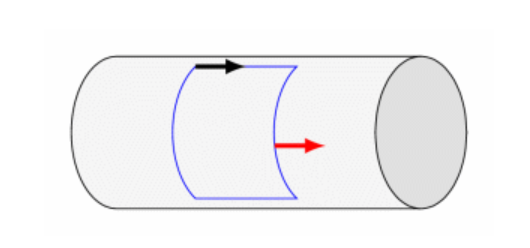}
\end{center}
\caption{
Parallel transport with respect to the metric connection: Curvature effect can be visualized as the angle defect along the parallel transport on smooth (infinitesimal) loops. For a sphere manifold, a vector parallel-transported along a loop does not coincide with itself, while it always conside with itself for a (flat) manifold.
Drawings are courtesy of \copyright{} CNRS, \protect\url{http://images.math.cnrs.fr/Visualiser-la-courbure.html}}\label{fig:sphercyl}
\end{figure}

In general, the parallel transport is {\em path-dependent}.
The {\em angle defect} of a vector transported on an {\em  infinitesimal closed loop} (a smooth curve with coinciding extremities) is related to the curvature.
However for a  {\em flat connection}, the parallel transport does not depend on the path, and yields {\em absolute parallelism geometry}~\cite{absoluteparallelism-2002}.
Figure~\ref{fig:sphercyl} illustrates the parallel transport along a curve for a curved manifold (the sphere manifold) and a flat manifold ( the cylinder manifold\footnote{The Gaussian curvature at of point of a manifold is the product of the minimal and maximal sectional curvatures: $\kappa_G\eqdef \kappa_{\min}\kappa_{\max}$ . For a cylinder, since $\kappa_{\min}=0$, it follows that the Gaussian curvature of a cylinder is $0$. Gauss's Theorema Egregium (meaning ``remarkable theorem'') proved that the Gaussian curvature is intrinsic and does not depend on how the surface is embedded into the ambient Euclidean space.
}).

\begin{figure}
\begin{center}
\includegraphics[width=\textwidth]{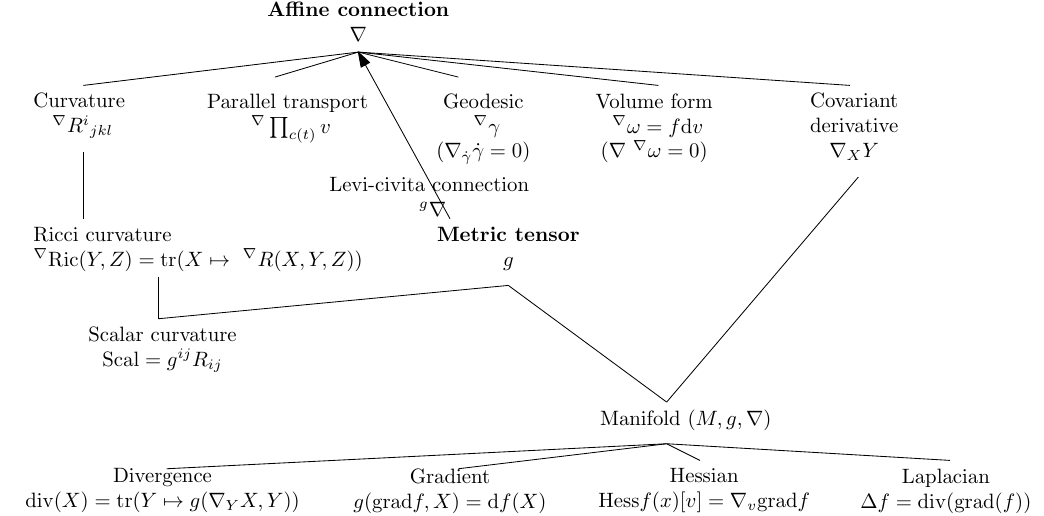}
\end{center}

\caption{Differential-geometric concepts associated to an affine connection $\nabla$ and a metric tensor $g$. 
\label{fig:dgconcepts}}

\end{figure}

An affine connection is a torsion-free linear connection.	
Figure~\ref{fig:dgconcepts}	summarizes the various concepts of differential geometry induced by an affine connection $\nabla$ and a metric tensor $g$.
	
Curvature is a fundamental concept inherent to geometry~\cite{Bourguignon-2009}:
There are several notions of curvatures: scalar curvature, sectional curvature, Gaussian curvature of surfaces to 
Riemannian-Christoffel $4$-tensor, Ricci symmetric $2$-tensor, synthetic Ricci curvature in Alexandrov geometry, etc.

\subsection{The fundamental theorem of Riemannian geometry: The Levi-Civita metric connection}\label{sec:fthmrie}
By definition, an affine connection $\nabla$ is said {\em metric compatible} with $g$ when it satisfies for any triple $(X,Y,Z)$ of vector fields the following equation:
\begin{equation}
X\inner{Y}{Z} = \inner{\nabla_X Y}{Z} +  \inner{Y}{\nabla_X Z},
\end{equation}
which can be written equivalently as:
\begin{equation}
Xg({Y},{Z}) = g({\nabla_X Y},{Z}) +  g({Y},{\nabla_X Z})
\end{equation}	
Using local coordinates and natural basis $\{\partial_i\}$ for vector fields, the metric-compatibility property amounts to check that we have: 
\begin{equation}
\partial_k g_{ij} = \inner{\nabla_{\partial_k}\partial_i}{\partial_j} +   \inner{\partial_i}{\nabla_{\partial_k}\partial_j}
\end{equation}
 
A property of using a metric-compatible connection is that the parallel transport $\prod^\nabla$ of vectors preserve the metric:
\begin{equation}
\inner{u}{v}_{c(0)}  = \Inner{\prod_{c(0)\rightarrow c(t)}^\nabla u}{\prod_{c(0)\rightarrow c(t)}^\nabla v}_{c(t)}\quad \forall t.
\end{equation}
That is, the parallel transport preserves angles (and orthogonality) and lengths of vectors in tangent planes when transported along a smooth curve.

The fundamental theorem of Riemannian geometry states the existence of a unique torsion-free metric compatible connection:
\begin{Theorem}[Levi-Civita metric connection]
There exists a unique torsion-free affine connection compatible with the metric called 
the {\em Levi-Civita connection} $\nablaLC$.
\end{Theorem}

The Christoffel symbols of the Levi-Civita connection can be expressed from the metric tensor $g$ as follows:
\begin{equation}
 {}^\LC\Gamma_{ij}^k \eqSum \frac{1}{2}  g^{kl}\left(
\partial_i g_{il} + \partial_j g_{il} -\partial_l g_{ij}
\right),
\end{equation}
	where $g^{ij}$ denote the matrix elements of the inverse matrix $g^{-1}$.
	
The Levi-Civita connection can also be defined coordinate-free with the {\em Koszul formula}:
\begin{equation}
2 g(\nabla_X Y, Z) =  X (g(Y,Z)) +  Y (g(X,Z)) -  Z (g(X,Y))  +  g([X,Y],Z) - g([X,Z],Y) - g([Y,Z],X).
\end{equation}
 
There exists metric-compatible connections with torsions studied in theoretical physics.
See for example the flat Weitzenb\"ock connection~\cite{TeleparallelGravity-2015}.

The metric tensor $g$ induces the torsion-free metric-compatible Levi-Civita connection that determines the {\em local structure} of the manifold. However, the metric $g$ does not fix the {\em global topological structure}: 
For example, although a cone and a cylinder have locally the same flat Euclidean metric, they exhibit different global structures.

\subsection{Preview: Information geometry versus Riemannian geometry}
In information geometry, we consider a pair of conjugate affine connections $\nabla$ and $\nabla^*$ (often but not necessarily torsion-free)  that are coupled to the metric $g$: The structure is conventionally written as $(M,g,\nabla,\nabla^*)$. The key property is that 
those conjugate connections are  metric compatible, and therefore the induced dual parallel transport preserves the metric:
\begin{equation}
\inner{u}{v}_{c(0)}  = \Inner{\prod_{c(0)\rightarrow c(t)}^\nabla u}{\prod_{c(0)\rightarrow c(t)}^{\nabla^*} v}_{c(t)}.
\end{equation}
Thus the Riemannian manifold $(M,g)$ can be interpreted as the self-dual information-geometric manifold obtained for $\nabla=\nabla^*=\nablaLC$ the unique torsion-free Levi-Civita metric connection: $(M,g) \equiv (M,g,\nablaLC,\nablaLC^*=\nablaLC)$.
However, let us point out that for a pair of self-dual Levi-Civita conjugate connections, the information-geometric manifold does not induce a distance.
This contrasts with the Riemannian modeling $(M,g)$ which provides a  Riemmanian metric distance $D_\rho(p,q)$ defined by the length of the geodesic $\gamma$ connecting the two points $p=\gamma(0)$ and $q=\gamma(1)$: 
\begin{eqnarray}
D_\rho(p,q) &\eqdef&\int_0^1 \|\gamma'(t)\|_{\gamma(t)} \dt =  \int_0^1 \sqrt{g_{\gamma(t)}(\dot\gamma(t),\dot\gamma(t))} \dt,\\
&=&  \int_0^1 \sqrt{{\dot\gamma}(t)^\top g_{\gamma(t)} {\dot\gamma}(t)} \dt.
\end{eqnarray}
This geodesic length distance $D_\rho(p,q)$ can also be interpreted as the shortest path linking point $p$ to point $q$: 
$D_\rho(p,q) = \inf_\gamma \int_0^1 \|\gamma'(t)\|_{\gamma(t)} \dt$ (with $p=\gamma(0)$ and $q=\gamma(1)$).

Usually, this Riemannian geodesic distance is not available in closed-form (and need to be approximated or bounded) because the geodesics cannot be explicitly parameterized (see geodesic shooting methods~\cite{GeodesicShooting-2011}).

We are now ready to introduce the key geometric structures of information geometry.

\section{Information manifolds}\label{sec:conjmfd}

\subsection{Overview} 
In this part, we explain the {\em dualistic structures} of manifolds in information geometry.
In \S\ref{sec:ccm}, we first present the core {\em Conjugate Connection Manifolds} (CCMs) $(M,g,\nabla,\nabla^*)$, and show how to build {\em Statistical Manifolds} (SMs) $(M,g,C)$ from a CCM in \S\ref{sec:statmfd}.
From any statistical manifold, we can build a $1$-parameter family  $(M,g,\nabla^{-\alpha},\nabla^\alpha)$ of CCMs, the information $\alpha$-manifolds. We state the fundamental theorem of information geometry in \S\ref{sec:fig}.
These CCMs and SMs structures are not related to any distance {\it a priori} but require at first a pair $(\nabla,\nabla^*)$ of conjugate connections
coupled to a metric tensor $g$. 
We show two methods to build an initial pair of conjugate connections.
A first method consists in building a pair of conjugate connections $(\leftsup{D}\nabla,\leftsup{D}\nabla^*)$ from any divergence $D$ in \S\ref{sec:divmfd}. 
Thus we obtain self-conjugate connections when the divergence is symmetric: $D(\theta_1:\theta_2)=D(\theta_2:\theta_1)$.
When the divergences are Bregman divergences (i.e., $D=B_F$ for a strictly convex and differentiable Bregman generator), we obtain Dually Flat Manifolds (DFMs) $(M,\nabla^2 F,\leftsup{F}\nabla,\leftsup{F}\nabla^*)$ in \S\ref{sec:dfm}.
DFMs nicely generalize the Euclidean geometry and exhibit  Pythagorean theorems.
We further characterize when orthogonal $\leftsup{F}\nabla$-projections and dual $\leftsup{F}\nabla^*$-projections of a point on submanifold a  is unique.\footnote{In Euclidean geometry, the orthogonal projection of a point $p$ onto an affine subspace $S$ is proved to be unique using the Pythagorean theorem.}
A second method to get a pair of conjugate connections $(\leftsup{e}\nabla,\leftsup{m}\nabla)$ consists in defining these connections from a regular parametric family of probability distributions $\calP=\{p_\theta(x)\}_\theta$. In that case, these `e'xponential connection $\leftsup{e}\nabla$ and `m'ixture connection $\leftsup{m}\nabla$ are coupled to the Fisher information metric $\leftsub{\calP}g$. A statistical manifold $(\calP,\leftsub{\calP}g,\leftsub{\calP}C)$ can be recovered by considering the skewness Amari-Chentsov cubic tensor
 $\leftsub{\calP}C$, and it follows a $1$-parameter family of CCMs, 
$(\calP,\leftsub{\calP}g,\leftsub{\calP}\nabla^{-\alpha},\leftsub{\calP}\nabla^{+\alpha})$, the statistical expected $\alpha$-manifolds.
In this parametric statistical context, these information manifolds are called {\em expected information manifolds} because the various quantities are expressed from statistical expectations $E_{\cdot}[\cdot]$.
Notice that these information manifolds can be used in  information sciences in general, beyond the traditional fields of statistics.
In statistics, we motivate the choice of the connections, metric tensors and divergences by studying statistical invariance criteria, 
in \S\ref{sec:criinv}.  We explain how to recover the expected $\alpha$-connections from standard $f$-divergences that are the only separable divergences that satisfy the property of information monotonicity.
Finally, in \S\ref{sec:FR}, the recall the Fisher-Rao expected Riemannian manifolds that are Riemannian manifolds $(\calP,\leftsub{\calP}g)$ equipped with a geodesic metric distance called the Fisher-Rao distance, or Rao distance for short.

\subsection{Conjugate connection manifolds: $(M,g,\nabla,\nabla^*)$}\label{sec:ccm}

We begin with a definition:

\begin{Definition}[Conjugate connections]
A connection $\nabla^*$ is said to be conjugate to a connection $\nabla$ with respect to the metric tensor $g$ if and only if  we have for any triple $(X,Y,Z)$ of smooth vector fields the following identity satisfied:
\begin{equation}
X\inner{Y}{Z} = \inner{\nabla_X Y}{Z} +  \inner{Y}{\nabla_X^* Z},\quad \forall X,Y,Z\in\frX(M). \label{eq:dc}
\end{equation}
\end{Definition}

We can notationally rewrite Eq.~\ref{eq:dc} as: 
\begin{equation}
X g({Y},{Z}) = g({\nabla_X Y},{Z}) +  g({Y},{\nabla_X^* Z}),
\end{equation}	
and further explicit that for each point $p\in M$, we have:
\begin{equation}
X_p g_p({Y}_p,{Z}_p) = g_p(({\nabla_X Y})_p,{Z}_p) +  g_p({Y}_p,({\nabla_X^* Z})_p).
\end{equation}
We check that the right-hand-side is a scalar and that the left-hand-side is a directional derivative of a real-valued function, that is also a scalar.

Conjugation is an involution: $({\nabla^*})^*=\nabla$.

\begin{Definition}[Conjugate Connection Manifold]
The structure of the Conjugate Connection Manifold (CCM) is denoted by $(M,g,\nabla,\nabla^*)$, where $(\nabla,\nabla^*)$ are conjugate connections with respect to the metric $g$.
\end{Definition}

A remarkable property is that the dual parallel transport of vectors preserves the metric.
That is, for any smooth curve $c(t)$, the inner product is conserved when we transport one of the vector $u$ using 
 the primal parallel transport $\prod^\nabla_c$ and the other vector $v$ using the dual parallel transport $\prod^{\nabla^*}_c$. 	
 
\begin{equation}
\inner{u}{v}_{c(0)}  = \Inner{\prod_{c(0)\rightarrow c(t)}^\nabla u}{\prod_{c(0)\rightarrow c(t)}^{\nabla^*} v}_{c(t)}.
\end{equation}

\begin{Property}[Dual parallel transport preserves the metric]
A pair $(\nabla,\nabla^*)$ of conjugate connections preserves the metric $g$ if and only if:
\begin{equation}
\forall t\in [0,1], \Inner{\prod_{c(0)\rightarrow c(t)}^\nabla u}{\prod_{c(0)\rightarrow c(t)}^{\nabla^*} v}_{c(t)}=\inner{u}{v}_{c(0)}.
\end{equation}
\end{Property}
	
\begin{Property}
Given a connection $\nabla$ on $(M,g)$ (i.e., a structure $(M,g,\nabla)$), there exists a unique conjugate connection $\nabla^*$
 (i.e., a dual structure $(M,g,\nabla^*)$).
\end{Property}
	
We consider a manifold $M$ equipped with a pair of conjugate connections $\nabla$ and $\nabla^*$ that are coupled with the metric tensor $g$ so that the dual parallel transport preserves the metric.	We define the mean connection $\bar\nabla$:
\begin{equation}
\bar\nabla=\frac{\nabla+\nabla^*}{2},
\end{equation} 
with corresponding Christoffel coefficients denoted by $\bar\Gamma$. 
This mean connection coincides with  the Levi-Civita metric connection:  
\begin{equation}
\bar\nabla=\nablaLC.
\end{equation} 

\begin{Property}
The mean connection $\bar\nabla$ is self-conjugate, and coincide with the Levi-Civita metric connection.
\end{Property}

\subsection{Statistical manifolds: $(M,g,C)$}\label{sec:statmfd}
Lauritzen introduced this corner structure~\cite{Lauritzen-1987} of information geometry in 1987.
Beware that although it bears the name ``statistical manifold,'' it is a purely geometric construction that may be used outside of the field
 of  Statistics.
However, as we shall mention later, we can always find a {\em statistical model} $\calP$ corresponding to a statistical manifold~\cite{smsm-2006}.
We shall see how we can convert a conjugate connection manifold into such a statistical manifold, and how we can subsequently derive  an infinite family of CCMs from a statistical manifold. 
In other words, once we have a pair of conjugate connections, we will be able to build a family of pairs of conjugate connections.

We  define a {\em totally symmetric}\footnote{This means that $C_{ijk}=C_{\sigma(i)\sigma(j)\sigma(k)}$ for any permutation $\sigma$. The metric tensor is totally symmetric.} cubic $(0,3)$-tensor (i.e., $3$-covariant tensor) called the {\em Amari-Chentsov tensor}:

\begin{equation}
C_{ijk} \eqdef \Gamma_{ij}^k-{\Gamma^{*}}_{ij}^k,
\end{equation}
or in coordinate-free equation:
\begin{equation}
C(X,Y,Z) \eqdef  \inner{\nabla_X Y-\nabla^*_X Y}{Z}.
\end{equation}

Using the local basis, this cubic tensor can be expressed as:
\begin{equation}
C_{ijk} = C(\partial_i,\partial_j,\partial_k)=\inner{\nabla_{\partial_i} {\partial_j}-\nabla^*_{\partial_i} {\partial_j}}{{\partial_k}}
\end{equation}

\begin{Definition}[Statistical manifold~\cite{Lauritzen-1987}]
A statistical manifold $(M,g,C)$ is a manifold $M$ equipped with a metric tensor $g$ and a totally symmetric cubic tensor $C$.
\end{Definition}


\subsection{A family $\{(M,g,\nabla^{-\alpha},\nabla^{\alpha}=(\nabla^{-\alpha})^*)\}_{\alpha\in\bbR}$ of conjugate connection manifolds}\label{sec:famccm}\label{sec:alphamfd}
For any pair $(\nabla,\nabla^*)$ of conjugate connections, we can define a {\em $1$-parameter family of connections} $\{\nabla^{\alpha}\}_{\alpha\in\bbR}$, called the {\em $\alpha$-connections} such that $(\nabla^{-\alpha},\nabla^{\alpha})$ are dually coupled to the metric, with $\nabla^{0}=\bar\nabla=\nablaLC$, $\nabla^{1}=\nabla$ and $\nabla^{-1}=\nabla^*$. 
By observing that the scaled cubic tensor $\alpha C$ is also a totally symmetric cubic $3$-covariant tensor, 
we can derive the $\alpha$-connections from a statistical manifold $(M,g,C)$ as:

\begin{eqnarray}
\Gamma^{\alpha}_{ij,k} &=& \Gamma^{0}_{ij,k} - \frac{\alpha}{2} C_{ij,k},\\
\Gamma^{-\alpha}_{ij,k} &=& \Gamma^{0}_{ij,k} + \frac{\alpha}{2} C_{ij,k},
\end{eqnarray}
where $\Gamma^{0}_{ij,k}$ are the Levi-Civita Christoffel symbols, and $\Gamma_{ki,j}\eqSum \Gamma_{ij}^l g_{lk}$ (by index juggling).

The $\alpha$-connection $\nabla^\alpha$ can also be defined as follows:
\begin{equation}
g(\nabla^\alpha_X Y,Z)=g(\nablaLC_X Y,Z)+\frac{\alpha}{2}C(X,Y,Z), \forall X,Y,Z\in\frX(M).
\end{equation}

\begin{Theorem}[Family of information $\alpha$-manifolds]
For any $\alpha\in\bbR$, $(M,g,\nabla^{-\alpha},\nabla^{\alpha}=(\nabla^{-\alpha})^*)$ is a conjugate connection manifold.
\end{Theorem}	

The $\alpha$-connections $\nabla^\alpha$ can also be constructed directly from a pair $(\nabla,\nabla^*)$ of conjugate connections by taking the following weighted combination:
\begin{equation}
\Gamma^{\alpha}_{ij,k}=  \frac{1+\alpha}{2} \Gamma_{ij,k} + \frac{1-\alpha}{2} \Gamma^*_{ij,k}.
\end{equation}

\subsection{The fundamental theorem of information geometry: $\nabla$ $\kappa$-curved $\Leftrightarrow$ $\nabla^*$ $\kappa$-curved}\label{sec:fig} 

We now state the fundamental theorem of information geometry and its corollaries:

\begin{Theorem}[Dually constant curvature manifolds]
If a torsion-free affine connection $\nabla$ has constant curvature $\kappa$ then its conjugate torsion-free connection $\nabla^*$ 
has necessarily the {\em same} constant curvature $\kappa$.
\end{Theorem}
The proof is reported in~\cite{IG-2014} (Proposition 8.1.4, page 226).	

A statistical manifold $(M,g,C)$ is said {\em $\alpha$-flat} if its induced $\alpha$-connection is flat.
It can be shown that $R^{\alpha}=-R^{-\alpha}$.

We get the following two corollaries:
	
\begin{Corollary}[Dually $\alpha$-flat manifolds]
A manifold $(M,g,\nabla^{-\alpha},\nabla^{\alpha})$ is $\nabla^{\alpha}$-flat if and only if it is  $\nabla^{-\alpha}$-flat.
\end{Corollary}

\begin{Corollary}[Dually flat manifolds  ($\alpha=\pm 1$)]
A manifold $(M,g,\nabla,\nabla^*)$ is $\nabla$-flat if and only if it is  $\nabla^*$-flat. 
 \end{Corollary}
(See Theorem 3.3 of~\cite{IG-2000})

Let us now define the notion of constant curvature of a statistical structure~\cite{Furuhata-2009}:
\begin{Definition}[Constant curvature $\kappa$]
A statistical structure $(M,g,\nabla)$ is said of constant curvature $\kappa$ when
$$
R^{\nabla}(X, Y) Z=\kappa\{g(Y, Z) X-g(X, Z) Y\}, \quad \forall X, Y, Z \in \Gamma(TM),
$$
where $\Gamma(TM)$ denote the space of smooth vector fields.
\end{Definition}

It can be proved that the Riemann-Christoffel  (RC) $4$-tensors of conjugate $\alpha$-connections~\cite{IG-2014} are related as follows:
\begin{equation}
g\left(R^{(\alpha)}(X, Y) Z, W\right)+g\left(Z, R^{(-\alpha)}(X, Y) W\right)=0.
\end{equation}
Thus we have $g\left(R^{\nabla^*}(X, Y) Z, W\right)=-g\left(Z, R^{\nabla}(X, Y) W\right)$.

Thus once we are given a pair of conjugate connections, we can always build a $1$-parametric family of manifolds.
Manifolds with constant curvature $\kappa$ are interesting from the computational viewpoint as dual geodesics have simple closed-form expressions.

\subsection{Conjugate connections from divergences: $(M,D)\equiv (M,\leftsup{D}g,\leftsup{D}\nabla,\leftsup{D}\nabla^*=\leftsup{D^*}\nabla)$}\label{sec:divmfd}

Loosely speaking, a divergence $D(\cdot:\cdot)$ is a smooth distance~\cite{DivGeo-Zhang-2014}, potentially asymmetric.
In order to define precisely a divergence, let us first introduce the following handy notations: $\partial_{i,\cdot}f(x,y)=\frac{\partial}{\partial x^i} f(x,y)$,  $\partial_{\cdot,j}f(x,y)=\frac{\partial}{\partial y^j} f(x,y)$, $\partial_{ij,k}f(x,y)=\frac{\partial^2}{\partial x^i\partial x^j}\frac{\partial}{\partial y^k} f(x,y)$ and
$\partial_{i,jk}f(x,y)=\frac{\partial}{\partial x^i}  \frac{\partial^2}{\partial y^j\partial y^k}f(x,y)$, etc.
	 
\begin{Definition}[Divergence]	
A {\em divergence} $D: M\times M \rightarrow [0,\infty)$ on a manifold $M$ with respect to a local chart $\Theta\subset \bbR^D$ is 
a $C^3$-function satisfying the following properties:
\begin{enumerate}
	\item $D(\theta:\theta')\geq 0$ for all $\theta,\theta'\in\Theta$ with equality holding iff $\theta=\theta'$ (law of the indiscernibles),
	
	\item $\partial_{i,\cdot} D(\theta:\theta')|_{\theta=\theta'}=\partial_{\cdot,j} D(\theta:\theta')|_{\theta=\theta'}=0$ for all $i,j\in [D]$,
	
	\item $-\partial_{\cdot,i}\partial_{\cdot,j} D(\theta:\theta')|_{\theta=\theta'}$ is positive-definite.
\end{enumerate}
\end{Definition}

The {\em dual   divergence}  is defined by swapping the arguments:
\begin{equation}
D^*(\theta:\theta') \eqdef D(\theta':\theta),
\end{equation}
and is also called the {\em reverse divergence} (reference duality in information geometry). 
Reference duality of divergences is an involution: $({D^*})^*=D$.

The Euclidean distance is a metric distance but not a divergence. The squared Euclidean distance is a non-metric symmetric divergence.
The metric tensor $g$ yields Riemannian metric distance $D_\rho$ but it is never a divergence.

From any given divergence $D$, we can define a conjugate connection manifold following the construction of Eguchi~\cite{Eguchi-1983,Eguchi-1985} (1983):

\begin{Theorem}[Manifold from divergence]
$(M,\gD,\nablaD,\nablaDstar)$ is an information manifold with:
	\begin{eqnarray}
	\gD &\eqdef & -\partial_{i,j} D(\theta:\theta')|_{\theta=\theta'} = \gDstar,\\ 
	\GammaD_{ijk}&\eqdef& -\partial_{ij,k} D(\theta:\theta')|_{\theta=\theta'},\\
  \GammaDstar_{ijk}&\eqdef& -\partial_{k,ij} D(\theta:\theta')|_{\theta=\theta'}.
\end{eqnarray}
\end{Theorem}
 
The associated statistical manifold is $(M,\gD,\CD)$ with:
\begin{equation}
\CD_{ijk}= \GammaDstar_{ijk}-\GammaD_{ijk}.
\end{equation}

 Since $\alpha\CD$ is a totally symmetric cubic tensor for any $\alpha\in\bbR$, we can derive a one-parameter family of conjugate connection manifolds: 
\begin{equation}
\left\{(M,\gD,\CD^\alpha) \equiv (M,\gD,\nablaD^{-\alpha},(\nablaD^{-\alpha})^*= \nablaD^{\alpha})\right\}_{\alpha\in\bbR}.
\end{equation}
 
In the remainder, we use the shortcut $(M,D)$ to denote the divergence-induced information  manifold $(M,\gD,\nablaD,\nablaD^*)$.
Notice that it follows from construction that:
\begin{equation}
\nablaD^*=\leftsup{D^*}\nabla.
\end{equation}

\subsection{Dually flat manifolds (Bregman geometry):  $(M,F)\equiv (M,\leftsup{B_F}g,\leftsup{B_F}\nabla,\leftsup{B_F}\nabla^*=\leftsup{B_{F^*}}\nabla)$}\label{sec:dfm}

We consider dually flat manifolds that satisfy asymmetric Pythagorean theorems.
These flat manifolds can be obtained from a canonical Bregman divergence.

Consider a {\em strictly convex smooth function} $F(\theta)$ called a {\em potential function}, with $\theta\in\Theta$ where $\Theta$ is an open convex domain. Notice that the function convexity does not change by an affine transformation.
We associate to the potential function $F$ a corresponding {\em Bregman  divergence} (parameter divergence):
\begin{equation}
B_F(\theta:\theta') \eqdef F(\theta)-F(\theta')-(\theta-\theta')^\top \nabla F(\theta').
\end{equation}

We write also the Bregman divergence between point $P$ and point $Q$ as $D(P:Q) \eqdef B_F(\theta(P):\theta(Q))$, where $\theta(P)$ denotes the coordinates of a point $P$.

The information-geometric structure induced by a Bregman generator\footnote{Here, we define a Bregman generator as a proper, lower semi-continuous, and strictly convex and $C^3$ differentiable real-valued function.} is $(M,\gF,\CF) \eqdef (M,\idxSup{B_F}g,\idxSup{B_F}C)$ with:
	
	\begin{eqnarray}
	\leftsup{F}g &\eqdef& \leftsup{B_F}g= - \left[{\partial_i}{\partial_j} B_F(\theta:\theta')|_{\theta'=\theta}\right] = \nabla^2 F(\theta), \\
	\leftsup{F}\Gamma &\eqdef& \leftsup{B_F}\Gamma_{ij,k}(\theta)=0,\label{eq:gammaf}\\
	\leftsup{F}C_{ijk} &\eqdef& \leftsup{B_F}C_{ijk}= {\partial_i}{\partial_j}{\partial_k} F(\theta).
	\end{eqnarray}

Since all coefficients of the Christoffel symbols vanish (Eq.~\ref{eq:gammaf}), the information manifold is {\em $\leftsup{F}\nabla$-flat}.
The Levi-Civita connection $\nablaLC$ is obtained from the metric tensor $\leftsup{F}g$ (usually not flat), and  
 we get the conjugate connection $(\leftsup{F}\nabla)^*=\leftsup{F}\nabla^{1}$ from $(M,\leftsup{F}g,\leftsup{F}C)$.

The Legendre-Fenchel transformation yields the {\em convex conjugate} $F^*$ that is interpreted as the {\em dual potential function}: 
\begin{equation}
F^*(\eta) \eqdef \sup_{\theta\in\Theta} \{\theta^\top\eta-F(\theta)\}.
\end{equation}

\begin{Theorem}[Fenchel-Moreau biconjugation~\cite{ConvexAnalysis-2012}]
If $F$  is a lower semicontinuous\footnote{A function $f$ is lower semicontinous (lsc) at $x_0$ iff $f(x_0)\leq \lim_{x\rightarrow x_0} \inf f(x)$. A function $f$ is lsc if it is lsc at $x$ for all $x$ in the function domain.  } and convex function, then its Legendre-Fenchel
transformation is involutive: $({F^*})^*=F$ (biconjugation).
\end{Theorem}

In a dually flat manifold, there exists two global dual affine coordinate systems $\eta=\nabla F(\theta)$ and $\theta=\nabla F^*(\eta)$,
 and therefore the manifold can be covered by a single chart.
Thus if a probability family belongs to an exponential family then its natural parameters cannot belong to, say, a spherical space (that requires at least two charts).
	
We have the Crouzeix~\cite{Crouzeix-1977} identity relating the Hessians of the potential functions: 
\begin{equation}
\nabla^2 F(\theta) \nabla^2 F^*(\eta)=I,
\end{equation}
where $I$ denote the $D\times D$ identity matrix.
This Crouzeix identity reveals that $B=\{\partial_i\}_i$ and $B^*=\{\partial^j\}_j$ are the primal and reciprocal basis, respectively.

The Bregman divergence can be reinterpreted using Young-Fenchel (in)equality as the {\em canonical divergence} $A_{F,F^*}$~\cite{canonicalDiv-2015}:
\begin{equation}
	B_F(\theta:\theta')=A_{F,F^*}(\theta:\eta')=F(\theta)+F^*(\eta')-\theta^\top\eta' =A_{F^*,F}(\eta':\theta).  
\end{equation}

The {\em dual Bregman divergence} ${B_F}^*(\theta:\theta')\eqdef B_F(\theta':\theta)=B_{F^*}(\eta:\eta')$ yields
		\begin{eqnarray}
	\leftsup{F}g^{ij}(\eta) &=& \partial^i\partial^j F^*(\eta),\quad \partial^l\eqNota \frac{\partial}{\partial\eta^l}\\
	{\leftsup{F}\Gamma^*}^{ijk}(\eta) &=& 0, \quad \leftsup{F}C^{ijk} =  \partial^i\partial^j\partial^k F^*(\eta)
	\end{eqnarray}
	
Thus the information manifold is both ${\leftsup{F}\nabla}$-flat and ${\leftsup{F}\nabla^*}$-flat: 
This structure is called a {\em dually flat manifold} (DFM).
In a DFM, we have two global affine coordinate systems $\theta(\cdot)$ and $\eta(\cdot)$ related by the Legendre-Fenchel transformation of a pair of potential functions $F$ and $F^*$. That is, $(M,F)\equiv (M,F^*)$, 
and the dual atlases are $\calA=\{(M,\theta)\}$ and $\calA^*=\{(M,\eta)\}$.

In a dually flat manifold,  any pair of points $P$ and $Q$ can  either be linked using the $\nabla$-geodesic (that is $\theta$-straight) or the $\nabla^*$-geodesic (that is $\eta$-straight).
In general, there are $2^3=8$ types of {\em geodesic triangles} in a dually flat manifold.

on a Bregman manifold, the primal parallel transport of a vector does not change the contravariant vector components, and the dual parallel transport does not change the covariant vector components. 
Because the dual connections are flat, the dual parallel transports are path-independent.

\begin{figure}
\begin{center}
\includegraphics[width=0.85\textwidth]{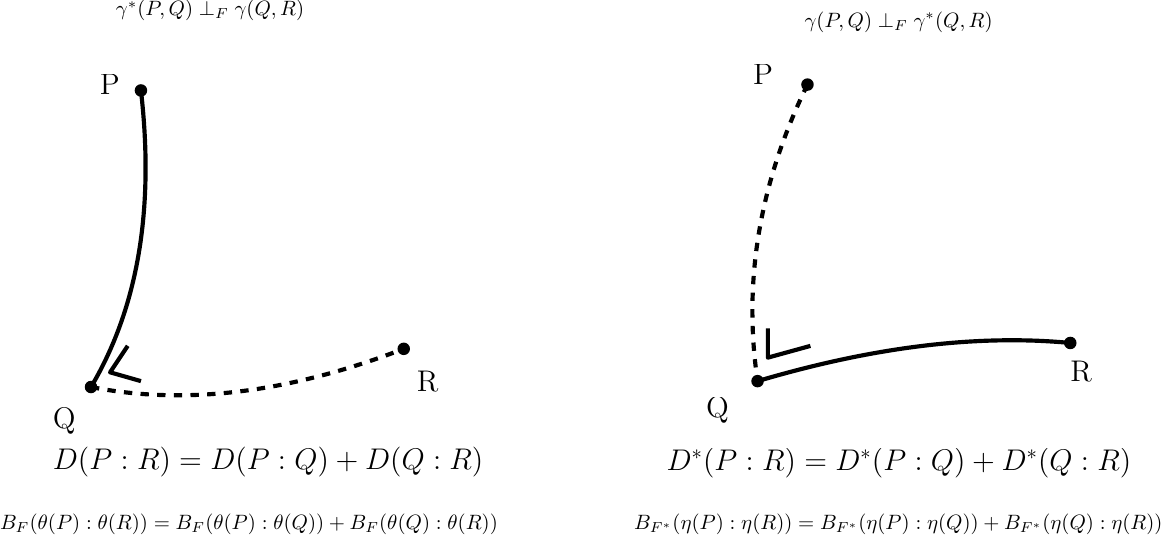}
\end{center}
\caption{Dual Pythagorean theorems in a dually flat space.\label{fig:Pytha}}
\end{figure}

Moreover,   the dual Pythagorean theorems~\cite{WhatIsInfoProj-2018} illustrated in Figure~\ref{fig:Pytha} holds.
Let $\gamma(P,Q)=\gamma_\nabla(P,Q)$ denote the $\nabla$-geodesic passing through points $P$ and $Q$, and 
$\gamma^*(P,Q)=\gamma_{\nabla^*}(P,Q)$ denote the $\nabla^*$-geodesic passing through points $P$ and $Q$.
Two curves $\gamma_1$ and $\gamma_2$ are orthogonal at point $p=\gamma_1(t_1)=\gamma_2(t_2)$ with respect to the metric tensor $g$ when 
$g(\dot\gamma_1(t_1),\dot\gamma_2(t_2))=0$.

\begin{Theorem}[Dual Pythagorean identities]
\begin{eqnarray*}
\gamma^*(P,Q)\perp \gamma(Q,R) &\Leftrightarrow& (\eta(P)-\eta(Q))^\top (\theta(Q)-\theta(R))\eqSum (\eta_i(P)-\eta_i(Q))(\theta_i(Q)-\theta_i(R))  =0,\\
\gamma(P,Q)\perp \gamma^*(Q,R) &\Leftrightarrow& (\theta(P)-\theta(Q))^\top (\eta(Q)-\eta(R))\eqSum (\theta_i(P)-\theta_i(Q))^\top (\eta_i(Q)-\eta_i(R))=0.
\end{eqnarray*}
\end{Theorem}

We can define dual Bregman projections and characterize when these projections are unique:
  A {\em submanifold} $S\subset M$  is said $\nabla$-flat ($\nabla^*$-flat) iff. it corresponds to an {\em affine subspace} in the $\theta$-coordinate system  (in the $\eta$-coordinate system, respectively).
 
\begin{Theorem}[Uniqueness of projections]
 The {\em $\nabla$-projection} $P_S$ of $P$ on $S$ is {\em unique} if $S$ is {\em $\nabla^*$-flat} and minimizes the 
	divergence $D(\theta(P):\theta(Q))$:
\begin{equation}
\mbox{$\nabla$-projection:} \quad  P_S=\arg\min_{Q\in S} D(\theta(P):{\theta(Q)}).
\end{equation}
 
The dual {$\nabla^*$-projection} $P_S^*$ is {\em unique} 
if $M\subseteq S$ is {\em $\nabla$-flat} and minimizes the  divergence $D(\theta(Q):\theta(P))$:

\begin{equation}
\mbox{$\nabla^*$-projection:} \quad {P_S^*=\arg\min_{Q\in S} D({\theta(Q)}:\theta(P))}.
\end{equation}
 \end{Theorem}

Let $S\subset M$ and $S'\subset M$, then we define the divergence between $S$ and $S'$ as
\begin{equation}
D(S:S') \eqdef \min_{s\in S,s'\in S'} D(s:s').
\end{equation}

When $S$ is a $\nabla$-flat submanifold and $S'$ $\nabla^*$-flat submanifold, the divergence $D(S:S')$ between submanifold $S$ and submanifold $S'$ can be calculated using the method of alternating projections~\cite{IG-2016}.
Let us remark that Kurose~\cite{alphaPythagoras-1994} reported a  Pythagorean theorem for dually constant curvature manifolds that generalizes the  Pythagorean theorems of dually flat spaces.

\def\Ball{\mathrm{Ball}}

We shall concisely explain the {\em space of Bregman spheres} explained in details in~\cite{BVD-2010}. 
Let $D$ denote the dimension of $\Theta$. 
We define the lifting of primal coordinates $\theta$  to the primal potential function $\calF=\{\hat{\theta}=(\theta,\theta_{D+1}=F(\theta))\ :\ \theta\in\Theta\}$  using an extra dimension $\theta_{D+1}$.
A Bregman ball $\Sigma$
\begin{eqnarray}
\Ball_F(C:r) := \{ P \st    F(\theta(P))+F^*(\eta(C))-\inner{\theta(P)}{\eta(C)} \leq  r \}
\end{eqnarray}
can then be lifted to $\calF$: $\hat{\Sigma}=\{\hat{\theta}(P)\ :\ P\in\sigma\}$.
The boundary Bregman sphere $\sigma=\partial\Sigma$ is lifted to $\partial\hat{\Sigma}=\hat{\sigma}$, and the lifted points are all supported by a supporting $(D+1)$-dimensional hyperplane (of dimension $D$): 
\begin{equation}
H_{\hat{\sigma}}: \theta_{D+1}=\inner{\theta-\theta(C)}{\eta(C)}+F(\theta(C))+r.
\end{equation}
Let $H_{\hat{\sigma}}^-$  denotes the halfspaces bounded by $H_{\hat{\sigma}}$ and containing $\hat\theta(C)=(\theta(C),F(\theta(C)))$.
A point $P$ belongs to a Bregman ball $\Sigma$ iff $\hat{\theta(P)}\in  H_{\hat{\sigma}}^-$, see~\cite{BVD-2010}.
Reciprocally, a $(D+1)$-dimensional hyperplane $H: \theta_{D+1}=\inner{\theta}{\eta_a}+b$ cutting the potential function $\calF$ yields a 
Bregman sphere 
$\sigma_H$ of center $C$ with $\theta(C)=\nabla F^*(\eta_a)$
and radius
$r=\inner{\nabla F^*(\eta_a)}{\eta_a}-F(\theta_a)+b=F^*(\eta_a)+b$, where $\theta_a=\nabla F^*(\eta_a)$.
It follows that the intersection of $k$ Bregman balls is a $(D-k)$-dimensional Bregman ball, and that a Bregman sphere can be defined by $D+1$ points in general position since an hyperplane in the augmented space is defined by $D+1$ points. 
We can test whether a point $P$ belongs to a Bregman ball with bounding Bregman sphere passing through $D+1$ points $P_1, \ldots, P_{D+1}$ or not by checking the sign of a $(D+2)\times (D+2)$ determinant:
\begin{equation}
\mathrm{InBregmanBall}_F(P_1,\ldots,P_{d+1};P):=\mathrm{sign}\left(\left| 
\begin{array}{llll}
1 & \ldots & 1 & 1\\
\theta(P_1) &\ldots & \theta(P_{D+1}) & \theta(P)\\
F(\theta(P_1)) & \ldots & F(\theta(P_{D+1})) & F(\theta(P))
\end{array}
\right|\right).
\end{equation}
We have:
\begin{equation}
\mathrm{InBregmanBall}_F(P_1,\ldots,P_{d+1};P):\left\{
\begin{array}{lll}
=-1 & \Leftrightarrow & P\in\mathrm{InBregmanBall}_F^\circ(P_1,\ldots,P_{D+1};P)\\
=0 & \Leftrightarrow & P\in\partial\mathrm{InBregmanBall}_F(P_1,\ldots,P_{D+1};P)\\
=+1 & \Leftrightarrow & P\not\in\mathrm{InBregmanBall}_F(P_1,\ldots,P_{D+1};P)
\end{array}
\right.
\end{equation}

Similarly, a dual-type Bregman ball $\Sigma^*$ can be defined by
\begin{eqnarray}
\Ball^*_F(C:r) := \{ P \st    F(\theta(C))+F^*(\eta(P))-\inner{\theta(C)}{\eta(P)} \leq  r \},
\end{eqnarray}
and be lifted to the dual potential function $\calF^*$.
Notice that $\Ball^*_F(C:r)=\Ball_{F^*}(C:r)$.

In general, we have the following quadrilateral relation for Bregman divergences:  

\begin{Property}[Bregman $4$-parameter property~\cite{Bregman4pt-1997}]
For any four points $P_1$, $P_2$, $Q_1$, $Q_2$, we have the following identity:
\begin{eqnarray}
B_{F}(\theta(P_1) : \theta(Q_1)) +  B_{F}(\theta(P_2) : \theta(Q_2)) 
 && - B_{F}(\theta(P_1) : \theta(Q_2)) - B_{F}(\theta(P_2) : \theta(Q_1))  \nonumber\\
&& -(\theta(P_2)-\theta(P_1))^\top (\eta(Q_1)-\eta(Q_2))=0.
\end{eqnarray}
\end{Property}

\begin{figure}
\begin{center}
\includegraphics[width=0.85\textwidth]{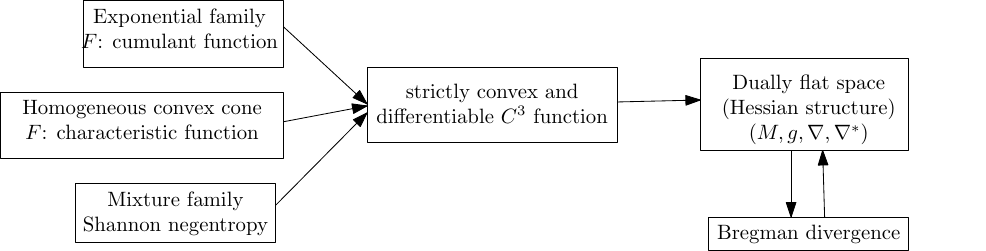}
\end{center}

\caption{Common dually flat spaces associated to smooth and strictly convex generators.
 \label{fig:cdfsF}
}

\end{figure}

In summary, to define a dually flat space, we need a convex Bregman generator. When the $\alpha$-geometries are neither dually flat (eg., Cauchy manifolds~\cite{VorCauchy-2020}, we may still  build a dually flat structure on the manifold by considering some Bregman generator (eg., Bregman-Tsallis generator for the dually flat Cauchy manifold~\cite{VorCauchy-2020}). 
The dually flat geometry can be investigated under the wider scope of {\em Hessian manifolds}~\cite{Shima-2007} which consider {\em locally} potential functions.
In general, a dually flat space can be built from any smooth strictly convex generator $F$.
For example, a dually flat geometry can be built on homogeneous cones with the characteristic function $F$ of the cone~\cite{Shima-2007}.
Figure~\ref{fig:cdfsF} illustrates several common constructions of dually flat spaces.

\subsection{Hessian $\alpha$-geometry: $(M,F,\alpha)\equiv (M,\leftsup{F}g,\leftsup{F}\nabla^{-\alpha},\leftsup{F}\nabla^\alpha$}
\label{sec:alphahessian}

The dually flat manifold is also called a manifold with a {\em Hessian structure}~\cite{Shima-2007} induced by a convex potential function $F$. Since we built two dual affine connections $\leftsup{B_F}\nabla=\leftsup{F}\nabla$ and 
$\leftsup{B_F}\nabla^*=\leftsup{F}\nabla^*=\leftsup{F^*}\nabla$, we can build a family of $\alpha$-geometry as follows:

\begin{equation}
\leftsup{F}g_{ij}(\theta)= \partial_i\partial_j F(\theta),\quad \leftsup{F}g^{ij}(\eta)=\partial^i\partial^j F(\eta),
\end{equation}
and
\begin{equation}
\leftsup{F}\Gamma_{ijk}^\alpha(\theta)=\frac{1-\alpha}{2} \partial_i\partial_j\partial_k F(\theta),
\quad  \leftsup{F}{\Gamma_{ijk}^\alpha}^*(\eta)=\leftsup{F^*}{\Gamma_{ijk}^\alpha}(\eta)=\frac{1+\alpha}{2} \partial^i\partial^j\partial^k F^*(\eta).
\end{equation}

Thus when $\alpha=\pm1$, the Hessian $\alpha$-geometry is dually flat.

We now consider information manifolds induced by parametric statistical models.

\subsection{Expected $\alpha$-manifolds of a family of parametric probability distributions: $(\calP,\leftsub{\calP}g,\leftsub{\calP}\nabla^{-\alpha},\leftsub{\calP}\nabla^{\alpha})$}\label{sec:expalphamfd}\label{sec:invexpalphamfd}

Informally speaking, an {\em expected manifold} is an information manifold built on a regular parametric family of distributions.
It is sometimes called ``expected'' manifold or ``expected'' geometry in the literature~\cite{RefRepDuality-2015} because the components of the metric tensor $g$ and the Amari-Chentsov cubic tensor $C$  are expressed using statistical expectations $E_{\cdot}[\cdot]$.

Let $\calP$ be a {\em parametric family} of probability distributions:
\begin{equation}
\calP \eqdef \left\{p_\theta(x)\right\}_{\theta\in\Theta},
\end{equation}
with $\theta$ belonging to the open parameter space $\Theta$. 
The order of the family is the dimension of its parameter space.
We define the
likelihood function\footnote{The likelihood function is an {\em equivalence class} of functions defined modulo a positive scaling factor.} $L(\theta;x) \eqdef p_\theta(x)$ as a function of $\theta$, and its corresponding {\em log-likelihood} function: 
\begin{equation}
l(\theta;x) \eqdef \log L(\theta;x)=\log p_\theta(x).
\end{equation}
The {\em score vector}:
\begin{equation}
s_\theta=\nabla_\theta l=(\partial_i l)_i,
\end{equation} 
indicates the sensitivity of the likelihood $\partial_i l \eqnota \frac{\partial}{\partial\theta_i} l(\theta;x)$.
	
The {\em Fisher information matrix} (FIM) of $D\times D$  for $\dim(\Theta)=D$ is defined by:
\begin{equation}
\leftsub{\calP}I(\theta) \eqdef E_\theta\left[ \partial_i l\partial_j l\right]_{ij}  \succeq 0,
\end{equation}
where $\succeq$ denotes the L\"owner order. 
That is, for two symmetric positive-definite matrices $A$ and $B$, $A \succeq B$ if and only if matrix $A-B$ is positive semidefinite.
For regular models~\cite{IG-2014}, the FIM is positive definite: $\leftsub{\calP}I(\theta)\succ 0$, where $A \succ  B$ if and only if matrix $A-B$ is positive-definite.

The FIM  is invariant by reparameterization of the sample space $\calX$, and covariant by reparameterization of the parameter space $\Theta$, see~\cite{IG-2014}.
That is, let $\bar{p}(x;\eta)=p(\theta(\eta);x)$.
Then we have:
\begin{equation}\label{eq:FIMcovariant}
\bar{I}(\eta) = \left[\frac{\partial \theta_i}{\partial \eta_j} \right]_{ij}^\top  I(\theta(\eta)) \left[\frac{\partial \theta_i}{\partial \eta_j} \right]_{ij}.
\end{equation}
Matrix $J_{ij}=\left[\frac{\partial \theta_i}{\partial \eta_j} \right]_{ij}$ is the Jacobian matrix.

\begin{Example}
For example, consider the family 
\begin{equation}
\mathcal{N}=\left\{p(x;\mu,\sigma)=\frac{1}{\sqrt{2\pi}\sigma} \exp(-\frac{(x-\mu)^2}{2\sigma^2})) \ :\ (\mu,\sigma)\in\bbR\times\bbR_{++}\right\}
\end{equation}
of univariate normal distributions. The 2D parameter vector is $\lambda=(\mu,\sigma)$ with $\mu$ denoting the mean and $\sigma$ the standard deviation. Another common parameterization of the normal family is $\lambda'=(\mu,\sigma^2)$. 
The $\lambda'$ parameterization extends naturally to $d$-variance normal distributions with $\lambda'=(\mu,\Sigma)$, where  $\Sigma$ denotes the covariance matrix (with $\Sigma=\sigma^2$ when $d=1$). For multivariate normal distributions, the $\lambda$-parameterization can be interpreted as $\lambda=(\mu,L^\top)$ where $L^\top$ is the upper triangular matrix in the Cholesky decomposition (when $d=1$, $L^\top=\sigma$).
We have the following Fisher information matrices in the $\lambda$-parameterization and  $\lambda'$-parameterization:
\begin{equation}
I_{\lambda}(\lambda)=\left[\begin{array}{cc}
\frac{1}{\lambda_{2}^{2}} & 0 \\
0 & \frac{2}{\lambda_{2}^{2}}
\end{array}\right]=\left[\begin{array}{cc}
\frac{1}{\sigma^{2}} & 0 \\
0 & \frac{2}{\sigma^{2}}
\end{array}\right]
\end{equation}
and
\begin{equation}
I_{\lambda^{\prime}}\left(\lambda^{\prime}\right)=\left[\begin{array}{cc}
\frac{1}{\lambda_{2}} & 0 \\
0 & \frac{1}{2 \lambda_{2}^{2}}
\end{array}\right]=\left[\begin{array}{cc}
\frac{1}{\sigma^{2}} & 0 \\
0 & \frac{1}{2 \sigma^{4}}
\end{array}\right]
\end{equation}

Since the FIM is covariant, we have the following the change of transformation:
\begin{equation}
I_{\lambda^{\prime}}\left(\lambda^{\prime}\right)=J_{\lambda, \lambda^{\prime}}^{\top} I_{\lambda}\left(\lambda\left(\lambda^{\prime}\right)\right) J_{\lambda, \lambda^{\prime}},
\end{equation}
with
\begin{equation}
J_{\lambda^{\prime}, \lambda}=\left[\begin{array}{cc}
1 & 0 \\
0 & 2 \sigma
\end{array}\right]
\end{equation}

Thus we check that
\begin{equation}
I_{\lambda}(\lambda)=\left[\begin{array}{cc}
1 & 0 \\
0 & 2 \sigma
\end{array}\right]\left[\begin{array}{cc}
\frac{1}{\sigma^{2}} & 0 \\
0 & \frac{1}{2 \sigma^{4}}
\end{array}\right]\left[\begin{array}{cc}
1 & 0 \\
0 & 2 \sigma
\end{array}\right]=\left[\begin{array}{cc}
\frac{1}{\sigma^{2}} & 0 \\
0 & \frac{2}{\sigma^{2}}
\end{array}\right]
\end{equation}

Notice that the infinitesimal length elements are invariant: $\ds_\lambda=\ds_{\lambda'}$.
\end{Example}

As a corollary, notice that we can recognize the Euclidean metric in any other coordinate system if the metric tensor $g$ can be written
$J_{\lambda, \lambda^{\prime}}^{\top}  J_{\lambda, \lambda^{\prime}}$. 
For example, the Riemannian geometry induced by a dually flat space with a separable potential function is Euclidean~\cite{RieSepBregman-2019}.

	In statistics, the FIM plays a role in the attainable precision of unbiased estimators.
For any  unbiased estimator, the Cram\'er-Rao lower bound~\cite{CRLB-IG-2013} on the variance  of the estimator is:
\begin{equation}
\mathrm{Var}_\theta[\hat{\theta}_n(X)]\succeq \frac{1}{n} \leftsub{\calP}I^{-1}(\theta).
\end{equation}
	
\begin{figure}
\begin{center}
\includegraphics[width=0.75\textwidth]{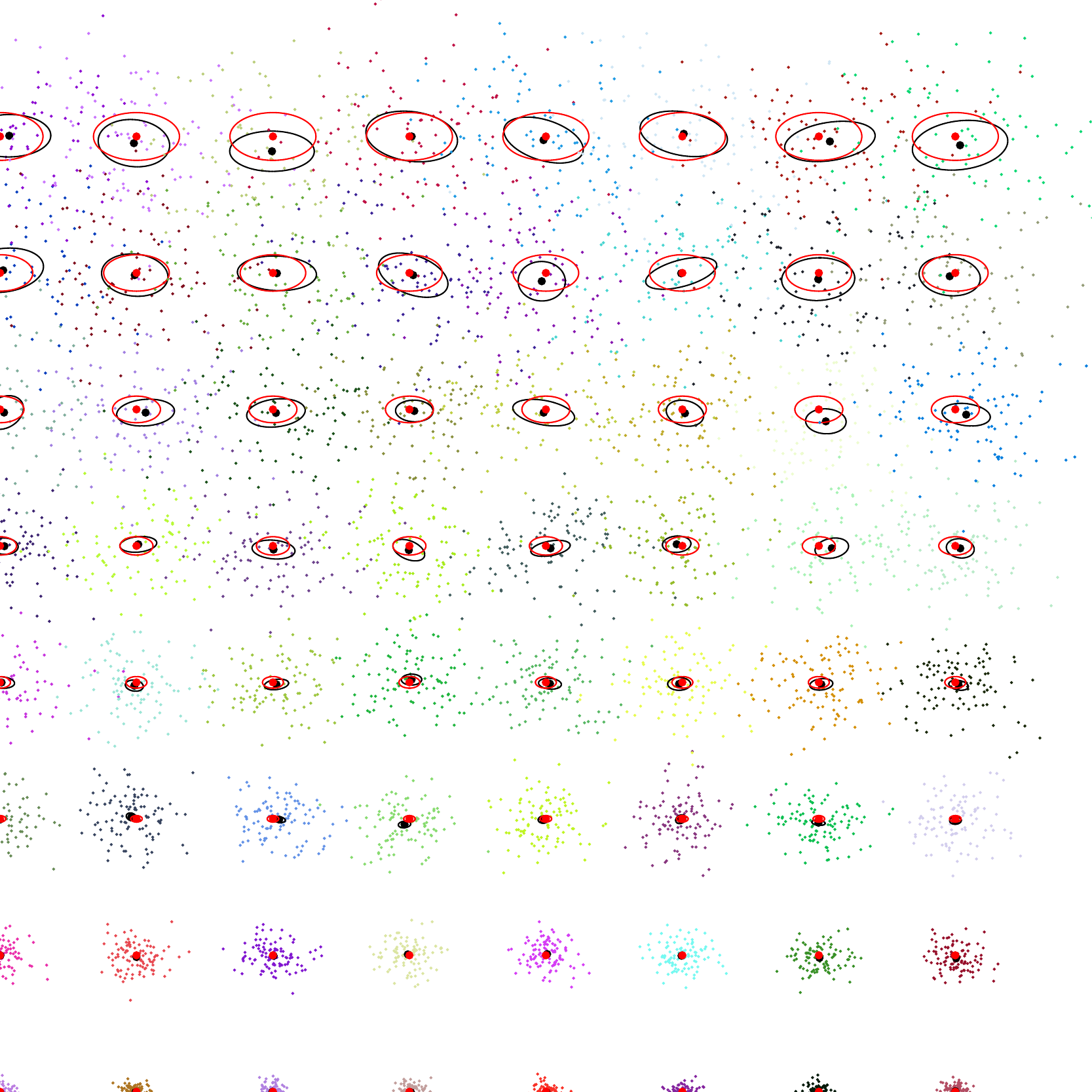}
\end{center}

\caption{Visualizing the Cram\'er-Rao lower bound: The red ellipses display the Fisher information matrix of normal distributions at grid locations. The black ellipses are sample covariance matrices centered at the sample means calculated by repeating $200$ runs of sampling $100$ iid variates for the normal parameters of the grid.
\label{fig:crlb}}

\end{figure}

Figure~\ref{fig:crlb}	illustrates the Cram\'er-Rao lower bound (CRLB) for the univariate distributions:
At regular grid locations $(\mu,\sigma)$ of the upper space of normal parameters, we repeat $200$ runs (trials) of estimating the normal parameters $\widehat{(\mu,\sigma)}$ using the MLE on $100$ iid samples $x_1,\ldots, x_n\sim N(\mu,\sigma)$. The sample mean and the sample covariance matrix are 
calculated for the number of trials and displayed as back ellipses. The Fisher information matrix is plotted as red ellipses at the grid locations: The red ellipses have semi-axis parallel to the coordinate system since the parameters $\mu$ and $\sigma$ are orthogonal (diagonal FIM). This is not true anymore for the sample covariance matrix of the MLE estimator, and the centers of the sample covariance matrices deviate from the grid locations.
	
We report the expression of the FIM for two important generic parametric family of probability distributions: 
(1) an exponential family (with its prominent multivariate normal family), and (2) a mixture family.

\begin{Example}[FIM of an exponential family $\calE$]
An {\em exponential family}~\cite{EF-2009} $\calE$ is defined for a sufficient statistic vector $t(x)=(t_1(x),\ldots, t_D(x))$, and an auxiliary carrier measure $k(x)$ by the following canonical density:
\begin{equation}\label{eq:efcd}
\calE=\left\{ p_\theta(x) = \exp\left(\sum_{i=1}^D t_i(x)\theta_i -F(\theta) + k(x) \right) \st \theta\in\Theta \right\},
\end{equation}
where $F$ is the strictly convex cumulant function (also called log-normalizer, and log partition function or free energy in statistical mechanics).
Exponential families include the  Gaussian family, the Gamma and Beta families, the probability simplex $\Delta$, etc.
The FIM of an exponential family is given by:
\begin{equation}
\leftsub{\calE}I(\theta) = \Cov_{X\sim p_\theta(x)}[t(x)] = \nabla^2 F(\theta)= (\nabla^2 F^*(\eta))^{-1} \succ 0.
\end{equation}	
Natural parameters beyond vector types can also be used in the canonical decomposition of the density of an exponential family:
For example, we may use a matrix type for defining the zero-centered multivariate Gaussian family or the Wishart family, a complex numbers for defining the complex-valued Gaussian distribution  family, etc. 
We then replace the term $\sum_{i=1}^D t_i(x)\theta_i$ in Eq.~\ref{eq:efcd} by an inner product defined for the natural parameter type (e.g., dot product for vectors, matrix product trace for matrices, etc). Furthermore, natural parameters can be of compound types:
For example, the multivariate Gaussian distribution can be written using $\theta=(\theta_v,\theta_M)$ where $\theta_v$ is a vector part and $\theta_M$ a matrix part, see~\cite{EF-2009}.

Let $\Sigma=[\sigma_{ij}]$ denote the {\em covariance matrix} and $\Sigma^{-1}=[\sigma^{ij}]$ the {\em precision matrix} of a multivariate normal distribution.
The Fisher information matrix   of the multivariate Gaussian~\cite{FisherRaoBivarNormal-1979,Skovgaard-1984} $N(\mu,\Sigma)$ is given  by
\begin{equation}
I(\mu,\Sigma) =
\begin{blockarray}{c|cc}
\mu & \Sigma=[\sigma_{ij}] \\
\begin{block}{[cc]c}
  \sigma^{ij} & 0 & \mu \\ 
  0 & \sigma^{il}\sigma^{jk}+\sigma^{ik}\sigma^{jl} &\Sigma=[\sigma_{kl}] \\
\end{block}
\end{blockarray}
\end{equation}
Notice that the lower right block matrix is a 4D  tensor of dimension $d\times d\times d\times d$.
The zero subblock matrices in the FIM indicate that the parameters $\mu$ and $\Sigma$ are orthogonal to each other.
In particular, when $d=1$, since $\sigma^{11}=\frac{1}{\sigma^2}$, we recover the Fisher information matrix of the univariate Gaussian:
\begin{equation}
I(\mu,\Sigma) =
\left[\begin{array}{cc}
\frac{1}{\sigma^{2}} & 0 \\
0 & \frac{1}{2 \sigma^{4}}
\end{array}\right]
\end{equation}
We refer to~\cite{Malago-2015} for the FIM of a Gaussian distribution using other canonical parameterizations (natural/expectation parameters of exponential family).
\end{Example}
	
\begin{Example}[FIM of a mixture family $\calM$]
A mixture family is defined for $D+1$ functions $F_1,\ldots, F_D$ and $C$ as:
\begin{equation}
\calM = \left\{ p_\theta(x) = \sum_{i=1}^D \theta_i F_i(x) + C(x) \st \theta\in\Theta \right\},
\end{equation}
where the functions $\{F_i(x)\}_i$ are linearly independent on the common support $\calX$ and satisfying 
$\int F_i(x) \dmu(x)=0$. Function $C$ is such that $\int C(x)\dmu(x)=1$.
Mixture families include statistical mixtures with prescribed component distributions and the probability simplex $\Delta$.
The FIM of a mixture family is given by:	
\begin{equation}
\leftsub{\calM}I(\theta) = E_{X\sim p_\theta(x)}\left[ \frac{F_i(x)F_j(x)}{(p_\theta(x))^2}\right]  = \int_\calX \frac{F_i(x)F_j(x)}{p_\theta(x)} \dmu(x) \succ 0.
\end{equation}

The family of Gaussian mixture model (GMM) with prescribed component distributions (ie., convex weight combinations of $D+1$ Gaussian densities) form a mixture family~\cite{geowmixtures-2018}. 
	\end{Example}
	
Notice that the probability simplex of discrete distributions can be {\em both} modeled as an exponential family or a mixture family~\cite{IG-2016}.	
 
The {\em expected $\alpha$-geometry} is built from the {\em expected dual $\pm\alpha$-connections}.
The Fisher ``{\em information metric}'' tensor is built from the FIM as follows:
\begin{equation}
	\leftsub{\calP}g(u,v) \eqdef (u)_\theta^\top\, \leftsub{\calP}I(\theta)\,  (v)_\theta
\end{equation}
	
The  expected {\em exponential connection}  and expected {\em mixture connection} are given by
\begin{eqnarray}
	\leftsubsup{\calP}{e}\nabla  &\eqdef& E_\theta\left[(\partial_i\partial_j l)(\partial_k l)\right],\\
	\leftsubsup{\calP}{m}\nabla   &\eqdef& E_\theta\left[(\partial_i\partial_j l+\partial_i l\partial_j l)(\partial_k l)\right].
\end{eqnarray}
	
The dualistic structure is denoted by $(\calP, \leftsub{\calP}g, \leftsubsup{\calP}{m}\nabla,\leftsubsup{\calP}{e}\nabla)$   with
	Amari-Chentsov cubic tensor called the {\em skewness tensor}: 
	\begin{equation}
	C_{ijk} \eqdef E_\theta\left[\partial_i l\partial_j l \partial_k l\right].
	\end{equation}
	
  It follows that we can build a one-family of expected information $\alpha$-manifolds: 
	\begin{equation}
	\left\{(\calP, \leftsub{\calP}g, \leftsub{\calP}\nabla^{-\alpha},\leftsub{\calP}\nabla^{+\alpha})\right\}_{\alpha\in\bbR},
	\end{equation}
	with
	\begin{eqnarray}
	\leftsub{\calP}{\Gamma^{\alpha}}_{ij,k}(\theta)  &\eqdef& E_\theta\left[\partial_i\partial_j l\partial_k l\right] + \frac{1-\alpha}{2}C_{ijk}(\theta),\\
	&=& E_\theta\left[\left(\partial_i\partial_j l+\frac{1-\alpha}{2}\partial_i l\partial_j l\right)(\partial_k l)\right].
	\end{eqnarray}
	
The Levi-Civita metric connection is recovered as follows:	
	
	\begin{equation}
	\leftsub{\calP}\bar\nabla = \frac{\leftsub{\calP}\nabla^{-\alpha}+\leftsub{\calP}\nabla^{\alpha}}{2}= \leftsubsup{\calP}{{\LC}}\nabla 
	\eqdef \leftsup{\LC}\nabla(\leftsub{\calP}g)
	\end{equation}
	
	The $\alpha$-Riemann-Christoffel curvature tensor is:
		\begin{equation}
	\leftsub{\calP}R_{ijkl} = \partial_i\Gamma_{jk,l}^\alpha-\partial_j\Gamma_{ik,l}^\alpha
	+g^{rs}\left(\Gamma_{ik,r}^\alpha\Gamma_{js,l}^\alpha - \Gamma_{jk,r}^\alpha\Gamma_{is,l}^\alpha\right),
	\end{equation}
	with $R_{ijkl}^\alpha=-R_{ijlk}^{-\alpha}$.
	We check that the expected $\pm\alpha$-connections are coupled with the metric: 
	$\partial_i g_{jk} = \Gamma_{ij,k}^\alpha +\Gamma_{ik,j}^{-\alpha}$.
	
In case of an exponential family $\calE$ or a mixture family $\calM$ equipped with the dual exponential/mixture connection, we get {\em dually flat manifolds}  
(Bregman geometry).

Indeed, for the exponential/mixture families, it is easy to check that the Christoffel symbols of $\nabla^e$ and  $\nabla^m$ vanish:
	\begin{equation}
	\leftsubsup{\calM}{e}\Gamma = \leftsubsup{\calM}{m}\Gamma = \leftsubsup{\calE}{e}\Gamma =\leftsubsup{\calE}{m}\Gamma =0.
	\end{equation}
	 
\subsection{Criteria for statistical invariance}\label{sec:criinv}
So far we have explained how to build an information manifold (or information $\alpha$-manifold) from a pair of conjugate connections.
Then we reported two ways to obtain such a pair of conjugate connections: (1) from a parametric divergence, or (2) by using the predefined expected exponential/mixture connections. 
We now ask the following question: Which information manifold makes sense in Statistics?
We can refine the question as follows:

\begin{itemize}
	\item Which metric tensors $g$ make sense in statistics?
	\item Which affine connections $\nabla$ make sense in statistics?
  \item Which statistical divergences make sense in statistics (from which we can get the metric tensor and dual connections)?
\end{itemize}
 
By definition, an {\em invariant metric tensor} $g$ shall preserve the inner product under important {\em statistical mappings} called Markov embeddings. Informally, we embed $\Delta_D$ into $\Delta_{D'}$ with $D'>D$ and the induced metric should be preserved (see~\cite{IG-2016}, page 62).
	                              
\begin{Theorem}[Uniqueness of Fisher information metric~\cite{Campbell-1986,uniqueFisher-2017}]
The Fisher information metric is the unique invariant metric tensor under Markov embeddings up  to a scaling constant.
\end{Theorem}

\begin{figure}
	\begin{center}
\includegraphics[width=0.4\textwidth]{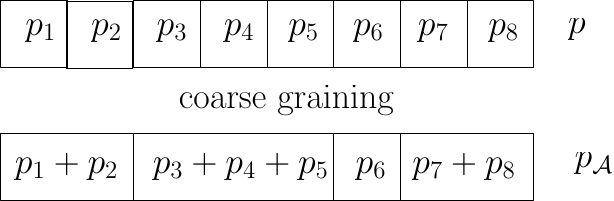}
\end{center}
	\caption{A divergence satisfies the property of information monotonicity iff $D(\theta_{\bar \calA}:\theta'_{\bar \calA})\leq D(\theta:\theta')$. Here, parameter $\theta$ represents a discrete distribution.}\label{fig:lumping}
	\end{figure}

A  $D$-dimensional parameter (discrete) divergence satisfies the {\em information monotonicity}\footnote{This property could be renamed as  the ``distance coarse-binning inequality property.''} if and only if: 
\begin{equation}
D(\theta_{\bar \calA}:\theta'_{\bar \calA})\leq D(\theta:\theta')
\end{equation}
	for any {\em coarse-grained partition} $\calA=\{\calA_i\}_{i=1}^{E}$ of $[D]=\{1,\ldots, D\}$ ($\calA$-lumping~\cite{TutCsiszar-2004}) with $E\leq D$, where 
	$\theta_{\bar \calA}^i=\sum_{j\in\calA_i } \theta^j$ for $i\in [E]$.
	This concept of coarse-graining is illustrated in Figure~\ref{fig:lumping}.
	
	A {\em separable divergence} $D(\theta_1:\theta_2)$ is a divergence that can be expressed as the sum of elementary {\em scalar divergences} $d(x:y)$:
\begin{equation}
D(\theta_1:\theta_2) \eqdef \sum_i d(\theta_1^i:\theta_2^j).
\end{equation}
For example, the squared Euclidean distance $D(\theta_1:\theta_2)=\sum_i (\theta_1^i-\theta_2^i)^2$ is a separable divergence for the scalar Euclidean divergence $d(x:y)=(x-y)^2$. 
The Euclidean distance $D_E(\theta_1,\theta_2)=\sqrt{\sum_i (\theta_1^i-\theta_2^i)^2}$  is {\em not} separable because of the square root operation. 
	
 The only invariant and {\em decomposable} divergences when $D>1$ are $f$-divergences~\cite{Jiao-2014} defined for a convex functional generator $f$:
	\begin{equation}
I_f(\theta:\theta') \eqdef \sum_{i=1}^D \theta_i f\left( \frac{\theta_i'}{\theta_i} \right)\geq f(1), \quad f(1)=0
	\end{equation}
	
The {\em standard $f$-divergences} are defined for $f$-generators satisfying $f'(1)=0$ 
	(choose $f_\lambda(u)\eqdef f(u)+\lambda (u-1)$ since $I_{f_\lambda}=I_f$), and  $f''(u)=1$ (scale fixed).

    Statistical $f$-divergences are {\em invariant}~\cite{Qiao-2010} under one-to-one/sufficient statistic transformations $y=t(x)$ of sample space:
		$p(x;\theta)=q(y(x);\theta)$:

\begin{eqnarray*}
	I_f[p(x;\theta):p(x;\theta')] &=&
	\int_\calX p(x;\theta) f\left( \frac{p(x;\theta')}{p(x;\theta)}\right)  \dmu(x), \\
	&=& \int_\calY q(y;\theta) f\left( \frac{q(y;\theta')}{q(y;\theta)}\right)  \dmu(y), \\
	&= & I_f[q(y;\theta):q(y;\theta')].
	\end{eqnarray*}

The dual $f$-divergences for reference duality is
\begin{equation}
	{{I_f}^*}[p(x;\theta):p(x;\theta')] = I_f[p(x;\theta'):p(x;\theta)] = 
	{I_{f^\diamond}}[p(x;\theta):p(x;\theta')]
\end{equation}
	for the standard conjugate $f$-generator (diamond $f^\diamond$ generator) with: 
\begin{equation}
	f^\diamond(u) \eqdef  uf\left(\frac{1}{u}\right).
\end{equation}
One can check that $f^\diamond$ is a standard $f$-generator when $f$ is standard.  

Let us report some common examples of $f$-divergences:

\begin{itemize}
	\item The family of {\em $\alpha$-divergences}: 
	\begin{equation}
	I_\alpha[p:q] \eqdef \frac{4}{1-\alpha^2} \left(1-\int p^{\frac{1-\alpha}{2}}(x) q^{\frac{1+\alpha}{2}}(x) \dmu(x)\right),
	\end{equation} 
	obtained for $f(u)=\frac{4}{1-\alpha^2}(1-u^{\frac{1+\alpha}{2}})$. The $\alpha$-divergences include:
	
	\begin{itemize}
	\item the {\em Kullback-Leibler} when $\alpha\rightarrow 1$:
	\begin{equation}
	\KL[p:q]=\int p(x)\log \frac{p(x)}{q(x)} \dmu(x),
	\end{equation}
	for $f(u)=-\log u$.
	
  \item the {\em reverse Kullback-Leibler} $\alpha\rightarrow -1$:
		\begin{equation} 
		\KL^*[p:q]=\int q(x)\log \frac{q(x)}{p(x)} \dmu(x) = \KL[q:p],
		\end{equation}
		for $f(u)=u\log u$.
	
	\item the symmetric squared {\em Hellinger divergence}: 
	\begin{equation}
	H^2[p:q]=\int (\sqrt{p(x)}-\sqrt{q(x)})^2 \dmu(x),
	\end{equation}
	for $f(u)=(\sqrt{u}-1)^2$ (corresponding to $\alpha=0$)
	
	\item the Pearson and Neyman chi-squared divergences~\cite{chisquarefdiv-2013}, etc.
\end{itemize}

\item the {\em Jensen-Shannon divergence}: 
\begin{equation}
\mathrm{JS}[p:q] = \frac{1}{2}\int \left(p(x)\log \frac{2p(x)}{p(x)+q(x)} +  q(x)\log \frac{2q(x)}{p(x)+q(x)}\right)\dmu(x),
\end{equation}
for $f(u)=  -(u+1)\log \frac{1+u}{2} + u\log u$.

\item the {\em Total Variation}  
\begin{equation}
\mathrm{TV}[p:q]=\frac{1}{2}\int |p(x)-q(x)| \dmu(x),
\end{equation}
 for $f(u)=\frac{1}{2} |u-1|$. 
The total variation distance is the only metric $f$-divergence.
\end{itemize}

The $f$-topology is the topology generated by open $f$-balls, open balls with respect to  $f$-divergences.
A topology $T$ is said stronger than a topology $T'$  if $T$ contains all the open sets of $T'$.
Csiszar's theorem~\cite{Csiszar-1967} states that when $|\alpha|<1$, the $\alpha$-topology is equivalent to the  topology induced by the total variation metric distance. Otherwise, the $\alpha$-topology  is stronger than the TV topology.

Let us state an important feature of $f$ divergences:
\begin{Theorem}
The $f$-divergences are invariant by diffeomorphisms $m(x)$ of the sample space $\calX$:
Let $Y=m(X)$, and $X_i~\sim p_i$ with $Y_i=m(X_i)\sim q_i$.
Then we have $I_f[q_1:q_2]=I_f[p_1:p_2]$.
\end{Theorem}

\begin{Example}
Consider the exponential distributions and the Rayleigh distributions which are related by:
$$
X \sim\mathrm{Exponential}(\lambda) \Leftrightarrow Y=m(X)=\sqrt{X} \sim \mathrm{Rayleigh}\left(\sigma=\frac{1}{\sqrt{2 \lambda}}\right).
$$
The densities of the exponential distributions are defined by
$$
p_{\lambda}(x)=\lambda \exp (-\lambda x) \text { with support } \mathcal{X}=[0, \infty),
$$
and the densities of the Rayleigh distributions are defined by
$$
q_{\sigma}(x)=\frac{x}{\sigma^{2}} \exp \left(-\frac{x^{2}}{2 \sigma^{2}}\right) \text { with support } \mathcal{X}=[0, \infty).
$$
We have
$$
D_{\mathrm{KL}}\left[q_{\sigma_{1}}: q_{\sigma_{2}}\right]=\log \left(\frac{\lambda_{2}^{2}}{\lambda_{1}^{2}}\right)+\frac{\sigma_{1}^{2}-\sigma_{2}^{2}}{\sigma_{2}^{2}}.
$$
It follows that
\begin{eqnarray*}
D_{\mathrm{KL}}\left[p_{\lambda_{1}}: p_{\lambda_{2}}\right] &=&D_{\mathrm{KL}}\left[q \frac{1}{\sqrt{2 \lambda_{1}}}: q \frac{1}{\sqrt{2 \lambda_{1}}}\right] \\
&=&\log \frac{2 \lambda_{1}}{2 \lambda_{2}}+2 \lambda_{2}\left(\frac{1}{2 \lambda_{1}}-\frac{1}{\lambda_{2}}\right) \\
&=&\log \left(\frac{\lambda_{1}}{\lambda_{2}}\right)+\frac{\lambda_{2}}{\lambda_{1}}-1.
\end{eqnarray*}
\end{Example}

A remarkable property is that invariant standard $f$-divergences yield the  Fisher information matrix and the $\alpha$-connections.
Indeed, the invariant   standard $f$-divergences is related infinitesimally to the Fisher metric as follows: 
	
	\begin{eqnarray}
	I_f[p(x;\theta):p(x;\theta+\dtheta)] &=&
	\int p(x;\theta) f\left( \frac{p(x;\theta+\dtheta)}{p(x;\theta)}\right)  \dmu(x)\\
	&\eqSum& \frac{1}{2} {{}_{F}g_{ij}(\theta)} \dtheta^i\dtheta^j
	\end{eqnarray}
	
 A	{\em statistical parameter divergence} $D$ on a parametric family of distributions $\calP$ yields an equivalent {\em parameter divergence}
 $\leftsub{\calP}D$:
\begin{equation}
\leftsub{\calP}D(\theta:\theta')  \eqdef D[p(x;\theta):p(x;\theta')] .
\end{equation}
Thus we can build the information manifold induced by this parameter divergence $\leftsub{\calP}D(\cdot:\cdot)$.
 For $\leftsub{\calP}D(\cdot:\cdot)=I_f[\cdot:\cdot]$, the induced $\pm 1$-divergence connections $\leftsubsup{\calP}{I_f}\nabla\eqdef \leftsup{\leftsub{\calP}{I_f}}\nabla$  and $\leftsubsup{\calP}{(I_f)^*}\nabla \eqdef \leftsup{\leftsub{\calP}{I_f^*}}\nabla$ are precisely
the {\em expected $\pm\alpha$-connections} (derived from the exponential/mixture connections) with: 
\begin{equation}
\alpha=2f'''(1)+3.
\end{equation}

Thus the invariant connections which coincide with the connections induced by the invariant statistical divergences 
are the expected $\alpha$-connections. Note that the curvature of an expected $\alpha$-connection depends both on $\alpha$ and on the considered statistical model~\cite{locationscaleMfdMitchell-1988}.

\subsection{Fisher-Rao expected Riemannian manifolds: $(\calP,\leftsub{\calP}g)$}\label{sec:FR}

Historically, a first manifold modeling of a regular parametric family of distributions $\calP=\{p_\theta(x)\}_\theta$
was to consider the {\em Fisher Information Matrix} (FIM) as  the Riemannian metric tensor $g$ (see~\cite{Hotelling-1930,Rao-1945}), with:
 
\begin{equation}
 \leftsub{\calP}I(\theta) \eqdef E_{p_\theta}\left[ \partial_i l \partial_j l \right],
\end{equation}
where $\partial_i l\eqnota \frac{\partial}{\partial\theta_i} \log p(x;\theta)$.
Under some regularity conditions, we can rewrite the FIM:
\begin{equation}
 \leftsub{\calP}I(\theta) \eqdef - E_{p_\theta}\left[ \partial_i\partial_j l \right].
\end{equation}

The Riemannian geodesic metric distance $D_\rho$ is commonly called the {\em Fisher-Rao distance}:
\begin{equation}
D_\rho(p_{\theta_1},p_{\theta_2}) =  \int_0^1 \sqrt{{\dot\gamma}(t)^\top g_{\gamma(t)} {\dot\gamma}(t)} \dt,
\end{equation}
where $\gamma$ denotes the geodesic passing through $\gamma(0)=\theta_1$ and  $\gamma(1)=\theta_2$.
The Fisher-Rao distance can also be defined as the shortest path length: $D_\rho(p_{\theta_1},p_{\theta_2}) =  \inf_\gamma \int_0^1 \sqrt{{\dot\gamma}(t)^\top g_{\gamma(t)} {\dot\gamma}(t)} \dt$.

\begin{Definition}[Fisher-Rao distance]
The Fisher-Rao distance is the geodesic metric distance of the Fisher-Riemannian manifold $(\calP,\leftsub{\calP}g)$.
\end{Definition}

Let us give some examples of Fisher-Riemannian manifolds:
\begin{itemize}
\item The Fisher-Riemannian manifold of the family of categorical distributions (also called finite discrete distributions in~\cite{IG-2016}) amount to the spherical geometry~\cite{KassVos-1997} (spherical manifold). 

\item  The Fisher-Riemannian manifold of the family of bivariate location-scale families amount to hyperbolic geometry (hyperbolic manifold).

\item  The Fisher-Riemannian manifold of the family of location  families amount to Euclidean geometry (Euclidean manifold).
\end{itemize}

The first fundamental form of the Riemannian geometry is $\ds^2=\inner{\dx}{\dx}\eqSum g_{ij}\dx^i\dx^j$ where $\ds$ denotes the line element.

This Riemannian geometric structure applied to a family of parametric probability distributions was first proposed  by Harold Hotelling~\cite{Hotelling-1930} (in a handwritten note of 1929, reprinted typeset in~\cite{Stigler-2007}) and independently later by C. R. Rao~\cite{Rao-1945} (1945, reprinted in~\cite{Rao-reproduced-1992}).
In a similar vein, Jeffreys~\cite{Jeffreys-1946} proposed to use the volume  element of a manifold as an invariant prior: The eponym Jeffreys prior in 1946.

Notice that for a parametric family of probability distributions $\calP$, the Riemannian structure $(\calP,\leftsub{\calP}g)$ coincides with the self-dual conjugate connection manifold $(\calP,\leftsub{\calP}g,\leftsubsup{\calP}{I_f}\nabla,\leftsubsup{\calP}{I_f}\nabla^*)$ induced by a {\em symmetric} $f$-divergence like the squared Hellinger divergence.

The exponential map $\exp_p$ at point $p\in M$ provides a way to map back a vector $v\in T_p$ to a point $\exp_p(v)\in M$ (when well-defined).
The exponential map can be used to parameterize a geodesic $\gamma$ with $\gamma(0)=p$ and unit tangent vector $\dot\gamma(0)=v$: $t\mapsto \exp_p(tv)$. For geodesically complete manifolds, the exponential map is defined everywhere.

\subsection{The monotone $\alpha$-embeddings and the metric gauge freedom}\label{eq:alphaembeddings}

Another common mathematically equivalent expression of the FIM~\cite{IG-2014} is given by:

\begin{equation}\label{eq:FIM0}
I_{ij}(\theta) \eqdef 4\int \partial_i\sqrt{p(x;\theta)}\partial_j\sqrt{p(x;\theta)}\dmu(x).
\end{equation}
This form of the FIM is well-suited to prove that the FIM is always a positive semi-definite matrix~\cite{IG-2014} ($I(\theta)\succeq 0$).
It turns out that we can define a family of {\em equivalent representations} of the FIM using the {\em $\alpha$-embedding}~\cite{embeddingZhang-2015} of the parametric family.

First, we define the {\em $\alpha$-representation} of densities $l^{\alpha}(x;\theta):=k_\alpha(p(x;\theta))$ with:

\begin{equation}\label{eq:alphaembedding}
k_\alpha(u) \eqdef \left\{
\begin{array}{ll}
\frac{2}{1-\alpha}u^{\frac{1-\alpha}{2}}, & \mbox{if $\alpha\not=1$,}\\
\log u, & \mbox{if $\alpha=1$}.
\end{array}
\right.
\end{equation}

The function $l^{\alpha}(x;\theta)$ is called the {\em $\alpha$-likelihood} function.
Then the $\alpha$-representation of the FIM, the $\alpha$-FIM for short, is expressed as:
\begin{equation}\label{eq:FIMalpha}
I_{ij}^{\alpha}(\theta) \eqdef \int \partial_i l^{\alpha}(x;\theta)\partial_j l^{-\alpha}(x;\theta) \dmu(x).
\end{equation}

We can rewrite compactly the $\alpha$-FIM, as $I_{ij}^{\alpha}(\theta)=\int  \partial_i l^{\alpha}\partial_j l^{-\alpha} \dmu(x)$.
Expanding the $\alpha$-FIM, we get:
\begin{equation}
I_{ij}^{\alpha}(\theta) = \left\{
\begin{array}{ll}
\frac{1}{1-\alpha^2}\int \partial_i p(x;\theta)^{\frac{1-\alpha}{2}}  \partial_j p(x;\theta)^{\frac{1+\alpha}{2}} \dmu(x) & \mbox{for $\alpha\not =\pm1$}\\
\int \partial_i \log p(x;\theta) \partial_j p(x;\theta) \dmu(x) & \mbox{for $\alpha\in \{-1,1\}$}\\
\end{array}
\right.
\end{equation}

The $1$-representation of the density is called the {\em logarithmic representation} (or $e$-representation), 
the $-1$-representation the {\em mixture representation} (or $m$-representation), and its  $0$-representation 
is called the {\em square root representation}.
The set of {\em $\alpha$-scores} vectors  $B_\alpha\eqdef \{\partial_i l^{\alpha}\}_i$  are interpreted as the tangent basis vectors of the $\alpha$-base $B_\alpha$. Thus the FIM is $\alpha$-independent.
 
Furthermore,  the $\alpha$-representation of the FIM can   be rewritten under mild conditions~\cite{IG-2014}  as:
\begin{equation}
I_{ij}^{\alpha}(\theta)=  -\frac{2}{1+\alpha} \int p(x;\theta)^{\frac{1+\alpha}{2}} \partial_i\partial_j l^{\alpha}(x;\theta)\dmu(x).
\end{equation}

Since we have:
\begin{equation}
 \partial_i\partial_j l^{\alpha}(x;\theta) = p^{\frac{1-\alpha}{2}}\left(
\partial_i\partial_j l + \frac{1-\alpha}{2} \partial_i l\partial_j l
 \right),
\end{equation}
it follows that:
\begin{equation}
I_{ij}^{\alpha}(\theta)= -\frac{2}{1+\alpha} \left(-I_{ij}(\theta)+ \frac{1-\alpha}{2} I_{ij}\right)=I_{ij}(\theta).
\end{equation}

Notice that when $\alpha=1$, we recover the equivalent expression of the FIM (under mild conditions):
\begin{equation}
I_{ij}^{1}(\theta)= -E[\nabla^2 \log p(x;\theta)].
\end{equation}
In particular, when the family is an exponential family~\cite{EF-2009} with cumulant function $F(\theta)$ (satisfying the mild conditions), we have:
\begin{equation}
I(\theta)=\nabla^2 F(\theta).
\end{equation}

Zhang~\cite{embeddingZhang-2015,NaudtsZhang-2018} further discussed the representation/reference biduality which was confounded in the $\alpha$-geometry.

Gauge freedom of the Riemannian metric tensor has been investigated under the framework of  $(\rho,\tau)$-monotone embeddings~\cite{embeddingZhang-2015,ConfDiv-2016,NaudtsZhang-2018} in information geometry:
Let $\rho$ and $\tau$ be two strictly increasing functions, and $f$ a strictly convex function such that $f'(\rho(u))=\tau(u)$ (with $f^*$ denoting its convex conjugate).  Observe that the set of strictly increasing real-valued univariate functions has a group structure for the group operation chosen as the functional composition $\circ$.
Let us write $p_\theta(x)=p(x;\theta)$. 

The $(\rho,\tau)$-metric tensor  $\leftsup{\rho,\tau}g(\theta)=[\leftsup{\rho,\tau}g_{ij}(\theta)]_{ij}$ can be derived  from the $(\rho,\tau)$-divergence:
\begin{equation}
D_{\rho,\tau}(p:q)=\int \left( f(\rho(p(x))) + f^*(\tau(q(x))) - \rho(p(x))\tau(q(x)) \right)  \dnu(x)
\end{equation}

We have: 
\begin{eqnarray}
\leftsup{\rho,\tau}g_{ij}(\theta) &=& \int \left(\partial_i\rho(p_\theta(x))\right)   \left(\partial_j\tau(p_\theta(x))\right) \dnu(x),\\
&=& \int \rho'(p_\theta(x)) \tau'(p_\theta(x))  \left(\partial_i p_\theta(x) \right) \left(\partial_j p_\theta(x) \right)\dnu(x),\\
&=& \int f''(\rho(p_\theta(x)))  \left(\partial_i\rho(p_\theta(x))\right) \left(\partial_j\rho(p_\theta(x))\right)\dnu(x),\\
&=& \int (f^*)''(\tau(p_\theta(x))) \left(\partial_i\tau(p_\theta(x))\right) \left(\partial_j\tau(p_\theta(x))\right) \dnu(x).
\end{eqnarray}

\subsection{Dually flat spaces and canonical Bregman divergences} 

We have described how to build a dually flat space from any strictly convex and smooth generator $F$:
A Hessian structure is built from $F(\theta)$ with  Riemannian Hessian metric $\nabla^2 F(\theta)$, and the convex conjugate $F^*(\eta)$
 (obtained by the Legendre-Fenchel duality) yields the dual Hessian structure with  Riemannian Hessian metric $\nabla^2 F^*(\eta)$.
The dual connections $\nabla$ and $\nabla^*$ are coupled with the metric. The connections are defined by their respective Christoffel symbols $\Gamma(\theta)=0$ and $\Gamma^*(\eta)=0$, showing that they are flat connections.

Conversely, it can be proved~\cite{IG-2016} that given two dually flat connections $\nabla$ and $\nabla^*$, we can reconstruct  two dual canonical strictly convex potential functions $F(\theta)$ and $F^*(\eta)$ such that $\eta=\nabla F(\theta)$ and $\theta=\nabla F^*(\eta)$.
The canonical divergence $A_{F,F^*}$ yields the dual Bregman divergences $B_F$ and $B_{F^*}$.

The only symmetric Bregman divergences are squared Mahalanobis distances $M_Q^2$~\cite{BVD-2010} with the Mahalanobis distance defined by:
\begin{equation}
M_Q(\theta,\theta') = \sqrt{(\theta'-\theta)^\top Q (\theta'-\theta)}.
\end{equation}

Let $Q=LL^\top$ be the Cholesky decomposition of a positive-definite matrix $Q\succ 0$. 
It is well-known that the Mahalanobis distance $M_Q$ amounts to the Euclidean distance on affinely transformed points:
\begin{eqnarray}
M_Q^2(\theta,\theta') &=& \Delta\theta^\top Q\Delta\theta,\\
&=&\Delta\theta^\top LL^\top\Delta\theta,\\
&=& M_I^2(L^\top\theta,L^\top\theta')=\|L^\top\theta-L^\top\theta'\|^2,
\end{eqnarray}
where $\Delta\theta=\theta'-\theta$.

The squared Mahalanobis distance $M_Q^2$ does not satisfy the triangle inequality, but the  Mahalanobis distance $M_Q$ is a metric distance.
We can convert a Mahalanobis distance $M_{Q_1}$ into another Mahalanobis distance $M_{Q_2}$, and vice versa, as follows:
\begin{proof}
Let us write matrix $Q=L^\top L\succ 0$ using the Cholesky decomposition.
Then we have
\begin{equation}
M_Q(\theta_1,\theta_2)= M_I(L^\top\theta_1,L^\top\theta_2)  \Leftrightarrow
M_I(\theta_1,\theta_2) = M_Q((L^\top)^{-1}\theta_1,((L^\top)^{-1}\theta_2).
\end{equation}
Then we have for two symmetric positive-definite matrices $Q_1=L_1^\top L_1\succ 0$ and $Q_2=L_2^\top L_2\succ 0$:
\begin{equation}
M_{Q_1}(\theta_1,\theta_2)=
M_I(L_1^\top\theta_1,L_1^\top\theta_2)= 
M_{Q_2}((L_2^\top)^{-1}L_1^\top\theta_1,(L_2^\top)^{-1}L_1^\top\theta_2).
\end{equation}
It follows that we have:
\begin{equation}
M_{Q_1}(\theta_1,\theta_2)=M_{Q_2}((L_2^\top)^{-1}L_1^\top\theta_1,(L_2^\top)^{-1}L_1^\top\theta_2).
\end{equation}
\end{proof}

We have $M_Q^2(\theta_1,\theta_2)=B_{F}(\theta_1,\theta_2)$ (Bregman divergence)
with $F(\theta)=\frac{1}{2}\theta^\top Q\theta$ for a positive-definite matrix $Q\succ 0$.
The convex conjugate $F^*(\eta)=\frac{1}{2}\eta^\top Q^{-1}\eta$ (with $Q^{-1}\succ 0$). We have 
$\eta=Q^{-1}\theta$ and $\eta=Q\theta$.
We have the following identity between the {\em dual Mahalanobis divergences} $M_Q^2$ and $M_{Q^{-1}}^2$:
\begin{equation}
M_Q^2(\theta_1,\theta_2)=M_{Q^{-1}}^2(\eta_1,\eta_2).
\end{equation}

When the Bregman generator is based on an integral, i.e., 
the log-normalizer $F(\theta)=\log\left( \int \exp(\inner{t(x)}{\theta}\dmu(x)\right)$ for exponential families $\calE$, or the 
negative Shannon entropy $F(\theta)=\int m_\theta(x)\log m(\eta)\dmu(x)$ for mixture families $\calM$, the associated 
Bregman divergences $B_{F,\calE}$ or $B_{F,\calM}$ can be relaxed and interpreted as a statistical distance. 
We explain how to obtain the reconstruction below:
 
\begin{itemize}
\item Consider an exponential family $\calE$ of order $D$ with densities defined according to a dominating measure $\mu$:
\begin{equation}
\calE=\{p_\theta(x)=\exp(\theta^\top t(x)-F(\theta))\ :\ \theta\in\Theta\},
\end{equation}
where the natural parameter $\theta$ and the sufficient statistic vector $t(x)$ belong to $\bbR^D$.
We have the integral-based Bregman generator:
\begin{equation}
F(\theta)=F_\calE(p_\theta)=\log\left(\int\exp(\theta^\top t(x))\dmu(x)\right),
\end{equation} 
and the dual convex conjugate 
\begin{equation}
F^*(\eta)=-h(p_\theta)=\int p(x)\log p(x)\dmu(x),
\end{equation}
where $h(p)=-\int p(x)\log p(x)\dmu(x)$ denotes Shannon's entropy. 

Let $\lambda(i)$ denotes the $i$-th coordinates of vector $\lambda$,
and let us calculate the {\em inner product} $\theta_1^\top\eta_2=\sum_i \theta_1(i)\eta_2(i)$ of the Legendre-Fenchel divergence.
We have $\eta_2(i)=E_{p_{\theta_2}}[t_i(x)]$.
Using the linear property of the expectation $E[\cdot]$, 
we find that $\sum_i \theta_1(i)\eta_2(i)= E_{p_{\theta_2}}\left[\sum_i \theta_1(i) t_i(x)\right]$.  
Moreover, we have $\sum_i \theta_1(i) t_i(x)=\left(\log p_{\theta_1}(x)\right)+F(\theta_1)$.
Thus  we have:
\begin{equation}
\theta_1^\top\eta_2 = E_{p_{\theta_2}}\left[\log p_{\theta_1}+F(\theta_1)\right]=F(\theta_1)+E_{p_{\theta_2}}\left[\log p_{\theta_1}\right].
\end{equation}

It follows that we get
\begin{eqnarray}
B_{F,\calE}[p_{\theta_1}:p_{\theta_2}] &=& F(\theta_1)+F^*(\eta_2)-\theta_1^\top\eta_2,\\
&=& F(\theta_1)-h(p_{\theta_2})-E_{p_{\theta_2}}[\log p_{\theta_1}]-F(\theta_1),\\
&=& E_{p_{\theta_2}}\left[\log \frac{p_{\theta_2}}{p_{\theta_1}}\right]=:D_{\KL^*}[p_{\theta_1}:p_{\theta_2}],
\end{eqnarray}

By relaxing the exponential family densities $p_{\theta_1}$ and $p_{\theta_2}$ to be arbitrary densities $p_1$ and $p_2$, we obtain
 the {\em reverse KL divergence} between $p_1$ and $p_2$ from the dually flat structure induced by the integral-based log-normalizer of an exponential family:
\begin{eqnarray}
D_{\KL^*}[p_1:p_2]&=&E_{p_{2}}\left[\log \frac{p_{2}}{p_{1}}\right]=\int p_2(x)\log \frac{p_2(x)}{p_1(x)}\dmu(x),\\
&=& D_{\KL}[p_2:p_1].
\end{eqnarray}

Thus we have recovered the reverse Kullback-Leibler divergence $D_{\KL^*}$ from $B_{F,\calE}$.

The dual divergence $D^*[p_1:p_2]:=D[p_2:p_1]$ is obtained by swapping the distribution parameter orders.
We have:
\begin{equation}
D_{\KL^*}^*[p_1:p_2]:=D_{\KL^*}[p_2:p_1]= E_{p_{1}}\left[\log \frac{p_{1}}{p_{2}}\right]=:D_{\KL}[p_1:p_2],
\end{equation}
and $D_{\KL^*}[p_1:p_2]=D_{\KL^*}^*[p_2:p_1]=D_{\KL}[p_2:p_1]$.

To summarize, the canonical Legendre-Fenchel divergence associated with the log-normalizer of an exponential family amounts to
 the statistical reverse Kullback-Leibler divergence between $p_{\theta_1}$ and $p_{\theta_1}$ (or the KL divergence between the swapped corresponding densities): $D_\KL[p_{\theta_1}:p_{\theta_2}]=B_F(\theta_2:\theta_1)=A_{F,F^*}(\theta_2:\eta_1)$. 
Notice that it is easy to check that $D_\KL[p_{\theta_1}:p_{\theta_2}]=B_F(\theta_2:\theta_1)$~\cite{KLEF-2001,Bregman-2005}.
Here, we took the opposite direction by constructing $D_\KL$ from $B_F$.

We may consider an auxiliary carrier term $k(x)$ so that the densities write
$p_\theta(x)=\exp(\theta^\top t(x)-F(\theta)+k(x))$. Then the dual convex conjugate writes~\cite{crossEF-2010} 
as $F^*(\eta)=-h(p_\theta)+E_{p_\theta}[k(x)]$.

Notice that since the Bregman generator is defined up to an affine term, we may consider the equivalent generator $F(\theta)=-\log p_\theta(\omega)$ instead of the integral-based generator. This approach yields ways to build formula bypassing the explicit use of the log-normalizer for calculating various statistical
distances~\cite{CumulantFree-2020}.

\item In this second example, we consider a mixture family 
\begin{equation}
\calM=\left\{m_\theta=\sum_{i=1}^D \theta_i p_i(x)+ (1-\sum_{i=1}^D \theta_i)p_0(x)\right\},
\end{equation}
 where $p_0,\ldots, p_D$ are $D+1$ linearly independent probability densities.
The integral-based Bregman generator $F$ is chosen as Shannon negentropy: 
\begin{equation}
F(\theta)=F_\calM(m_\theta)=-h(m_\theta)=\int m_\theta(x)\log m_\theta(x)\dmu(x).
\end{equation}

We have 
\begin{equation}
\eta_i=[\nabla F(\theta)]_i=\int (p_i(x)-p_0(x))\log m_\theta(x)\dmu(x),
\end{equation} 
and the dual convex potential function 
is 
\begin{equation}
F^*(\eta)=-\int p_0(x)\log m_\theta(x)\dmu(x)=h^\times(p_0:m_\theta),
\end{equation}
i.e., the cross-entropy between the density $p_0$ and the mixture $m_\theta$.
Let us calculate the inner product $\theta_1^\top\eta_2$ of the Legendre-Fenchel divergence  as follows:

\begin{eqnarray}
\sum_i \theta_1(i)\int (p_i(x)-p_0(x))\log m_{\theta_2}(x)\dmu(x) &=& 
\int  \sum_i \theta_1(i) p_i(x)\log m_{\theta_2}(x)\dmu(x)  \nonumber \\ 
&& -\sum_i \theta_1(i)  p_0(x)\log m_{\theta_2}(x)\dmu(x).
\end{eqnarray}

That is
\begin{equation}
\theta_1^\top \eta_2=\int  \sum_i \theta_1(i) p_i\log m_{\theta_2}\dmu-\sum_i \theta_1(i)  p_0\log m_{\theta_2}\dmu.
\end{equation}

Thus it follows that we have the following statistical distance:
\begin{eqnarray}
B_{F,\calM}[m_{\theta_1}:m_{\theta_2}] &:=& F(\theta_1)+F^*(\eta_2)-\theta_1^\top\eta_2,\\
&=& -h(m_{\theta_1})-\int p_0(x)\log m_{\theta_2}(x)\dmu(x) - 
\int  \sum_i \theta_1(i) p_i(x)\log m_{\theta_2}(x)\dmu(x) \nonumber \\ 
&&+\sum_i \theta_1(i)  p_0(x)\log m_{\theta_2}(x)\dmu(x),\\
&=&-h(m_{\theta_1})-\int ((1-\sum_i \theta_1(i))p_0(x)+\sum_i \theta_1(i)p_i(x))\log m_{\theta_2}(x)\dmu(x),\\
&=& -h(m_{\theta_1})-\int m_{\theta_1}(x)\log m_{\theta_2}(x)\dmu(x),\\
&=& \int  m_{\theta_1}(x)\log \frac{m_{\theta_1}(x)}{m_{\theta_2}(x)} \dmu(x),\\
&=& D_\KL[m_{\theta_1}:m_{\theta_2}].
\end{eqnarray}

Thus we have $D_\KL[m_{\theta_1}:m_{\theta_2}]=B_F(\theta_1:\theta_2)$.
By relaxing the mixture densities $m_{\theta_1}$ and $m_{\theta_2}$ to arbitrary densities $m_1$
 and $m_2$, we find that the dually flat geometry induced by the negentropy of densities of a mixture family induces a 
statistical distance which corresponds to the (forward) KL divergence.
That is, we have recovered the statistical distance $D_{\KL}$ from $B_{F,\calM}$.
Note that in general the entropy of a mixture is not available in closed-form (because of the log sum term), except when the component distributions have pairwise disjoint supports.
This latter case includes the case of Dirac distributions whose mixtures represent the categorical distributions.
\end{itemize}
 
Dually flat spaces can be built from any strictly convex $C^3$ generator $F$.
Vinberg and Koszul~\cite{Shima-2007} showed how to obtain such a convex generator for homogeneous cones.
A cone $\mathcal{C}$ in a vector space $V$ yields a dual cone of positive linear functionals in the dual vector space $V^*$:
\begin{equation}
\mathcal{C}^{*}:=\left\{\omega \in V^{*}: \forall v \in \mathcal{C}, \omega(v) \geq 0\right\}.
\end{equation}
The characteristic function of the cone is defined by
\begin{equation}
\chi_{\mathcal{C}}(\theta):=\int_{\mathcal{C}^{*}} \exp (-\omega(\theta)) \mathrm{d} \omega \geq 0,
\end{equation}
and the function $\log\chi_{\mathcal{C}}(\theta)$ defines a Bregman generator which induces a Hessian structure and a dually flat space.

\begin{sidewaysfigure}
\includegraphics[width=\columnwidth]{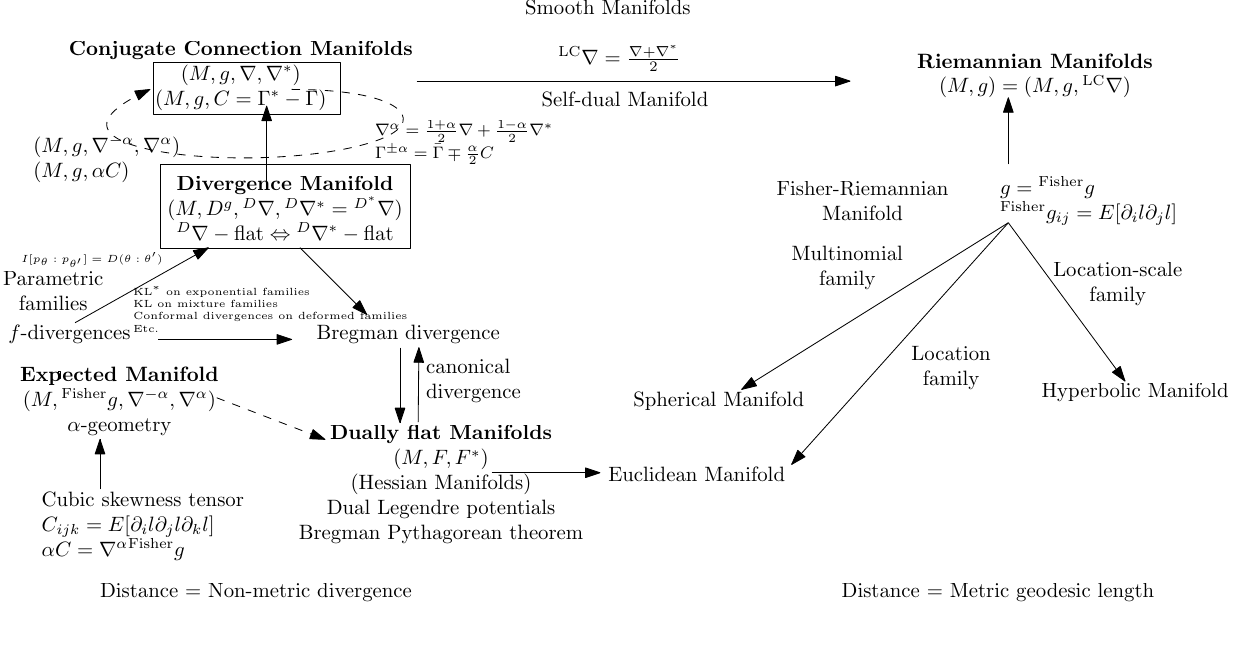}%
\caption{Overview of the main types of information manifolds with their relationships in information geometry.}%
\label{fig:overviewIGMfd}%
\end{sidewaysfigure}

Figure~\ref{fig:overviewIGMfd} displays the main types of information manifolds encountered in information geometry with their relationships.

\def\diag{\mathrm{diag}}

\section{Some applications of information geometry}\label{sec:AppIG}

Information geometry~\cite{IG-2016} found broad applications in information sciences.
For example, we can mention:

\begin{itemize}
	\item Statistics:
	Asymptotic inference, Expectation-Maximization (EM and the novel information-geometric em), time series (AutoRegressive Moving Average model, ARMA) models,
 
	\item Machine learning:
	Restricted Boltzmann machines (RBMs), neuromanifolds and natural gradient~\cite{RFIM-2017},
 
	\item Signal processing:
	Principal Component Analysis (PCA), Independent Component Analysis (ICA), Non-negative Matrix Factorization (NMF),
 
	\item Mathematical programming:
	Barrier function of interior point methods,
	
	\item Game theory:
	Score functions.
\end{itemize}

Next, we  shall describe a few applications, starting with the celebrated natural gradient descent.

\subsection{Natural gradient in Riemannian space}\label{sec:appng}
The {\em Natural Gradient}~\cite{Amari-1998} (NG) is an extension of the ordinary (Cartesian) gradient of Euclidean geometry to the gradient in a Riemannian space analyzed in an arbitrary coordinate system. We explain the natural gradient
 
\subsubsection{The vanilla gradient descent method}

Given a real-valued function $L_\theta(\theta)$ parameterized by a a $D$-dimensional vector $\theta$  on parameter space 
$\theta\in\Theta\subset\bbR^D$, we wish to minimize $L_\theta$, i.e., solve $\min_{\theta\in\Theta} L_\theta(\theta)$.
The {\em gradient descent} (GD) method, also called the {\em steepest descent} method, is a first-order local optimization procedure which starts by initializing the parameter to an arbitrary value (say, $\theta_0\in\Theta$), and then iteratively updates at stage $t$  
the current location of $\theta_t$  to $\theta_{t+1}$ as follows:
\begin{equation}\label{eq:gd}
\mathrm{GD}:\quad \theta_{t+1} = \theta_t - \alpha_t \nabla_\theta L_\theta(\theta_t).
\end{equation}

The scalar $\alpha_t>0$ is called the {\em step size} or {\em learning rate} in machine learning.
The ordinary gradient (OG) $\nabla_\theta F_\theta(\theta)$ (vector of partial derivatives) represents the {\em steepest vector} at $\theta$ of the function graph $\calL_\theta=\{(\theta,L_\theta(\theta))\ :\ \theta\in\Theta\}$.
The GD method was pioneered by Cauchy~\cite{Cauchy-1847} (1847) and its convergence proof to a {\em stationary point} was first reported in Curry~\cite{Curry-1944} (1944).

If we {\em reparameterize} the function $L_\theta$ using a one-to-one and onto differentiable mapping $\eta=\eta(\theta)$ (with reciprocal inverse mapping $\theta=\theta(\eta)$), the GD update rule transforms as:
\begin{equation}
\eta_{t+1} = \eta_t-\alpha_t \nabla_\eta L_\eta(\eta_t),
\end{equation}
where 
\begin{equation}
L_\eta(\eta):=L_\theta(\theta(\eta)).
\end{equation}

Thus in general, the two gradient descent location sequences $\{\theta_t\}_t$ and $\{\eta_t\}_t$ (initialized at $\theta_0=\theta(\eta_0)$ and $\eta_0=\eta(\theta_0)$) are  {\em different} (because usually $\eta(\theta)\not=\theta$), and the two GDs may potentially reach different stationary points.
In other words, the GD local optimization {\em  depends on the choice of the parameterization} of the function $L$ (i.e., $L_\theta$ or $L_\eta$).
For example,
 minimizing with the gradient descent a temperature function $L_\theta(\theta)$ with respect to Celsius degrees $\theta$ may yield a different result 
than minimizing the same temperature function $L_\eta(\eta)=L_\theta(\theta(\eta))$ expressed with respect to Fahrenheit degrees $\eta$.
That is, the GD optimization is {\em extrinsic} since it depends on the choice of the parameterization of the function, and does not take into account the underlying geometry of the parameter space $\Theta$.

The natural gradient precisely addresses  this problem and solves it by choosing {\em intrinsically} the steepest direction with respect to a Riemannian metric tensor field on the parameter manifold. We shall explain the natural gradient descent method and highlight its connections with the Riemannian gradient descent, the mirror descent and even the ordinary gradient descent when the parameter space is dually flat. 

\subsubsection{Natural gradient and its connection with the Riemannian gradient}

Let $(M,g)$ be a $D$-dimensional Riemannian space~\cite{DG-2016} equipped with a metric tensor $g$, and $L\in C^\infty(M)$ a smooth function to minimize on  the manifold $M$.
The {\em Riemannian gradient}~\cite{Bonnabel-2013} uses the Riemannian  {\em exponential map} 
$\exp_{p}: T_p\rightarrow M$ to update the sequence of points 
$p_t$'s on the manifold as follows:
\begin{equation}
\mathrm{RG}:\quad p_{t+1} =  \exp_{p_t}(-\alpha_t \nabla_{M} L(p_t)),
\end{equation}
where the Riemannian gradient $\nabla_{M}$ is defined according to a {\em directional derivative} $\nabla_v$ by:
\begin{equation}
\nabla_{M} L(p):=\left.\nabla_v\left(L\left(\exp_p(v)\right)\right)\right|_{v=0},
\end{equation}
with
\begin{equation}
\nabla_{v} L(p) :=\lim_{h\rightarrow 0} \frac{L(p+hv)-L(p)}{h}.
\end{equation}

However, the Riemannian exponential mapping $\exp_p(\cdot)$ is often computationally intractable since it requires to solve a system of second-order differential equations~\cite{DG-2016,MatrixMfd-2009}.
Thus instead of using $\exp_p$,  we shall rather use a computable {\em Euclidean retraction} $R: T_p\rightarrow\bbR^D$ of the exponential map 
 expressed in a local $\theta$-coordinate system as:
\begin{equation}
\mathrm{RetG}:\quad \theta_{t+1} =  R_{\theta_t}\left(-\alpha_t \nabla_{\theta} L_\theta(\theta_t)\right).
\end{equation}

Using the retraction~\cite{MatrixMfd-2009} $R_p(v)=p+v$ which corresponds to a first-order Taylor approximation of the exponential map, 
we recover the {\em natural gradient descent}~\cite{Amari-1998}:   
\begin{equation}\label{eq:ngd}
\mathrm{NG}: \theta_{t+1}= \theta_t-\alpha_t g^{-1}_\theta(\theta_t) \nabla_{\theta} L_\theta(\theta_t).
\end{equation}

The {\em natural gradient}~\cite{Amari-1998} (NG)
\begin{equation}
\nablaNG L_\theta(\theta) := g^{-1}_\theta(\theta) \nabla_\theta L_\theta(\theta)
\end{equation}
 encodes the {\em Riemannian steepest descent} vector, and the natural gradient descent method yields the following update rule
\begin{equation}
\mathrm{NG}: \theta_{t+1}= \theta_t-\alpha_t \nablaNG L_\theta(\theta_t).
\end{equation}

Notice that the natural gradient is a {\em contravariant vector} while the ordinary gradient is a {\em covariant vector}.
Recall that a covariant vector $[v_i]$ is transformed into a contravariant vector $[v^i]$ by $v^i=\sum_j g^{ij}v_i$, 
that is by using the dual Riemannian metric $g^*_\eta(\eta)=g_\theta(\theta)^{-1}$.
The natural gradient is {\em invariant} under an invertible smooth change of parameterization.
However, the natural gradient {\em descent}  does {\em not} guarantee that the locations $\theta_t$'s always stay on the manifold:
Indeed, it may happen that for some $t$, $\theta_t\not\in\Theta$ when $\Theta\not=\bbR^D$.

\begin{Property}[\cite{Bonnabel-2013}]
The natural gradient descent approximates the intrinsic Riemannian gradient descent using a contravariant gradient vector induced by the Riemannian metric tensor $g$. The natural gradient is invariant to coordinate transformations.
\end{Property}

Next, we shall explain how the natural gradient descent is related to the {\em mirror descent} and the {\em ordinary gradient} when the Riemannian space $\Theta$ is dually flat.

\subsubsection{Natural gradient in dually flat spaces: Connections to Bregman mirror descent and ordinary gradient}

Recall that a dually flat space $(M,g,\nabla,\nabla^*)$ is a manifold $M$ equipped with a pair $(\nabla,\nabla^*)$ of dual torsion-free flat connections which are coupled to the Riemannian metric tensor $g$~\cite{IG-2016,DFS-2019} in the sense that
 $\frac{\nabla+\nabla^*}{2}=\leftsup{LC}\nabla$, where 
$\leftsup{LC}\nabla$ denotes the unique metric torsion-free Levi-Civita connection.

On a dually flat space, there exists a pair of dual global {\em Hessian structures}~\cite{Shima-2007} with dual canonical Bregman divergences~\cite{Bregman-1967,IG-2016}. The dual Riemannian metrics can be expressed as the Hessians of dual convex potential functions $F$ and $F^*$. 
Examples of Hessian manifolds are the {\em manifolds of exponential families} or the {\em manifolds of mixture families}~\cite{MCIG-2019}.
On a dually flat space induced by a strictly convex and $C^3$ function $F$ (Bregman generator), we have two dual global coordinate system:
$\theta(\eta)=\nabla F^*(\eta)$ and $\eta(\theta)=\nabla F(\theta)$, where $F^*$ denotes the Legendre-Fenchel convex conjugate function~\cite{LegendreIG-2010,CRLB-IG-2013}.
The Hessian metric expressed in the primal $\theta$-coordinate system is $g_\theta(\theta)=\nabla^2 F(\theta)$, and the dual Hessian metric expressed
 in the dual coordinate system is $g^*_\eta(\eta)=\nabla^2 F^*(\eta)$. 
Crouzeix's identity~\cite{Crouzeix-1977} shows that $g_\theta(\theta)g_\eta(\eta)=I$, where $I$ denotes the $D\times D$ matrix identity.

The ordinary gradient descent method can be extended using a  {\em proximity function} $\Phi(\cdot,\cdot)$ as follows:
\begin{equation}
\mathrm{PGD}:\quad \theta_{t+1} = \arg \min_{\theta\in\Theta} \left\{ \inner{\theta}{\nabla L_\theta(\theta_t)}+\frac{1}{\alpha_t} \Phi(\theta,\theta_t)\right\}.
\end{equation}
When $\Phi(\theta,\theta_t)=\frac{1}{2}\|\theta-\theta_t\|^2$, the PGD update rule becomes the ordinary GD update rule.

Consider a Bregman divergence~\cite{Bregman-1967} $B_F$ for the proximity function $\Phi$: $\Phi(p,q)=B_F(p:q)$.
Then the PGD yields the following {\em mirror descent} (MD):
\begin{equation}
\mathrm{MD}:\quad\theta_{t+1} = \arg \min_{\theta\in\Theta} \left\{ \inner{\theta}{\nabla L(\theta_t)}+\frac{1}{\alpha_t} B_F(\theta:\theta_t)\right\}.
\end{equation}

This mirror descent can be interpreted as a natural gradient descent as follows:

\begin{Property}[\cite{MDIG-2015}]
Bregman mirror descent on the Hessian manifold $(M,g=\nabla^2 F(\theta))$ 
is equivalent to natural gradient descent on the dual Hessian manifold $(M,g^*=\nabla^2 F(\eta))$, where $F$ is a Bregman generator, $\eta=\nabla F(\theta)$ and $\theta=\nabla F^*(\eta)$.
\end{Property}

Indeed, the mirror descent rule yields the following natural gradient update rule:
\begin{eqnarray}
\mathrm{NG}^*:\eta_{t+1} &=&\eta_t-\alpha_t (g^*_\eta)^{-1}(\eta_t)\nabla_{\eta} L_\theta(\theta(\eta_t)),\\
&=& \eta_t-\alpha_t (g^*_\eta)^{-1}(\eta_t)\nabla_{\eta} L_\eta(\eta_t),
\end{eqnarray}
where $g^*_\eta(\eta)=\nabla^2 F^*(\eta)=(\nabla^2_\theta F(\theta))^{-1}$ and $\theta(\eta)=\nabla F^*(\theta)$.

The method is called mirror descent~\cite{Bubeck-2015} because it performs that gradient step in the {\em dual space} (ie., mirror space) $H=\{\eta=\nabla F(\theta)\ : \ \theta\in\Theta\}$, and thus solves the inconsistency contravariant/covariant type problem of subtracting a covariant vector from a contravariant vector of the ordinary GD (Eq.~\ref{eq:gd}).

Let us prove now the following property of the natural gradient in a dually flat space or Bregman manifold~\cite{DFS-2019}:

\begin{Property}[\cite{Zhang-2018}]
In a dually flat space induced by potential convex function $F$, the natural gradient amounts to the ordinary gradient on the dually parameterized function: $\nablaNG L_\theta(\theta)=\nabla_\eta L_\eta(\eta)$ where $\eta=\nabla_\theta F(\theta)$ and $L_\eta(\eta)=L_\theta(\theta(\eta))$. 
\end{Property}

\begin{proof}
Let $(M,g,\nabla,\nabla^*)$ be a dually flat space. We have $g_\theta(\theta)=\nabla^2 F(\theta)=\nabla_\theta \nabla_\theta F(\theta)=\nabla_\theta\eta$ since $\eta=\nabla_\theta F(\theta)$.
The function to minimize can be written either as $L_\theta(\theta)=L_\theta(\theta(\eta))$ or as $L_\eta(\eta)=L_\eta(\eta(\theta))$.
Recall the chain rule in the calculus of differentiation:
\begin{equation}
\nabla_\theta L_\theta(\theta)= \nabla_\theta (L_\eta(\eta(\theta)))=
 (\nabla_\theta\eta) (\nabla_\eta L_{\eta}(\eta)).
\end{equation}

Thus we have:
\begin{eqnarray}
\nablaNG L_\theta(\theta) &\eqdef& g_\theta^{-1}(\theta) \nabla_\theta L_\theta(\theta),\\
&=& (\nabla_\theta \eta)^{-1} (\nabla_\theta\eta) \nabla_\eta L_{\eta}(\eta),\\
&=& \nabla_\eta L_\eta(\eta).
\end{eqnarray}
\end{proof}

It follows that the natural gradient descent on a loss function $L_\theta(\theta)$ amounts to an ordinary gradient descent on the {\em dually parameterized} loss function $L_\eta(\eta):=L_\theta(\theta(\eta))$.
In short, $\nablaNG_\theta L_\theta=\nabla_\eta L_\eta$.

\subsubsection{An application of the natural gradient: Natural Evolution Strategies (NESs)}

\def\grad{\nabla}
\def\agrad{\stackrel{\sim}{\nabla}}
\def\ngrad{ \nablaNG_\theta}
\def\angrad{\widetilde{\nablaNG}_\theta}

A nice family of applications of the natural gradient are the Natural Evolution Strategies (NESs) for black-box minimization~\cite{Beyer-2002}:
Let $f(x)$ for $x\in\bbX\subset\bbR^d$ be a real-valued function to minimize.
Berny~\cite{SearchGrad-2000} proposed to {\em relax} the optimization problem $\min_{x\in\bbX} f(x)$ by considering a parametric search distribution $p_\lambda$, and minimize instead:
\begin{equation}
\min_{\lambda\in\Lambda} E_{p_\lambda}[f(x)],
\end{equation}
where $\lambda\in\Lambda\subset\bbR^D$ denotes the parameter space of the search distributions.
Let $J(\lambda)=E_{p_\lambda}[f(x)]$.
Minimizing $J(\lambda)$ instead of $f(x)$ is particularly useful when $\bbX$ is a discrete space: 
Indeed, the {\em combinatorial optimization}~\cite{SearchGrad-2000} $\min_{x\in\bbX} f(x)$ is replaced by a continuous optimization $\min_{\lambda\in\Lambda} J(\lambda)$ when $\Lambda$ is a continuous parameter, and the ordinary or natural GD methods can be used.
The gradient $\grad J(\lambda)$ is called the {\em search gradient}, and it can be approximated stochastically using the log-likelihood trick~\cite{NES-2014} as
\begin{equation}
\agrad J(\lambda):=\frac{1}{n}\sum_{i=1}^n f(x_i)\grad \log p_\lambda(x_i)\approx \grad J(\lambda),
\end{equation}
where $x_1,\ldots, x_n\sim p_\lambda$.
Similarly, the Fisher information matrix (FIM) may be approximated by the following empirical FIM:
\begin{equation}
\tilde{I}(\lambda) = \frac{1}{n} \sum_{i=1}^n \nabla_\lambda l_\lambda(x_i) (\nabla_\lambda l_\lambda(x_i))^\top \approx I(\lambda),
\end{equation}
where $l_\lambda(x):=\log p_\lambda(x)$ denote the log-likelihood function.
Notice that the approximated FIM may potentially be degenerated and may not respect the structure of the true FIM.
For example, we have $\nabla_\mu l(x;\mu,\sigma^2)= \frac{x-\mu}{\sigma^2}$ and $\nabla_{\sigma^2}= \frac{(x-\mu)^2}{2\sigma^4}-\frac{1}{2\sigma^2}$. The non-diagonal of the approximate FIM $\tilde{I}(\lambda)$ are close to but usually non-zero although the expected FIM is diagonal $I(\mu,\sigma^2)=\diag\left(\frac{1}{\sigma^2},\frac{1}{2\sigma^4}\right)$.
Thus we may estimate the FIM until the non-diagonal elements have absolute values less than a prescribed $\epsilon>0$.
For multivariate normals, we have $\grad_\mu l(x;\mu,\Sigma)=\Sigma^{-1}(x-\mu)$ and
$\grad_\Sigma l(x;\mu,\Sigma)=\frac{1}{2} (\grad_\mu l(x;\mu,\Sigma)\grad_\mu l(x;\mu,\Sigma)^\top-\Sigma^{-1})$.

%

\subsection{Some illustrating applications of dually flat manifolds}\label{sec:AIG}

In this part, we describe how to use the dually flat structures for handling an exponential family $\calE$ (in a hypothesis testing problem detailed in~\S\ref{sec:ht}) and the mixture family $\calM$ (clustering statistical mixtures~\S\ref{sec:cm}).
Note that for a general divergence, neither $(\calE,D)$ nor $(\calM,D)$ is  dually flat. However, when $D=\KL$, the Kullback-Leibler divergence, we get dually flat spaces that are computationally attractive since the primal/dual geodesics are straight lines in the corresponding global affine coordinate system.

\subsection{Hypothesis testing in the dually flat exponential family manifold $(\calE,\KL^*)$}\label{sec:ht}

\begin{figure} 
\begin{center}
\includegraphics[width=8cm]{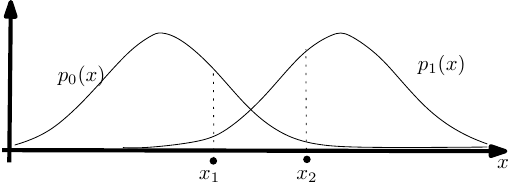}
\end{center}
\caption{Statistical Bayesian hypothesis testing: The best Maximum A Posteriori (MAP) rule chooses to classify an observation from the class that yields the maximum likelihood.\label{fig:ht}}
\end{figure}

Given two probability distributions $P_0\sim p_0(x)$ and $P_1\sim p_1(x)$, we ask to 
classify a set of iid. observations $X_{1:n}=\{x_1,\ldots, x_n\}$    as either sampled from $P_0$ or from $P_1$?
This is a statistical decision problem~\cite{HT-2013}. For example, $P_0$ can represent the signal distribution and $P_1$ the noise distribution.
Figure~\ref{fig:ht} displays the probability distributions and the unavoidable error that is made by any statistical decision rule (on observations $x_1$ and $x_2$).

Assume that both distributions $P_0\sim P_{\theta_0}$ and $P_1\sim P_{\theta_1}$
 belong to the same {\em exponential family} $\calE=\{P_{\theta} : \theta\in\Theta\}$, and consider the
 exponential family manifold with the dually flat structure  $(\calE,{{}_\calE}g,{{}_\calE}\nabla^e,{{}_\calE}\nabla^m)$.
That is, the manifold equipped with the Fisher information metric tensor field and the expected exponential connection and conjugate expected mixture connection.
More generally, the expected $\alpha$-geometry of an exponential family $\calE$ with cumulant function $F$ is given by:
\begin{eqnarray}
g_{ij}(\theta) &=& \partial_i\partial_j F(\theta),\\
\Gamma_{ij,k}^\alpha &=& \frac{1-\alpha}{2}\partial_i\partial_j\partial_k F(\theta).
\end{eqnarray}
When $\alpha=1$, $\Gamma_{ij,k}^\alpha=0$ and $\nabla^1$ is flat, and so is $\nabla^{-1}$ by using the fundamental theorem of information geometry. 

The $\pm 1$-structure can also be derived from a {\em divergence manifold structure} by choosing the reverse Kullback-Leibler divergence $\KL^*$:
\begin{equation}
(\calE,{{}_\calE}g,{{}_\calE}\nabla^e,{{}_\calE}\nabla^m) \equiv (\calE,\KL^*).
\end{equation}

Therefore, the Kullback-Leibler divergence $\KL[P_\theta:P_{\theta'}]$ amounts to a Bregman divergence (for the cumulant function of the exponential family):

\begin{equation}
\KL^*[P_{\theta'}:P_{\theta}] = \KL[P_\theta:P_{\theta'}] = B_F(\theta':\theta).
\end{equation}

The {\em best exponent error} $\alpha^*$ of the best  {\em Maximum A Priori} (MAP) decision rule is found by minimizing the
{\em Bhattacharyya distance} to get the {\em Chernoff information}~\cite{TensorHT-2018}:
\begin{equation}
C[P_1,P_2] = -\log \min_{\alpha\in(0,1)} \int_{x\in\calX} p_1^{\alpha}(x)p_2^{1-\alpha}(x) \dmu(x)\geq 0.
\end{equation}
 
On the exponential family manifold $\calE$, the {\em Bhattacharyya distance}:
\begin{equation}
B_\alpha[p_1:p_2]=-\log \int_{x\in\calX} p_1^{\alpha}(x)p_2^{1-\alpha}(x) \dmu(x),
\end{equation}
 amounts to a {\em skew Jensen parameter divergence}~\cite{BR-2011} (also called Burbea-Rao divergence):
\begin{equation}
J_{F}^{\alpha}(\theta_1 : \theta_2)= \alpha F(\theta_1)+ (1-\alpha)F(\theta_2)-  F(\theta_1+ (1-\alpha)\theta_2).
\end{equation}
 
It can be shown that the Chernoff information (that minimizes $\alpha$) is equivalent to a Bregman divergence:
Namely, the Bregman divergence for exponential families at the optimal exponent value $\alpha^*$.

\begin{Theorem}[Chernoff information~\cite{HT-2013}] 
The Chernoff information between two distributions belonging to the same exponential family amount to a Bregman divergence:
\begin{equation}
C[P_{\theta_1}:P_{\theta_2}]=B(\theta_1:\theta_{12}^{\alpha^*})= B(\theta_2:\theta_{12}^{\alpha^*}),
\end{equation}
where $\theta_{12}^{\alpha}=(1-\alpha)\theta_1+\alpha\theta_2$, and $\alpha^*$ denote the best exponent error.
\end{Theorem}
 
Let $\theta_{12}^*\eqdef \theta_{12}^{\alpha^*}$ denote the best exponent error.
The geometry~\cite{HT-2013} of the best error exponent can be  explained on the dually flat exponential family manifold as follows:
\begin{equation}
P^* = P_{\theta_{12}^*} = {G_e(P_1,P_2)} \cap {\Bi_m(P_1,P_2)},
\end{equation}
where $G_e$ denotes the exponential geodesic $\gamma_{\nabla^e}$ and $\Bi_m$ the $m$-bisector:
\begin{equation}
\Bi_m(P_1,P_2) = \{ P \ :\ F(\theta_1)-F(\theta_2) + \eta(P)^\top (\theta_2-\theta_1)=0\}.
\end{equation}

\begin{figure}
\begin{center}
\includegraphics[width=0.45\textwidth]{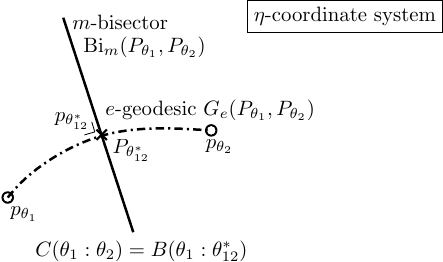}
\end{center}
\caption{Exact geometric characterization (not necessarily i closed-form) of the best exponent error rate $\alpha^*$.\label{fig:errexp}}
\end{figure}

Figure~\ref{fig:errexp} illustrates how to retrieve the best error exponent from an exponential arc ($\theta$-geodesic) intersecting the  $m$-bisector.

\begin{figure}
\begin{center}
 \includegraphics[width=0.6\textwidth]{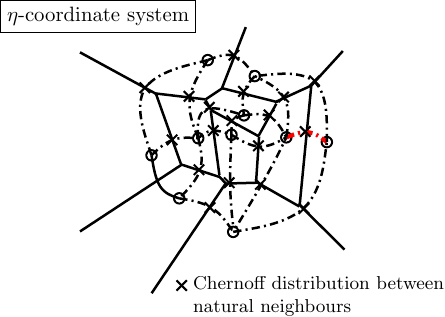}
\end{center}
\caption{Geometric characterization of the best exponent error rate in the multiple hypothesis testing case.\label{fig:merrexp}}
\end{figure}

Furthermore, instead of considering two distributions for this statistical binary decision problem, 
we may consider a set of $n$ distributions  of $P_1,\ldots, P_n\in\calE$.
The geometry of the error exponent in this multiple  hypothesis testing setting has been investigated in~\cite{MHT-2013}.
On the dually flat exponential family manifold, it corresponds to check the exponential arcs between {\em natural neighbors} (sharing Voronoi subfaces) of a
Bregman Voronoi diagram~\cite{BVD-2010}. See Figure~\ref{fig:merrexp} for an illustration.

\subsection{Clustering mixtures in the dually flat mixture family manifold $(\calM,\KL)$}\label{sec:cm}

	Given a set of $k$ prescribed statistical distributions $p_0(x),\ldots, p_{k-1}(x)$, all sharing the same support $\calX$ (say, $\bbR$), a
 {\em mixture family} $\calM$ of order $D=k-1$ consists of all {\em strictly convex combinations} of these component distributions~\cite{geowmixtures-2018}:
\begin{equation}
\calM \eqdef  \left\{ m(x;\theta)= \sum_{i=1}^{k-1} \theta_ip_i(x)+ \left(1-\sum_{i=1}^{k-1} \theta_i\right)p_0(x) \st \theta_i>0, \sum_{i=1}^{k-1} \theta_i<1 \right\}.
\end{equation}

\begin{figure}
\begin{center}
 \includegraphics[width=0.6\textwidth]{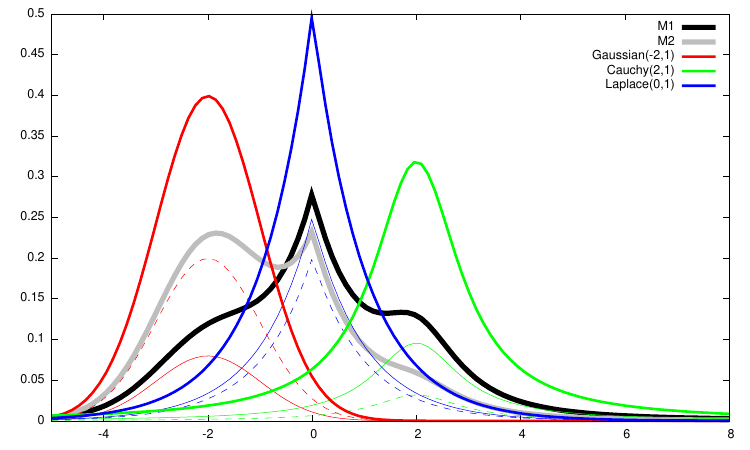}
\end{center}
\caption{Example of a mixture family of order $D=2$ ($3$ components: Laplacian, Gaussian and Cauchy prefixed distributions).}\label{fig:exmf}
\end{figure}

Figure~\ref{fig:exmf} displays two mixtures obtained as convex combinations of prescribed Laplacian, Gaussian and Cauchy component distributions ($D=2$). When considering a set of prescribed Gaussian component distributions, we obtain a $w$-Gaussian Mixture Model, or $w$-GMM for short.

We consider the expected information manifold $(\calM,{{}^\calM}g,{{}^\calM}\nabla^{m},{{}^\calM}\nabla^{e})$ which is dually flat and
 equivalent to $(M_\Theta,\KL)$. That is, the KL between two mixtures with prescribed components ($w$-mixtures, for short) is equivalent to a Bregman divergence for $F(\theta)=-h(m_{\theta})$, where $h(p)=\int p(x)\log p(x)\dmu(x)$ 
is the differential Shannon information (negative entropy)~\cite{geowmixtures-2018}:

\begin{equation}
\KL[m_{\theta_1}:m_{\theta_2}] = B_F(\theta_1:\theta_2).
\end{equation}

Consider a set $\{m_{\theta_1},\ldots, m_{\theta_n}\}$ of $n$ $w$-mixtures~\cite{geowmixtures-2018}.
Because $F(\theta)=-h(m(x;\theta))$ is the {\em negative differential entropy} of a mixture (not available in closed form~\cite{LSE-2016}), we approximate the untractable $F$ by another close tractable generator $\tilde{F}$.
We use Monte Carlo stochastic sampling to get Monte-Carlo convex $\tilde{F}_\calS$ for an independent and identically distributed sample $\calS$.

Thus we can build a {\em nested sequence} $(\calM,\tilde{F}_{\calS_1}), \ldots, (\calM,\tilde{F}_{\calS_m})$ of tractable dually flat manifolds for nested sample sets $\calS_1\subset \ldots\subset \calS_m$  converging to the ideal mixture manifold $(\calM,F)$: 
$\lim_{m\rightarrow\infty} (\calM,\tilde{F}_{\calS_m})=(\calM,F)$ (where convergence is defined with respect to the induced canonical Bregman divergence).
A key advantage of this approach is that for a given sample $\calS$, all computations carried inside the 
dually flat manifold $(\calM,\tilde{F}_\calS)$ are {\em consistent}, see~\cite{geowmixtures-2018}.

\begin{figure}
 \begin{center}
 \includegraphics[width=0.6\textwidth]{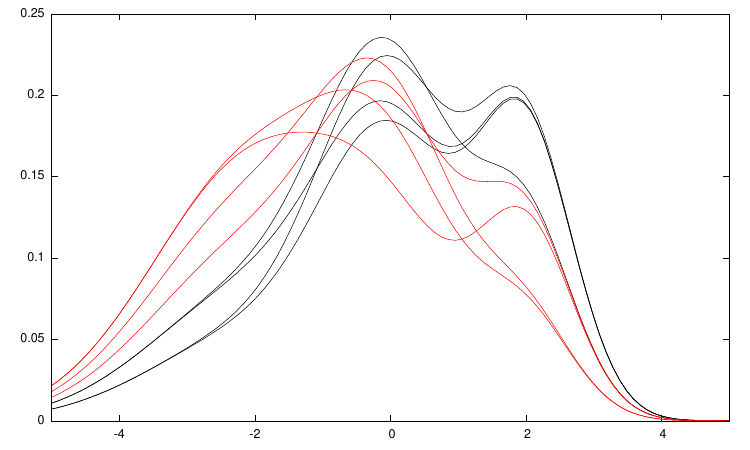}
\end{center}
\caption{Example of $w$-GMM clustering into $k=2$ clusters. }\label{fig:wgmmclust}
\end{figure}

For example, we can apply Bregman $k$-means~\cite{SBC-2009} on these Monte Carlo dually flat spaces~\cite{MCIG-2018} of $w$-GMMs (Gaussian Mixture Models) to cluster a set of $w$-GMMs.
Figure~\ref{fig:wgmmclust} displays the result of such a clustering.

We have briefly described two applications using dually flat manifolds: (1) the  {\em dually flat exponential manifold} induced by the statistical reverse Kullback-Leibler divergence on an exponential family (structure $(\calE,\KL^*)$), and (2) the {\em dually flat mixture manifold} induced by the statistical Kullback-Leibler divergence on a mixture family (structure $(\calM,\KL)$).
There are  many other dually flat structures that can be met in a statistical context:
For example, two other dually flat structures for the $D$-dimensional probability simplex $\Delta_D$ are reported in Amari's textbook~\cite{IG-2016}:
(1) the conformally deforming of the $\alpha$-geometry  (page 88, Eq. 4.95 of~\cite{IG-2016}), and 
(2) the $\chi$-escort geometry 
(page 91, Eq. 4.114 of~\cite{IG-2016}).


\section{Conclusion: Summary, historical background, and perspectives}\label{sec:concl}

\subsection{Summary}
We explained the dualistic nature of information manifolds $(M,g,\nabla,\nabla^*)$ in information geometry.
The dualistic structure is defined by  a pair of conjugate connections coupled with the metric tensor that provides a  dual parallel transport that preserves the metric. 
We showed how to extend this structure to a $1$-parameter family of structures:
From a pair of conjugate connections, the pipeline to build this $1$-parameter family of structures can be informally summarized as:
\begin{equation}
(M,g,\nabla,\nabla^*) \Rightarrow (M,g,C) \Rightarrow (M,g,\alpha C) \Rightarrow (M,g,\nabla^{-\alpha},\nabla^\alpha),\quad \forall\alpha\in\bbR. 
\end{equation}

We stated the fundamental theorem of information geometry on dual constant-curvature manifolds, including the special but important case of dually flat manifolds on which there exists two potential functions and global affine coordinate systems related by the Legendre-Fenchel transformation. Although, information geometry historically started with the Riemannian modeling $(\calP,\leftsub{\calP}g)$ of a parametric family of probability distributions $\calP$ by letting the metric tensor be the Fisher information matrix, we have emphasized   the dualistic view of information geometry  which considers non-Riemannian manifolds that can be derived from any divergence, and not necessarily tied to a statistical context (e.g., information manifold can be used in mathematical programming~\cite{MP-2007}).
Let us notice that for any symmetric divergence (e.g. any symmetrized $f$-divergence like the squared Hellinger divergence), the induced conjugate connections coincide with the Levi-Civita connection but the Fisher-Rao metric distance does not coincide with the squared Hellinger divergence.

On one hand, a Riemannian metric distance $D_\rho$ is {\em never} a divergence because the rooted distance functions fail to be smooth at the extremities but a squared Riemmanian metric distance is always a divergence.
On the other hand, taking the power $\delta$ of a divergence $D$ (i.e., $D^\delta$) for some $\delta>0$ may yield a metric distance (e.g., the square root of the Jensen-Shannon divergence~\cite{JSmetric-2004}), but this may not always be the case: The powered Jeffreys divergence $J^\delta$ is never a metric distance (see~\cite{MetricDiv-2009}, page 889).
Recently, the {\em Optimal Transport} (OT) theory~\cite{OT-2008} gained interest in statistics and machine learning.
But the  optimal transport between two members of a same elliptically-contoured family has the same optimal transport formula distance (see~\cite{FrechetGaussian-1982} Eq.~16 and Eq.~17, although they have different Kullback-Leibler divergences). 
Another essential difference is that the Fisher-Rao manifold of location-scale families is hyperbolic but the Wasserstein manifold of location-scale families has positive curvature~\cite{FrechetGaussian-1982,OTGaussian-2011}.

Notice that we may convert back and forth a similarity $S(p, q)\in (0,1]$ to a dissimilarity $D(p, q)\in[0, \infty)$ as follows:
\begin{eqnarray}
S(p, q) &=&\exp (-D(p, q)) \in(0,1] \\
D(p, q)&=&-\log S(p, q) \in[0, \infty)
\end{eqnarray}
When the dissimilarity satisfies the (additive) triangle inequality (i.e., $D(p, q)+D(q, r) \geq D(p, r)$ for any triple $(p,q,r)$) 
then the corresponding similarity satisfies the multiplicative triangle inequality: $S(p, q) \times S(q, r) \leq S(p, r)$.
A metric transform on a metric distance $D$ is a transformation $T$ such that $T(D(p,q))$ is a metric.
The transformation $T(u)=\frac{1}{1+u}$ is a metric transform which bounds potentially unbounded metric distances:
That is, if $D$ is an unbounded metric, then $T(D(p,q))=\frac{D(p,q)}{1+D(p,q)}$ is a bounded metric distance.
The transformation $S(u)=u^2$ is not a metric transform since the squared of the Euclidean metric distance is not a metric distance.


\subsection{A brief historical review of information geometry\label{sec:history}}

The field of Information Geometry (IG) was historically motivated by providing some differential-geometric structures to statistical models in order to reason geometrically about statistical problems with the endeavor goal of geometrizing mathematical statistics~\cite{Chentsov-1982,Amari-1985,Dodson-1987,DGStat-1993,IG-JP-Amari-1993,KassVos-1997,IG-2000,AIGStat-2009}:
Professor Harold Hotelling~\cite{Hotelling-1930}  first considered in the late 1920's the {\em Fisher Information Matrix} (FIM) $I$ as a Riemannian metric tensor $g$ (ie., the {\em Fisher Information metric}, FIm), and interpreted a parametric family of probability distributions $M$
as a Riemannian manifold $(M,g)$.
Historically speaking, Hotelling attended the  American Mathematical
Society's Annual Meeting in Bethlehem (Pennsylvania, USA) on December 26--29, 1929, but left before his  scheduled talk on December 27.
His handwritten notes on the ``Spaces of Statistical Parameters'' was read by a colleague and are fully typeset in~\cite{Stigler-2007}. We warmly thank Professor Stigler for sending us the scanned handwritten notes and for discussing by emails some historical aspects of the birth of information geometry.
In this pioneering work, Hotelling mentioned that location-scale probability families yield Riemannian manifolds of constant non-positive curvatures.
This Riemannian modeling of parametric family of densities was further independently studied by Calyampudi Radhakrishna Rao (C.R. Rao) in his celebrated paper~\cite{Rao-1945} (1945) that also includes the Cram\'er-Rao lower bound~\cite{CRLB-IG-2013} and the Rao-Blackwellization technique used in statistics.
Nowadays the induced Riemannian metric distance is often called the {\em Fisher-Rao distance}~\cite{FRMetric-2011} or Rao distance~\cite{RaoGamma-2003}. Yet another use of Riemannian geometry in statistics was pioneered by Harold Jeffreys~\cite{Jeffreys-1946} that proposed to use as an invariant prior the normalized volume element of the Fisher-Riemannian manifold.
In those seminal papers, there was no theoretical justification of using the Fisher information matrix as a metric tensor (besides the fact that it is a well-defined positive-definite matrix for regular identifiable models). Nowadays, this Riemmanian metric tensor is called the {\em information metric} for short. Modern information geometry considers a generalization of this approach using a non-Riemannian dualistic modeling $(M,g,\nabla,\nabla^*)$ which coincides with the Riemannian manifold when $\nabla=\nabla^*=\idxSup{\LC}\nabla$, the Levi-Civita connection (the unique torsion-free affine connection compatible with the metric tensor).
The Fisher-Rao geometry has also been explored in thermodynamics yielding the Ruppeiner geometry~\cite{Ruppeiner-2019}, and the geometry of thermodynamics is called nowadays called {\em geometrothermodynamics}~\cite{Geometrothermodynamics-2007}.

In the 1960's, Nikolai Chentsov (also commonly written {\v{C}}encov) studied the algebraic category of all statistical decision rules with its induced geometric structures:
Namely, the  $\alpha$-geometries (``equivalent differential geometry'') and the dually flat manifolds (``Nonsymmetric Pythagorean geometry'' of the exponential families with respect to the Kullback-Leibler divergence).
In the preface of the english translation of his 1972's russian monograph~\cite{Chentsov-1982}, the field of investigation
 is defined as ``geometrical statistics.'' 
However in the original Russian monograph, Chentsov used the russian term {\em geometrostatistics}.
According to Professor Alexander Holevo, the geometrostatistics term was  coined by Andrey Kolmogorov to define the field of differential geometry of statistical models.
In the monograph of Chentsov~\cite{Chentsov-1982}, the Fisher information metric is shown to be the unique metric tensor (up to a scaling factor) yielding statistical invariance under Markov morphisms (see~\cite{Campbell-1986} for a simpler proof that generalizes to positive measures).

The dual nature of the information geometry was thoroughly investigated  by Professor Shun-ichi Amari~\cite{Amari-1980}.
In the preface of his 1985's monograph~\cite{Amari-1985}, Professor Amari coined the term {\em information geometry} as follows:
``The differential-geometrical method developed in statistics is also
applicable to other fields of sciences such as information theory
and systems theory... They together will
open a new field, which I would like to call {\em information geometry}.''
Professor Amari mentioned in~\cite{Amari-1985} that he considered the Gaussian Riemannian manifold as a hyperbolic manifold in 1959, and was strongly influenced by Efron's paper on statistical curvature~\cite{Efron-1975} (1975) to study the family of $\alpha$-connections in the 1980's~\cite{Amari-1980,NA-IG-1982}.
Professor Amari prepared his PhD under the supervision of Professor Kondo~\cite{Croll-2007}, an expert of differential geometry in touch with Professor Kawaguchi~\cite{Kawaguchi-1960}.
The role of differential geometry in statistics has been discussed in~\cite{DGStat-1986}.

Note that the dual affine connections of information geometry have also been investigated independently in {\em affine differential geometry}~\cite{AffDG-1994} which considers  invariance under volume-preserving affine transformations by defining a volume form instead of a metric form for Riemannian geometry.
The notion of dual parallel transport compatible with the metric is due to Aleksandr Norden~\cite{ConjugateConnections-1945} and Rabindra Nath Sen~\cite{RieParallelismI-1944,RieParallelismII-1945,RieParallelismIII-1946} (See the Senian geometry in \url{http://insaindia.res.in/detail/N54-0728}).

We summarize the main fundamental structures of information manifolds below:
 \vskip 0.3cm

\begin{supertabular}{ll}
$(M,g)$ & Riemannian manifold\\
$(\calP,\leftsub{\calP}g)$ & Fisher-Riemannian (expected) Riemannian manifold\\
$(M,g,\nabla)$ & Riemannian manifold $(M,g)$ with affine connection $\nabla$\\
$(\calP,\leftsub{\calP}g,\leftsub{\calP}\leftsup{e}\nabla^\alpha)$ & Chentsov's manifold with affine exponential $\alpha$-connection\\
$(M,g,\nabla,\nabla^*)$ & Amari's dualistic information manifold\\
$(\calP,\leftsub{\calP}g,\leftsub{\calP}\nabla^{-\alpha},\leftsub{\calP}\nabla^\alpha)$ & Amari's (expected) information $\alpha$-manifold, $\alpha$-geometry\\
$(M,g,C)$ & Lauritzen's statistical manifold~\cite{Lauritzen-1987}\\
$(M,\gD,\nablaD,\nablaDstar)$  & Eguchi's conjugate connection manifold induced by divergence $D$\\
$(M,\gF,\CF)$ & Chentsov/Amari's dually flat manifold induced by convex potential $F$\\
\end{supertabular}
 \vskip 0.3cm

We use the $\equiv$ symbol to denote the equivalence of geometric structures.
For example, we have $(M,g)\equiv (M,g,\nablaLC,\nablaLC^*=\nablaLC)$.

\subsection{Perspectives}
We recommend the two recent textbooks~\cite{IG-2014,IG-2016} for an indepth covering of (parametric) information geometry, and the book~\cite{infIG-2015} for a thorough description of some infinite-dimensional statistical models. (Japanese readers may refer to~\cite{IG-JP-Amari-2014,IG-Fujiwara-2015}.)
We did not report the various coefficients of the metric tensors, Christoffel symbols and skewness tensors for the expected $\alpha$-geometry of common parametric models like the multivariate Gaussian distributions, the Gamma/Beta distributions, etc. They can be found in~\cite{IG-2008,IG-2014} and in various articles dealing with less common family of 
distributions~\cite{locationscaleMfdMitchell-1988,EllipticalIG-Mitchell-1989,invGaussian-IG-2007,IG-inversegamma-2008,DirichletInfoMfd-2008,ParetoInfoMfd-2007,IG-2008}.
Although we have focused on the finite parametric setting, information geometry is also considering non-parametric families of distributions~\cite{nparIG-2013}, and quantum information geometry~\cite{QT-2006}.

\begin{figure}
\begin{center}
\includegraphics[width=\textwidth]{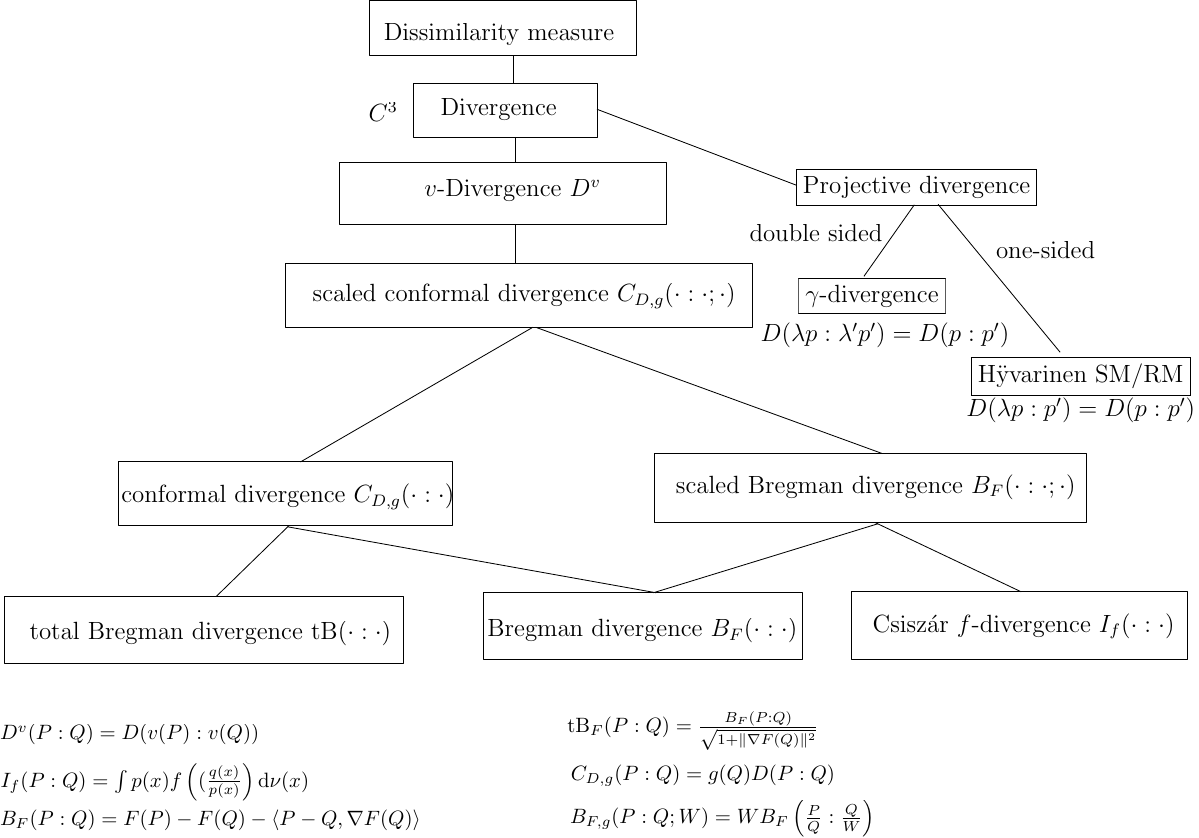}
\caption{{Principled classes of distances/divergences}}\label{fig:classD}
\end{center}
\end{figure}

We have shown that we can always create an information manifold $(M,D)$ from {\em any} divergence function $D$.
It is therefore important to consider generic classes of divergences in applications, that are ideally axiomatized and shown to have exhaustive characteristics. 
Beyond the three main Bregman/Csisz\'ar/Jensen classes (theses classes overlap~\cite{Classes-1997}), we may also mention the 
class of conformal divergences~\cite{ConfDiv-2016,tJ-2015}, the class of projective divergences~\cite{ProjDivPM-2016,HolderDiv-2017}, etc.
Figure~\ref{fig:classD} illustrates the relationships between the principal classes of distances.

There are many perspectives on information geometry as attested by the new Springer journal\footnote{'Information Geometry', \url{https://www.springer.com/mathematics/geometry/journal/41884}}, and the biannual international conference ``Geometric Sciences of Information'' (GSI)~\cite{GSI-13,GSI-15,GSI-17} with its collective post-conference edited books~\cite{IG-GSI-2018,IG-GTI-2014}. 
We also mention the edited book~\cite{IG-IGAIA-2018} on the Occasion of Shun-ichi Amari's 80th birthday.

\noindent {Acknowledgments}:{FN would like to thank the organizers of the {\em Geometry In Machine Learning}  workshop in 2018 (GiMLi, \url{http://gimli.cc/2018/}) for their kind keynote talk invitation, and specially Professor S\o{}ren Hauberg (Technical University of Denmark, DTU). This survey is based on the talk given at GiMLi. I am very thankful to Professor Stigler (University of Chicago, USA) and Professor Holevo (Steklov Mathematical Institute, Russia) for providing me feedback on some historical aspects of the field of information geometry. 
Finally, I would like to express my sincere thanks to Ga\"etan Hadjeres (Sony Computer Science Laboratories Inc, Paris) for his careful proofreading and feedback.
}

\appendix

\section{Monte Carlo estimations of $f$-divergences}\label{sec:MCfdiv}
Let $(X,F,\mu)$ be a probability space~\cite{Keener-2011} with $X$ denoting the sample space, $F$ the $\sigma$-algebra, and $\mu$ a reference positive measure.
The $f$-divergence~\cite{Csiszar-1967,chisquarefdiv-2013} between two probability measures $P$ and $Q$  both absolutely continuous with respect to  a positive measure $\mu$ for a convex generator $f:(0,\infty)\rightarrow\bbR$ strictly convex at $1$ and satisfying $f(1)=0$ is
\begin{equation}
I_f(P:Q)= I_f(p:q) = \int p(x) f\left(\frac{q(x)}{p(x)}\right) \dmu(x),
\end{equation}
where $P=p\dmu$ and $Q=q\dmu$ (i.e., $p$ and $q$ denote the Radon-Nikodym derivatives with respect to $\mu$).
We use the following conventions:
\begin{equation}
0f\left(\frac{0}{0}\right)=0,\quad f(0)=\lim_{u\rightarrow 0^+} f(u),\quad \forall a>0, 0f\left(\frac{a}{0}\right)=\lim_{u\rightarrow 0^+} uf\left(\frac{a}{u}\right)=a\lim_{u\rightarrow \infty} \frac{f(u)}{u}.
\end{equation}

When $f(u)=-\log u$, we retrieve the Kullback-Leibler divergence (KLD):
\begin{equation}
D_\KL(p:q)=\int p(x)\log\frac{p(x)}{q(x)}\dmu(x).
\end{equation}

The KLD is usually difficult to calculate in closed-form, say, for example, between statistical mixture models~\cite{KLD-2016}.
A common technique is to estimate the KLD using Monte Carlo sampling using a proposal distribution $r$:
\begin{equation}
\widehat{\KL}_n(p:q) = \frac{1}{n} \sum_{i=1}^n \frac{p(x_i)}{r(x_i)}\log \frac{p(x_i)}{q(x_i)},
\end{equation}
where $x_1, \ldots, x_n\sim_{\mathrm{iid}} r$.
When $r$ is chosen as $p$, the KLD can be estimated as: 
\begin{equation}\label{eq:klmc}
\widehat{\KL}_{n}(p:q) = \frac{1}{n} \sum_{i=1}^n \log \frac{p(x_i)}{q(x_i)}.
\end{equation}
Monte Carlo estimators are consistent under mild conditions: $\lim_{n\rightarrow\infty} \widehat{\KL}_{n}(p:q)=\KL(p:q)$.

In practice, one problem when implementing Eq.~\ref{eq:klmc}, is that we may end up potentially with $\widehat{\KL}_{n}(p:q)<0$.
This may have  disastrous consequences as algorithms implemented by programs consider non-negative divergences to execute a correct workflow. 
The potential negative value problem of Eq.~\ref{eq:klmc} comes from the fact that $\sum_i p(x_i)\not=1$ and $\sum_i q(x_i)\not=1$.
One way to circumvent this problem is to consider the  {\em extended $f$-divergences}:

\begin{Definition}[Extended $f$-divergence]
The extended $f$-divergence for a convex generator $f$, strictly convex at $1$ and satisfying $f(1)=0$ is defined by 
\begin{equation}
I_f^e(p:q)=\int p(x)\left( f\left(\frac{q(x)}{p(x)}\right)-f'(1)\left(\frac{q(x)}{p(x)}-1\right)   \right)\dmu(x).
\end{equation}
\end{Definition}

Indeed, for a strictly convex generator $f$, let us consider the {\em scalar Bregman divergence}~\cite{Bregman-1967}:
\begin{equation}\label{eq:bds}
B_f(a:b)= f(a)-f(b)-(a-b)f'(b)\geq 0.
\end{equation}

Setting $a=\frac{q(x)}{p(x)}$ and $b=1$ in Eq.~\ref{eq:bds}, and using the fact that $f(1)=0$, we get
\begin{equation}
f\left(\frac{q(x)}{p(x)}\right)-\left(\frac{q(x)}{p(x)}-1\right)f'(1)\geq 0.
\end{equation}

Therefore we define the {\em extended $f$-divergences} as 
\begin{equation}
I_f^e(p:q) = \int p(x) B_f\left(\frac{q(x)}{p(x)}:1\right)\dmu(x)\geq 0.
\end{equation}

That is, the formula for the extended $f$-divergences is
\begin{equation}
I_f^e(p:q)= \int p(x) \left( f\left(\frac{q(x)}{p(x)}\right) - f'(1)\left(\frac{q(x)}{p(x)}-1\right) \right)\dmu(x) \geq 0.
\end{equation}

Then we estimate the extended $f$-divergence using importance sampling of the integral with respect to distribution $r$, using $n$ variates $x_1,\ldots, x_n\sim_{\mathrm{iid}} p$ as:

\begin{equation}
\hat{I}_{f,n}(p:q) =   \frac{1}{n} \sum_{i=1}^n f\left(\frac{q(x_i)}{p(x_i)}\right)-f'(1)\left(\frac{q(x_i)}{p(x_i)}-1\right) \geq 0.
\end{equation}

For example, for the KLD, we obtain the following Monte Carlo estimator:
\begin{equation}\label{eq:eklmc}
\widehat{\KL}_{n}(p:q) = \frac{1}{n} \sum_{i=1}^n \left(\log \frac{p(x_i)}{q(x_i)}+ \frac{q(x_i)}{p(x_i)}-1\right)\geq 0,
\end{equation}
since the extended KLD is
\begin{equation}
D_{\KL^e}(p:q)=\int \left(p(x)\log\frac{p(x)}{q(x)}+q(x)-p(x)\right)\dmu(x).
\end{equation}
Eq.~\ref{eq:eklmc} can be interpreted as a sum of scalar Itakura-Saito divergences since the Itakura-Saito divergence is scale-invariant:
$\widehat{\KL}_{n}(p:q)= \frac{1}{n} \sum_{i=1}^n D_\IS(p(x_i):q(x_i))$ with the scalar Itakura-Saito divergence
\begin{equation}
D_\IS(a:b)=D_\IS\left(\frac{a}{b}:1\right)=\frac{a}{b} - \log \frac{a}{b} -1\geq 0,
\end{equation}
a Bregman divergence obtained for the generator $f(u)=-\log u$.

Notice that the extended $f$-divergence is a $f$-divergence for the generator
\begin{equation}
f_e(u)=f(u)-f'(1)(u-1).
\end{equation}
We check that the generator $f_e$ satisfies both $f(1)=0$ and $f'(1)=0$, and we have $I_f^e(p:q)=I_{f_e}(p:q)$.
Thus $D_{\KL^e}(p:q)=I_{f_\KL^e}(p:q)$ with $f_\KL^e(u)=-\log u+u-1$.

Let us remark that we only need to have the scalar function strictly convex at $1$ to ensure that $B_f\left(\frac{a}{b}:1\right)\geq 0$.
Indeed, we may use the definition of Bregman divergences extended to strictly convex functions but not necessarily smooth functions~\cite{Gordon-1999,BHC-2012}:
\begin{equation}
B_f(x:y)=\max_{g(y)\in \partial f(y)} \{f(x)-f(y)-(x-y)g(y)\},
\end{equation}
where $\partial f(y)$ denotes the subderivative  of $f$ at $y$.

Furthermore, noticing that $I_{\lambda f}(p:q)=\lambda I_f(p:q)$ for $\lambda>0$, we may enforce that $f''(1)=1$, and obtain a {\em standard $f$-divergence}~\cite{IG-2016} 
which enjoys the property that 
$I_f(p_\theta(x):p_{\theta+\dtheta}(x))=\dtheta^\top I(\theta)\dtheta$, where $I(\theta)$ denotes the Fisher information matrix of the parameteric family $\{p_\theta\}_\theta$ of densities.

\section{The multivariate Gaussian family: An exponential family}\label{sec:mvn}

We report the canonical decomposition of the multivariate Gaussian~\cite{Yoshizawa-1999} family $\{N(\mu,\Sigma) \st \mu\in\bbR^d, \Sigma\succ 0\}$ following~\cite{JS-Nielsen-2019}.
The multivariate Gaussian family is also called the {\em MultiVariate Normal} family, or~MVN family for short.

Let $\lambda \eqdef(\lambda_v,\lambda_M)=(\mu,\Sigma)$ denote the {\em composite} (vector,matrix) parameter of an MVN.
The~$d$-dimensional MVN density is given by
\begin{eqnarray}\label{eq:mvnl}
p_\lambda(x;\lambda) &\eqdef&  \frac{1}{(2\pi)^{\frac{d}{2}}\sqrt{|\lambda_M|}}  \exp\left(-\frac{1}{2} (x-\lambda_v)^\top \lambda_M^{-1} (x-\lambda_v)\right),
\end{eqnarray} 
where $|\cdot|$ denotes the matrix determinant.
The natural parameters $\theta$ are also expressed using both a {vector parameter}  $\theta_v$ and a  {matrix parameter}  $\theta_M$ in a compound object $\theta=(\theta_v,\theta_M)$.
By defining the following {\em compound inner product} on a composite (vector,matrix) object
\begin{equation}
\inner{\theta}{\theta'}\eqdef \theta_v^\top \theta_v'+ \mathrm{tr}\left({\theta_M'}^\top\theta_M\right),
\end{equation}
where $\tr(\cdot)$ denotes the matrix trace, we rewrite the MVN density of Equation (\ref{eq:mvnl}) in the canonical form of an exponential family~\cite{EF-2009}:
\begin{eqnarray}
p_\theta(x;\theta) &\eqdef& \exp\left(\inner{t(x)}{\theta}-F_\theta(\theta)\right) = p_\lambda(x;\lambda(\theta)),
\end{eqnarray} 
where 
\begin{equation}
\theta=(\theta_v,\theta_M)=\left(\Sigma^{-1}\mu,-\frac{1}{2}\Sigma^{-1}\right)=\theta(\lambda)=\left(\lambda_M^{-1}\lambda_v,-\frac{1}{2}\lambda_M^{-1}\right),
\end{equation}
 is the {\em compound natural parameter} and 
 \begin{equation}
t(x)=(x,-xx^\top)
\end{equation}
 is the {\em compound sufficient statistic}.
The  function $F_\theta$ is the strictly convex and continuously differentiable log-normalizer defined by:
\begin{equation}
F_\theta(\theta)= \frac{1}{2}\left( d\log\pi -\log |\theta_M|+\frac{1}{2} \theta_v^\top \theta_M^{-1} \theta_v \right),
\end{equation}

The log-normalizer can be expressed using the ordinary parameters, $\lambda=(\mu,\Sigma)$,  as:
\begin{eqnarray}
F_\lambda(\lambda) &=& \frac{1}{2}\left(\lambda_v^\top \lambda_M^{-1}\lambda_v+\log |\lambda_M| + d\log2\pi \right),\\
&=& \frac{1}{2}\left(\mu^\top \Sigma^{-1}\mu +\log |\Sigma| + d\log2\pi \right).
\end{eqnarray}

The {\em moment/expectation parameters}~\cite{IG-2016} are 
\begin{equation}
\eta=(\eta_v,\eta_M)=E[t(x)]=\nabla F(\theta).
\end{equation}

We report the conversion formula between the three types of coordinate systems (namely the ordinary parameter $\lambda$, the~natural parameter $\theta$ and the moment parameter $\eta$) as follows:
\begin{eqnarray}
\left\{
\begin{array}{ll}
\theta_v(\lambda)=\lambda_M^{-1}\lambda_v=\Sigma^{-1}\mu\\
\theta_M(\lambda)=\frac{1}{2}\lambda_M^{-1}=\frac{1}{2}\Sigma^{-1}
\end{array}
\right.  &\Leftrightarrow &
\left\{
\begin{array}{ll}
\lambda_v(\theta) = \frac{1}{2}\theta_M^{-1}\theta_v=\mu\\
\lambda_M(\theta) = \frac{1}{2}\theta_M^{-1}=\Sigma
\end{array}
\right.\\
\left\{
\begin{array}{ll}
\eta_v(\theta)= \frac{1}{2}\theta_M^{-1}\theta_v\\
\eta_M(\theta)= -\frac{1}{2}\theta_M^{-1}-\frac{1}{4}(\theta_M^{-1}\theta_v)(\theta_M^{-1}\theta_v)^\top
\end{array}
\right.
  &\Leftrightarrow &
	\left\{
\begin{array}{ll}
\theta_v(\eta)=  -(\eta_M+\eta_v\eta_v^\top)^{-1}\eta_v\\
\theta_M(\eta)=  -\frac{1}{2}(\eta_M+\eta_v\eta_v^\top)^{-1}
\end{array}
\right.\\
\left\{
\begin{array}{ll}
\lambda_v(\eta)=  \eta_v = \mu\\
\lambda_M(\eta)=  -\eta_M-\eta_v\eta_v^\top=\Sigma
\end{array}
\right.
  &\Leftrightarrow &
	\left\{
\begin{array}{ll}
\eta_v(\lambda)=   \lambda_v=\mu\\
\eta_M(\lambda)=   -\lambda_M-\lambda_v\lambda_v^\top =-\Sigma -\mu\mu^\top
\end{array}
\right.
\end{eqnarray}

The dual Legendre convex conjugate~\cite{IG-2016} is 
\begin{equation}
F^*_\eta(\eta)=-\frac{1}{2}\left( \log(1+\eta_v^\top \eta_M^{-1} \eta_v) + \log |-\eta_M| +d (1+\log 2\pi) \right),
\end{equation}
and $\theta=\nabla_\eta F_\eta^*(\eta)$.

We check the Fenchel-Young equality when $\eta=\nabla F(\theta)$ and $\theta=\nabla F^*(\eta)$:
\begin{equation}
F_\theta(\theta) + F^*_\eta(\eta)  -\inner{\theta}{\eta} =0.
\end{equation}

The Kullback–Leibler divergence between two $d$-dimensional Gaussians distributions $p_{(\mu_1,\Sigma_1)}$ and $p_{(\mu_2,\Sigma_2)}$ 
(with $\Delta_\mu=\mu_2-\mu_1$) is
\begin{eqnarray}
\KL(p_{(\mu_1,\Sigma_1)}:p_{(\mu_2,\Sigma_2)}) 
&=& \frac{1}{2} \left\{ \tr(\Sigma_2^{-1}\Sigma_1) + \Delta_\mu^\top \Sigma_2^{-1}\Delta_\mu +  \log \frac{|\Sigma_2|}{|\Sigma_1|} -d \right\} =\KL(p_{\lambda_1}:p_{\lambda_2}).
\end{eqnarray}

We check that $\KL(p_{(\mu,\Sigma)}:p_{(\mu,\Sigma)})=0$ since $\Delta_\mu=0$ and $\tr(\Sigma^{-1}\Sigma)=\tr(I)=d$.
Notice that when $\Sigma_1=\Sigma_2=\Sigma$, we have
\begin{equation}
\KL(p_{(\mu_1,\Sigma)}:p_{(\mu_2,\Sigma)})  = \frac{1}{2}  \Delta_\mu^\top \Sigma^{-1}\Delta_\mu= \frac{1}{2}D_{\Sigma^{-1}}^2(\mu_1,\mu_2),
\end{equation}
that is half the squared Mahalanobis distance for the precision matrix $\Sigma^{-1}$ (a positive-definite matrix: $\Sigma^{-1}\succ 0$), where the Mahalanobis distance is defined for any positive matrix $Q\succ 0$ as follows:
\begin{equation}
D_Q(p_1:p_2) = \sqrt{(p_1-p_2)^\top Q (p_1-p_2)}.
\end{equation}

The Kullback–Leibler divergence between two probability densities of the same exponential families amount to a Bregman divergence~\cite{IG-2016}:
 \begin{equation}
\KL(p_{(\mu_1,\Sigma_1)}:p_{(\mu_2,\Sigma_2)})  = \KL(p_{\lambda_1}:p_{\lambda_2}) = B_F(\theta_2:\theta_1)=B_{F^*}(\eta_1:\eta_2),
\end{equation}
where the Bregman divergence is defined by
\begin{equation}\label{eq:bdip}
B_F(\theta:\theta') \eqdef F(\theta)-F(\theta')-\inner{\theta-\theta'}{\nabla F(\theta')},
\end{equation}
with $\eta'=\nabla F(\theta')$.
Define the canonical divergence~\cite{IG-2016}
\begin{equation}
A_F(\theta_1:\eta_2)=F(\theta_1)+F^*(\eta_2)-\inner{\theta_1}{\eta_2}=A_{F^*}(\eta_2:\theta_1),
\end{equation}
since ${F^*}^*=F$. We have $B_F(\theta_1:\theta_2)=A_F(\theta_1:\eta_2)$.

\section*{Notations}\label{sec:notations}

\vskip 0.3cm
Below is a list of notations we used in this document:
\vskip 0.3cm
\begin{supertabular}{ll}
$[D]$ & $[D]\eqdef\{1,\ldots, D\}$\\
$\inner{\cdot}{\cdot}$ & inner product\\
$M_Q(u,v)=\|u-v\|_Q$ & Mahalanobis distance $M_Q(u,v)=\sqrt{ \sum_{i,j} (u^i-v^i)(u^j-v^j) Q_{ij}}$, $Q\succ 0$\\
$D(\theta:\theta')$ & parameter divergence\\
$D[p(x):p'(x)]$ & statistical divergence \\
$D$, $D^*$ & Divergence and dual (reverse) divergence\\
Csisz\'ar divergence $I_f$ & $I_f(\theta:\theta') \eqdef \sum_{i=1}^D \theta_i f\left( \frac{\theta_i'}{\theta_i} \right)$ with $f(1)=0$\\
Bregman divergence $B_F$ & $B_F(\theta:\theta') \eqdef F(\theta)-F(\theta')-(\theta-\theta')^\top \nabla F(\theta')$\\
Canonical divergence $A_{F,F^*}$ & $A_{F,F^*}(\theta:\eta')=F(\theta)+F^*(\eta')-\theta^\top\eta'$\\
Bhattacharyya distance & $B_\alpha[p_1:p_2]=-\log \int_{x\in\calX} p_1^{\alpha}(x)p_2^{1-\alpha}(x) \dmu(x)$\\	
Jensen/Burbea-Rao divergence & $J_{F}^{(\alpha)}(\theta_1 : \theta_2)= \alpha F(\theta_1)+ (1-\alpha)F(\theta_2)-  F(\theta_1+ (1-\alpha)\theta_2)$\\
Chernoff information & $C[P_1,P_2] = -\log \min_{\alpha\in(0,1)} \int_{x\in\calX} p_1^{\alpha}(x)p_2^{1-\alpha}(x) \dmu(x)$\\
$F$, $F^*$ & Potential functions related by Legendre-Fenchel transformation\\
$D_\rho(p,q)$ & Riemannian distance $D_\rho(p,q) \eqdef \int_0^1 \|\gamma'(t)\|_{\gamma(t)} \dt$\\
\\	
$B$, $B^*$ & basis, reciprocal basis\\
$B=\{e_1=\partial_1,\ldots, e_D=\partial_D\}$  & natural basis\\
$\{\dx_i\}_i$ & covector basis (one-forms)\\
$(v)_B\eqdef (v^i)$ & contravariant components of vector $v$\\
$(v)_{B^*}\eqdef (v_i)$ & covariant components of vector $v$\\
$u\perp v$ & vector $u$ is perpendicular to vector $v$ ($\inner{u}{v}=0$)\\
$\|v\|=\sqrt{\inner{v}{v}}$ & induced norm, length of a vector $v$\\
\\
$M$, $S$ & Manifold, submanifold \\
$T_p$ & tangent plane at $p$ \\
$TM$ & Tangent bundle $TM=\cup_p T_p=\{(p,v), p\in M, v\in T_p\}$\\
$\frF(M)$ & space of smooth functions on $M$\\
$\frX(M)=\Gamma(TM)$ & space of smooth vector fields on $M$\\
$v f$ & direction derivative of $f$ with respect to vector $v$\\
$X, Y, Z\in\frX(M)$ & Vector fields\\
$g \eqsum g_{ij} \dx_i \otimes \dx_j$ & metric tensor (field)\\
$(\calU,x)$ & local coordinates $x$ in a chat $\calU$\\
$\partial_i\eqNota \frac{\partial}{\partial x_i}$ & natural basis vector\\
$\partial^i\eqNota \frac{\partial}{\partial x^i}$ & natural reciprocal basis vector\\
\\
$\nabla$ & affine connection\\
$\nabla_X Y$ & covariant derivative\\
$\prod^\nabla_c$ & parallel transport of vectors along a smooth curve $c$\\
 $\prod^\nabla_c v$ & Parallel transport of $v\in T_{c(0)}$ along a smooth curve $c$\\
$\gamma$, $\gamma_\nabla$& geodesic, geodesic with respect to connection $\nabla$\\
$\Gamma_{ij,l}$ & Christoffel symbols of the first kind (functions)\\
$\Gamma_{ij}^k$ & Christoffel symbols of the second kind (functions)\\
$R$ & Riemann-Christoffel curvature  tensor \\
$[X,Y]$ & Lie bracket $[X,Y](f)=X(Y(f))-Y(X(f)), \forall f\in\frF(M)$\\
$\nabla$-projection & $P_S=\arg\min_{Q\in S} D(\theta(P):{\theta(Q)})$\\
$\nabla^*$-projection & $P_S^*=\arg\min_{Q\in S} D({\theta(Q)}:\theta(P))$\\
$C$ & Amari-Chentsov totally symmetric cubic $3$-covariant tensor\\
\\
$\calP=\{p_\theta(x)\}_{\theta_in\Theta}$ & parametric family of probability distributions\\
$\calE,\calM,\Delta_D$ & exponential family, mixture family, probability simplex\\
$\leftsub{\calP}I(\theta) $ Fisher information matrix\\
$\leftsub{\calP}I(\theta)$ & Fisher Information Matrix (FIM) for a parametric family $\calP$\\
$\leftsub{\calP}g$ & Fisher information metric tensor field\\
exponential connection $\leftsubsup{\calP}{e}\nabla$	& $\leftsubsup{\calP}{e}\nabla  \eqdef E_\theta\left[(\partial_i\partial_j l)(\partial_k l)\right]$\\
mixture connection $\leftsubsup{\calP}{m}\nabla$ &	$\leftsubsup{\calP}{m}\nabla   \eqdef E_\theta\left[(\partial_i\partial_j l+\partial_i l\partial_j l)(\partial_k l)\right]$\\
expected skewness tensor $C_{ijk}$ &	$C_{ijk} \eqdef E_\theta\left[\partial_i l\partial_j l \partial_k l\right]$\\
expected $\alpha$-connections &	$\leftsub{\calP}{\Gamma^{\alpha}}_{ij}^k \eqdef -\frac{1+\alpha}{2} C_{ijk} =
	E_\theta\left[\left(\partial_i\partial_j l+\frac{1-\alpha}{2}\partial_i l\partial_j l\right)(\partial_k l)\right]$\\
$\equiv$ & equivalence of geometric structures\\
\end{supertabular}

\bibliographystyle{plain}
\bibliography{ElementaryIGBib}

\end{document}